\pdfoutput=1
\documentclass[lettersize,journal]{IEEEtran}
\usepackage{amsmath,amsfonts}
\usepackage{algorithm}
\usepackage{array}
\usepackage{textcomp}
\usepackage{stfloats}
\usepackage{url}
\usepackage{verbatim}
\usepackage{graphicx}
\usepackage{cite}
\usepackage{color}
\usepackage{colortbl}
\usepackage{soul}
\usepackage{multirow}
\usepackage{multicol}
\usepackage{makecell}
\usepackage{booktabs}
\usepackage{hyperref}
\usepackage{amssymb}
\usepackage{pifont}
\usepackage{wasysym}
\usepackage{subfigure}
\usepackage{amsfonts}
\usepackage{enumitem}
\usepackage{bm}
\usepackage{algpseudocode}
\algrenewcommand\algorithmicindent{1.0em}
\usepackage[utf8]{inputenc}
\usepackage{threeparttable}
\usepackage[table]{xcolor}
\usepackage{xspace}
\newcolumntype{P}[1]{>{\centering\arraybackslash}p{#1}}
\newcolumntype{C}[1]{>{\centering\arraybackslash}m{#1}}
\newcommand{\rt}[1]{#1}
\newcommand{\bt}[1]{#1}

\newcommand{\diag}{{\rm{diag}} }
\newcommand{\gt}{\textcolor[rgb]{0,1,0}}

\newcommand{\ti}{\textit}
\newcommand{\tb}{\textbf}

\newcommand{\methodname}{\textsc{OpenNavMap}\xspace}
\newcommand{\ie}{\textit{i.e.,\ }}
\newcommand{\eg}{\textit{e.g.,\ }}
\begin{document}

\title{\methodname: Multi-Session Appearance-Based Topometric Mapping for Scalable Visual Navigation}

\author{
  Jianhao Jiao$^{1}$,
  Changkun Liu$^{2}$, Jingwen Yu$^{3}$,
  Boyi Liu$^{3}$, Qianyi Zhang$^{4}$,
  Yue Wang$^{5}$, Dimitrios Kanoulas$^{1, 6}$
  \thanks{\textsuperscript{1}Robot Perception and Learning Lab, Dept. of Computer Science, University College London, London WC1E 6BT, UK. {\tt\small \{ucacjji, d.kanoulas\}@ucl.ac.uk}.}
  \thanks{\textsuperscript{2}Dept. of Computer Science and Engineering, Hong Kong University of Science and Technology, Hong Kong, China.}
  \thanks{\textsuperscript{3}Dept. of Electronic and Computer Engineering, Hong Kong University of Science and Technology, Hong Kong, China.}
  \thanks{\textsuperscript{4}Institute of Robotics and Automatic Information System, Nankai University, Tianjin 300350, China.}
  \thanks{\textsuperscript{5}Zhejiang University, Hangzhou, Zhejiang, China.}
  \thanks{\textsuperscript{6}D. Kanoulas is also with the AI Centre, Dept. of Computer Science, University College London, London WC1E 6BT, UK, and Archimedes/Athena RC, Greece.}
  \thanks{This work was supported by the UKRI Future Leaders Fellowship [MR/V025333/1] (RoboHike).  For the purpose of Open Access, the author has applied a CC BY public copyright license to any Author Accepted Manuscript version arising from this submission.}
}

\maketitle

\begin{abstract}
  \bt{
    Scalable and maintainable maps are fundamental to large-scale navigation and the long-term deployment of robots in real-world environments.
    However, conventional maps that explicitly maintain dense geometry or 3D landmarks incur high storage and maintenance costs, while the core challenge of scaling to multi-session mapping is visual localization under sparse viewpoint overlap, temporal appearance shifts, and cross-device variance.
    To address this, we propose \methodname, a lightweight, landmark-free topometric mapping system that organizes image nodes into covisibility, odometry, and traversability graphs and delegates local geometry recovery to 3D geometric foundation models (GFMs) on demand.
    For localization, dynamic-programming-based sequence matching narrows candidate correspondences for a GFM, reducing global estimation to a lightweight, pose-only optimization; for mapping, a lifelong pipeline fuses multi-session, multi-device data via cross-device merging and node culling.
    On a $19$km dataset across four real-world environments, \methodname attains a state-of-the-art $0.62$m translation error on the Map-Free benchmark, bounds the absolute trajectory error below $3$m across $15.7$km without depth sensors, and completes $12$ autonomous image-goal visual navigation tasks on both simulated and physical robots.
    Code and datasets will be made publicly available at \url{https://rpl-cs-ucl.github.io/OpenNavMap_page}.}
\end{abstract}

%%%%%%%%%%%% Version 1
% This paper proposes \methodname, a multi-session mapping system designed for scalable visual navigation.
%   Rather than relying on the 3D structure-based representation of the environment, \methodname adopts a robust collaborative localization strategy to facilitate map merging, taking only 2D images as input.
%   The resulting topometric map is thus lightweight and structure-free, composed of three layered graphs: covisibility, odometry, and traversability.
%   This design enables autonomous visual navigation without the need for prior structure-based maps.
%   \methodname also supports cross-device localization, making crowdsourced mapping feasible and significantly improving mapping efficiency and coverage.
%   Extensive experiments on collaborative localization and map merging demonstrate that \methodname achieves high accuracy ($<3m$ ATE over $15km$) and strong robustness to challenging conditions such as day-night transitions and large viewpoint changes. The system has been successfully deployed on a quadruped robot using only monocular RGB inputs for image-goal visual navigation.
%   The code and datasets will be publicly available.
% The code and datasets will be publicly available at \url{https://rpl-cs-ucl.github.io/opennavmap}.
%%%%%%%%%%%%%%%%%%%%%%%%%

\begingroup
\section{Introduction}
\label{sec:introduction}

%%%%%%%%%%%%%%%%%%%%%%%%%%%%%%%%%%%%%%%%%%%%%%%%%%%%%%%%%%%%
%%%%%%%%%%%%%%%%%%%%%%%%%%%%%%%%%%%%%%%%%%%%%%%%%%%%%%%%%%%%
%%%%%%%%%%%%%%%%%%%%%%%%%%%%%%%%%%%%%%%%%%%%%%%%%%%%%%%%%%%%
\subsection{Motivation}
\label{sec:intro_motivation}

\begin{figure}[t]
  \begin{center}
    \includegraphics[width=0.94\linewidth]{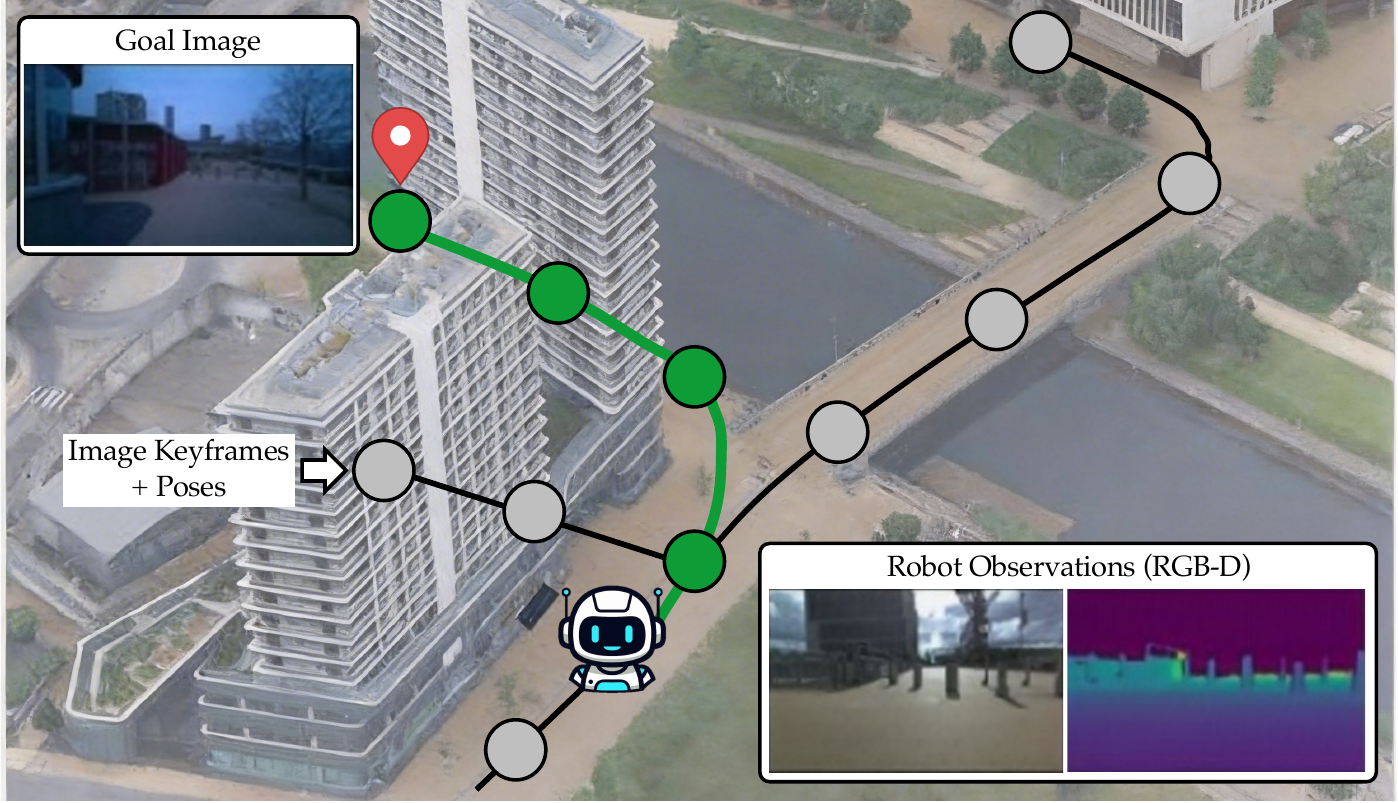}
  \end{center}
  \vspace{-0.3cm}
  \caption{\bt{Conceptual illustration of the topometric map built by \methodname from multi-session mapping, shown with the traversability layer. Nodes in grey form a global traversable graph; the shortest path to a user-specified image goal is highlighted in green. At runtime, the robot uses RGB-D sensing to localize against the map and follow the planned path.}}
  \label{fig:viz_cover}
  \vspace{-0.4cm}
\end{figure}

\bt{As robots scale toward large-scale, long-term real-world autonomy, scalable environmental mapping becomes indispensable for providing a globally consistent navigation reference.
Common navigation systems \cite{lee2024learning,oleynikova2017voxblox} depend on dense metric maps, such as point clouds and occupancy grids, to explicitly reconstruct fine-grained 3D geometry.
In a dedicated prior mapping phase, it extracts traversable space and landmarks and encodes offline cues (\eg out-of-view or occluded regions), relaxing the robot's need to recover complete cues from uncertain online observations.
However, for all its accuracy, this map-intensive design faces two limitations: such maps incur substantial storage overhead and are brittle to dynamics, as moving objects corrupt the reconstruction and force elaborate post-processing to sustain it \cite{duberg2024dufomap}.
Even sparse landmark-based maps reduce this cost \cite{campos2021orb,qin2018vins,xu2024d}, but the scalability bottleneck persists: jointly optimizing a growing set of camera poses and 3D landmarks is increasingly expensive, and lifelong operation demands continual storage, data association, and re-optimization of these global landmarks \cite{sarlin2019coarse}.}

\bt{To overcome this scalability issue, we eliminate the reliance on global metric structures and adopt an alternative design principle: decoupling the coarse-grained global representation from fine-grained local geometry.
A representative paradigm is visual teach-and-repeat (VT\&R) navigation \cite{furgale2010visual}, which replays recorded image sequences without dense geometry or global landmark optimization, but is commonly restricted to single-session mapping.
Building on recent advances in map-free visual localization \cite{arnold2022map} and feed-forward 3D geometric foundation models (GFMs) \cite{leroy2024grounding,wang2025vggt,wang2024dust3r,wen2025foundationstereo}, we generalize this idea into an appearance-based topometric map \cite{paton2016bridging}.
Unlike traditional approaches \cite{campos2021orb}, it avoids optimizing global 3D structures, modeling the environment as lightweight 2D image nodes connected by relational edges (\eg spatial or covisibility): this coarse-grained topology handles global pose consistency and path planning, while fine-grained obstacle is perceived online via GFM-inferred local depth.
This scheme thus keeps the map lightweight and eases the demands on sensing hardware and data quality.
Crucially, this allows us to move beyond single-session VT\&R and scale to large-scale, multi-session, crowdsourced mapping with everyday consumer devices.}

\subsection{Challenges}
\label{sec:intro_challenge}

\bt{Scaling to multi-session mapping allows distributed agents asynchronously aggregate observations beyond the spatio-temporal coverage of a single agent can reach.
We favor a vision-centric solution, since cameras are small, low-cost, and ubiquitous on personal devices (\eg smartphones and AR/VR glasses), relaxing the strict reliance on LiDAR \cite{lee2024learning} or GNSS.
The core problem of multi-session vision-based mapping is visual localization, \ie accurately localizing independently captured sessions against one another to fuse them into one globally consistent map.
This poses two coupled challenges:}

\subsubsection{Visual Localization on a Landmark-Free Map}

\bt{
Even though the map stores no prior 3D landmarks and relies solely on keyframe images, its visual localization is required to match the accuracy of landmark-based methods.
This is hard because unconstrained crowdsourced capture offers no control over the acquisition process, breaking three assumptions that conventional landmark-based pipelines rely on \cite{jafarzadeh2021crowddriven,campos2021orb}, such as dense viewpoints, stable appearance, and a single calibrated camera, and thereby imposing three practical challenges:
\ti{a) Sparse Spatial Overlap:} arbitrary user trajectories overlap only marginally, leaving wide baselines and scarce co-visible regions under which the recall of place recognition and sequence matching drops sharply \cite{milford2012seqslam,lajoie2019modeling}.
\ti{b) Temporal Appearance Shift:} keyframes captured at disparate times (\eg day versus night or across seasons) undergo severe illumination and appearance changes that break appearance-based matching \cite{berton2025megaloc,yin2025general}.
\ti{c) Cross-Device Variance:} heterogeneous devices yield diverse camera and projection models (\eg perspective and panoramic), whose inconsistent fields of view and distortions degrade cross-device matching \cite{sarlin2022lamar,huang2024360loc,blum2025crocodl}.}

% \bt{The map must deliver localization accuracy comparable to landmark-based methods, even though it stores no prior 3D landmarks and relies solely on keyframe images.
% Unconstrained crowdsourced capture makes this hard by violating the assumptions of dense viewpoints, stable appearance, and a single camera model that conventional pipelines rely on \cite{jafarzadeh2021crowddriven}, in three respects:
% \ti{a) Sparse Spatial Overlap:} arbitrary user trajectories overlap only marginally, leaving wide baselines and few co-visible regions, under which the recall of place recognition and sequence matching drops sharply \cite{milford2012seqslam,lajoie2019modeling}.
% \ti{b) Temporal Appearance Shift:} data captured across disparate times (\eg day versus night, across seasons) induces severe illumination and appearance changes that cause appearance-based matching to fail \cite{berton2025megaloc,yin2025general}.
% \ti{c) Cross-Device Variance:} heterogeneous devices yield diverse camera types and projection models (\eg perspective and panoramic), causing inconsistent fields of view and distortions that degrade cross-device matching \cite{sarlin2022lamar,huang2024360loc,blum2025crocodl}.}

\subsubsection{A Globally Consistent and Unified Lifelong Mapping System for Navigation}

\bt{Beyond per-session localization accuracy, these disconnected sessions must be fused into a single global topometric map that remains a unified and reliable navigation reference throughout its lifetime.
This requires two key properties: global consistency and lifelong updatability.
First, the sessions share no common reference frame, so their alignment relies solely on robust cross-session localization under the challenges above; any residual drift then accumulates and breaks the global consistency that navigation demands.
Second, and most critically, the map must be lifelong: as sessions are continually appended, it grows unbounded and demands a pruning mechanism that curtails redundancy without eroding localization accuracy.}

\subsection{Contributions}
\label{sec:intro_contribution}

\bt{This paper introduces \methodname, a vision-centric crowdsourcing system designed for large-scale, multi-session map construction and continual expansion.
The prefix ``Open'' reflects our vision for a publicly accessible mapping solution, inspired by platforms such as Google Street View \cite{anguelov2010google} and OpenStreetMap \cite{OpenStreetMap}, where crowdsourced data aggregation inherently fosters scalable environmental mapping.
At its core, \methodname represents the world as a unified, appearance-based topometric map organized into three graph layers that share the same underlying image nodes:
\ti{1) Covisibility Graph:} encodes strong visual overlap between nodes to ensure robust appearance-based localization.
\ti{2) Odometry Graph:} records sequential connections during data acquisition, serving as the foundational skeleton for pose graph optimization (PGO) and map maintenance.
\ti{3) Traversability Graph:} establishes edges between physically adjacent nodes, providing the topological basis for shortest-path global planning.
Based on this representation, \methodname addresses the two challenges above through the following \ti{contributions}:}
\begin{enumerate}
    \item \bt{\tb{Robust Hierarchical Visual Localization:} We propose a dynamic-programming (DP)-based flexible sequence matching for topological localization, with geometric verification (GV) to reject false positives. 
    Once keyframe correspondences are fixed, we employ a 3D GFM to unify multi-device, multi-image, and multi-time observations into a single metric localization, yielding notable gains under the sparse keyframe distribution of crowdsourced sequences. Unlike the 3D landmark-based paradigm \cite{campos2021orb,schoenberger2016sfm}, we recover local landmarks online with the GFM, removing the need for dense landmark storage and global structure optimization; the global estimation thus reduces to a low-complexity pose-only optimization. A confidence map calibration step further suppresses the GFM feed-forward prediction errors (Sec.~\ref{sec:collaborative_mapping}).}
    \item \bt{\tb{Lifelong Topometric Mapping System:} Enabled by the above shift in localization, we design a topometric mapping system that fuses multi-session, multi-device data into one globally consistent map and sustains it through continual updating, including cross-device merging, PGO, and accuracy-preserving node culling. This advances beyond conventional VT\&R toward an updatable, multi-purpose visual navigation map (Sec.~\ref{sec:crowdsourcing} and Sec.~\ref{sec:method_data_pruning}).}
\end{enumerate}

\bt{To the best of our knowledge, \methodname is the first system to successfully integrate crowdsourced multi-session mapping with an appearance-based topometric map explicitly designed for navigation.
Its modular architecture seamlessly accommodates evolving visual place recognition (VPR) models and 3D GFMs, ensuring sustained relevance alongside algorithmic advancements.
We rigorously validate the proposed mapping system across four diverse real-world environments, including laboratories, campuses, shopping centers, and vineyards, using a $19$km dataset captured via heterogeneous devices over several months.
For metric localization, \methodname achieves state-of-the-art (SoTA) accuracy: it outperforms both structure-from-motion (SfM) and regression-based baselines on the Map-Free dataset, attaining an average translation error of $0.62$m despite a drastically different map representation.
For full-system mapping, it sustains a globally consistent topometric map across $15.7$km of multi-session data, restricting the absolute trajectory error (ATE) to below $3$m without relying on depth sensors (see Tab.~\ref{tab:exp_collaborative_mapping}).}

\bt{
This multi-layer design makes the lightweight map actionable: the covisibility and traversability graphs jointly supply visual localization and planning cues while remaining free of dense geometry, supporting the full real-time image-goal visual navigation pipeline. We substantiate this utility through $12$ autonomous navigation tasks across simulated and physical robots in Fig.~\ref{fig:exp_simu_vnav}-\ref{fig:exp_vnav_ops_around}, confirming that crowdsourced multi-session maps transfer to real-world deployment. To foster future research, code and datasets will be open-sourced.}

\section{Related Work}
\label{sec:related_work}

This section briefly reviews the literature on related map representations for visual navigation and vision-based localization for multi-session mapping.

\subsection{Map Representations for Visual Navigation}
\label{sec:related_work_map_rep}

\bt{Mapping is the process of constructing a symbolic model of the environment using sensor data~\cite{slam-handbook}. 
Existing map representations for navigation can be classified based on how they encode geometric information.}

\bt{\tb{\ti{Dense Geometric Maps}} represent continuous environments precisely via primitives such as point clouds and occupancy grids, explicit 3D Gaussians~\cite{kerbl20233d}, or implicit neural networks such as NeRF~\cite{martin2021nerf}.
However, these maps are fundamentally bound by the need to enforce global metric accuracy, making them prone to error accumulation.
\tb{\ti{Sparse Landmark-Based Maps}} reduce this burden by discarding continuous surfaces in favor of discrete 3D feature points or landmarks~\cite{mur2015orb}.
\tb{\ti{Landmark-Free Maps}} prioritize maps' lightweightness through abstract graph representations, such as topological maps for environment connectivity \cite{Dang2020}, or topometric maps with spatial node positions \cite{jiao2025litevloc}.
However, this efficiency comes at the expense of explicit 3D geometry, thus limiting fine-grained obstacle avoidance and path planning.}

\begin{figure*}[t]
  \begin{center}
    \includegraphics[width=0.95\linewidth]{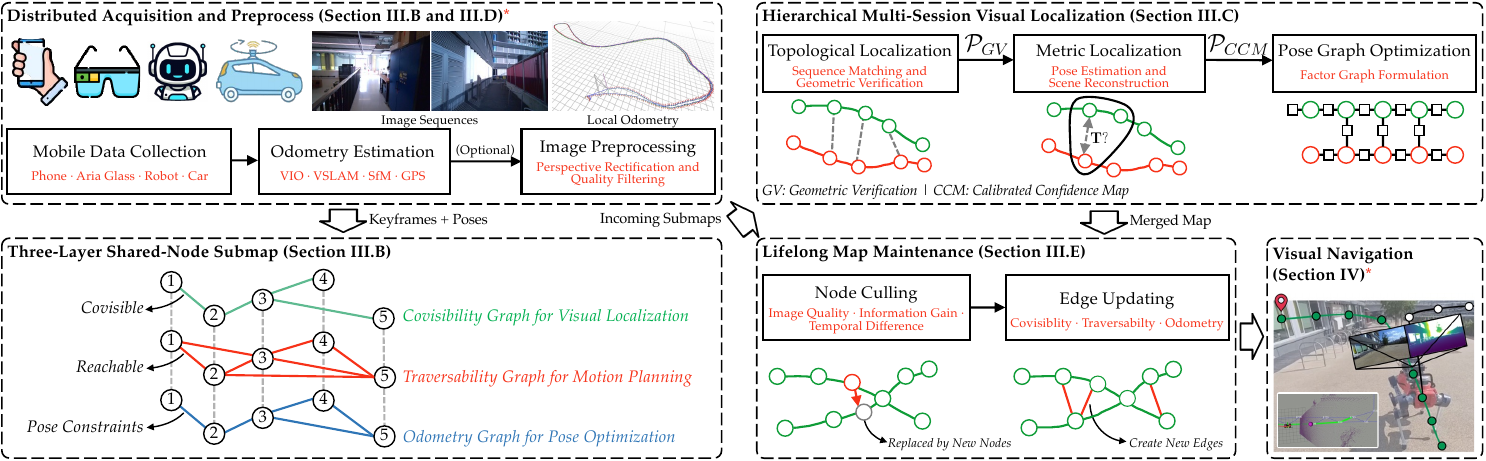}
  \end{center}
  \vspace{-0.3cm}
  \caption{\bt{The \methodname pipeline, with several modules. (a) Disconnected submaps are built independently from distributed devices, with optional image preprocessing steps for low-quality images. (b) Submaps are represented as a three-layer shared-node submap. (c) Hierarchical multi-session visual localization computes inter-submap transformations and merge them. (d) The lifelong map maintenance prunes low-information nodes and updates edge connectivity. (e) The resulting map deployed directly for the downstream image-goal visual navigation. The ``*'' denotes navigation logs as a prospective acquisition source for lifelong deployment as the future work.}}
  \label{fig:overview_system}
  \vspace{-0.4cm}
\end{figure*}

\bt{\tb{\ti{Hybrid Maps}} integrate the complementary strengths of metric and landmark-free representations to seek a balance between geometric accuracy and storage efficiency.
A common strategy involves maintaining explicit dense geometric maps alongside corresponding topometric maps derived from navigable free space; the former facilitates precise navigation, while the latter supports efficient global planning~\cite{lee2024learning}. 
More complex hierarchical representations, such as the scene graphs~\cite{rosinol2021kimera,bavle2022situational,bavle2023s,hughes2022hydra}, organize metric-semantic information across multiple abstraction layers, enabling the efficient management of object- and room-level details.
Additionally, several hybrid approaches construct the topometric map that retain only essential local geometric details~\cite{furgale2010visual}.
This solution forms the foundation for systems such as VT\&R \cite{walcott2012dynamic,furgale2010visual,paton2016bridging} and experience-based navigation~\cite{churchill2013experience}.}

\bt{Unlike maps that store explicit global or local geometry, \methodname adopts an appearance-based topometric map of image-snapshot nodes and relative-pose edges, delegating local reconstruction and pose estimation to on-demand 3D GFMs \cite{leroy2024grounding}.
This revisits the VT\&R paradigm: the \ti{Teach} phase replaces a single manual demonstration with crowdsourced multi-session data from heterogeneous devices, whose recorded paths encode traversable regions, while the \ti{Repeat} phase replaces brittle direct visual tracking with global topometric planning and local depth-aware obstacle avoidance for robustness to viewpoint and dynamic changes.}

\subsection{Visual Localization for Multi-Session Mapping}
\label{sec:related_world_vloc}
\bt{Visual localization estimates the 6-DoF camera pose against a pre-built reference map, providing the spatial alignment required to merge multi-session observations. We classify existing approaches into hierarchical and direct methods.}

\bt{\tb{\ti{Hierarchical Methods}} operate in two stages:
topological localization to identify corresponding images between sessions, followed by metric localization to compute precise camera poses.
The topological stage commonly relies on VPR, employing global image representations such as Bag-of-Words~\cite{galvez2012bags} or learned descriptors such as NetVLAD~\cite{arandjelovi2015netvlad}.
Sequence-based methods \cite{milford2012seqslam,garg2021seqnet} improve robustness to perceptual aliasing~\cite{lajoie2019modeling} by exploiting temporal coherence across consecutive images, though they require careful tuning of hyper-parameters such as sequence length.
Metric localization typically refines poses on explicit landmark-based maps~\cite{pan2024robust,liu2024slideslam,tian2022kimera} via Perspective-n-Point (PnP)~\cite{gao2003complete} with RANSAC~\cite{fischler1981random}, or, in map-free variants, via depth-prediction PnP or RAFT-based backbones coupling optical flow with differentiable solvers~\cite{lipson2024multi}.
Coarse-to-fine frameworks currently dominate multi-session mapping~\cite{karrer2018cvi,tian2022kimera,xu2024d}.}

\bt{\tb{\ti{Direct Methods}} bypass explicit topological localization and instead estimate the final 6-DoF pose in a single step.
Typical approaches include scene coordinate regression (SCR)~\cite{brachmann2023accelerated}, which predicts 3D world coordinates for 2D query pixels before solving for pose, and absolute pose regression (APR)~\cite{kendall2015posenet}, which regresses the camera pose directly in a single feed-forward pass.
However, both require training a scene-specific model and remain insufficiently robust for large scenes, particularly compared to hierarchical pipelines.}

\bt{\tb{\ti{Cross-Device Localization}} addresses the challenge of registering observations from heterogeneous devices with differing motion patterns, viewpoints, and sensor modalities~\cite{sarlin2022lamar,wei2024fusionportablev2,blum2025crocodl}.
The CrowdDriven~\cite{jafarzadeh2021crowddriven} and CroCoDL~\cite{blum2025crocodl} benchmarks quantified these limitations, revealing failures under drastic device heterogeneity and viewpoint changes, while $360$Loc~\cite{huang2024360loc} specifically examined perspective-to-omnidirectional alignment.
Cross-modal matchers such as XoFTR~\cite{tuzcuouglu2024xoftr} and MatchAnything~\cite{he2025matchanything} have recently addressed these gaps through large-scale pre-training.}

\bt{Building on the hierarchical framework, \methodname refines both stages for multi-session mapping.
In the topological stage, it designs the DP-based sequence matching that removes fixed sequence-length hyper-parameters and improves recall under severe appearance and viewpoint changes, paired with a GV module to reject false matches.
In the metric stage, it replaces pre-stored 3D point clouds with on-demand reconstruction from 3D GFMs, and introduces a residual-based confidence calibration mechanism~\cite{lu2025lora3d} with closed-loop feedback to suppress their forward prediction errors during joint optimization of local structure and camera poses.}

\section{Methodology}
\label{sec:methodology}

\bt{
    This section presents the methodology of \methodname, beginning with the following system overview that introduces the overall pipeline and cross-references the components detailed in the subsequent subsections.}

\subsection{System Overview}
\label{sec:system}

\bt{
    As illustrated in Fig.~\ref{fig:overview_system}, \methodname is a scalable, appearance-based, multi-session topometric mapping system for visual navigation.
    Its map adopts a three-layer structure comprising the covisibility graph, odometry graph, and traversability graph (Sec.~\ref{sec:preliminary}), which support visual localization, scene geometry recovery, and path planning, respectively.
    Each device first builds a local submap independently, and these submaps are then merged on a server into a globally consistent frame through a hierarchical multi-session localization pipeline (Sec.~\ref{sec:collaborative_mapping}): inter-submap topological localization with GV, metric localization of matched node pairs, and PGO.
    To accommodate heterogeneous devices, the system adapts to variable camera intrinsics and perceptual degradation (Sec.~\ref{sec:crowdsourcing}), while a probabilistic node-culling strategy bounds map growth for lifelong scalability (Sec.~\ref{sec:method_data_pruning}).
    The resulting map enables vision-only image-goal navigation in GNSS-denied environments, guiding robots to autonomously reach query targets (Sec.~\ref{sec:vnav_system}).}
\subsection{Topometric Mapping Design}
\label{sec:preliminary}

\subsubsection{Map Representation}
\label{sec:map_representation}
\bt{
We formally represent the environment as a topometric map $\mathcal{M}^W = \{\mathcal{G}^W_C, \mathcal{G}^W_O, \mathcal{G}^W_T\}$ defined in the global world frame $W$, comprising three graph layers: the \tb{covisibility graph} $\mathcal{G}^W_C$, the \tb{odometry graph} $\mathcal{G}^W_O$, and the \tb{traversability graph} $\mathcal{G}^W_T$.}
\bt{Each layer is an undirected graph $\mathcal{G} = \{\mathcal{N}, \mathcal{E}\}$, where $\mathcal{N}$ is the set of nodes and $\mathcal{E}$ is the set of edges.
All three layers share a common node set $\mathcal{N}^{W}_{C} = \{\mathbf{n}^{W}_{C_{i}}\}_{i=1}^{N}$ defined by the covisibility graph, with $N$ denoting the total number of nodes.
Only covisibility graph nodes carry the full visual attributes required for localization; odometry and traversability graph nodes use only the shared pose attributes.
Each covisibility graph node $\mathbf{n}^{W}_{C_{i}}$ stores:
\begin{itemize}[leftmargin=0.7cm]
  \item \textbf{Visual Data:} The RGB image ($\mathbf{I}_{i}\in\mathbb{R}^{w\times h\times 3}$), its global descriptor ($\mathbf{d}_{i}\in\mathbb{R}^{d}$), and its timestamp ($\tau_{i}\in\mathbb{R}^{+}$).
  \item \textbf{Pose Information:} The globally consistent pose ($\mathbf{T}^W_{i}\in SE(3)$) w.r.t. the world frame.
  \item \textbf{Quality Metric:} An image quality assessment (IQA) score ($q_{i}\in[0, 100]$) used for node culling.
\end{itemize}}

\bt{
  The three layers differ in their edge semantics.
  Covisibility graph edges encode the covisibility strength ($v_{C_{ij}}\in\mathbb{R}^{+}$) between nodes, supporting visual localization.
  Odometry graph edges encode relative pose constraints $(\mathbf{T}_{ij},\mathbf{\Sigma}_{ij})$ derived from local odometry or relative pose estimation (Sec.~\ref{sec:collaborative_mapping}), forming a pose graph for PGO.
  Traversability graph edges store traversability costs ($v_{T_{ij}}\in\mathbb{R}^{+}$) reflecting motion feasibility, supporting global path planning.
  Crucially, connectivity across layers is independent: two nodes may share a covisibility edge and an odometry edge yet have no traversability edge if a physical obstacle blocks their direct path.}

\subsubsection{Submap Construction from Distributed Devices}
\label{sec:submap_construction}
\bt{
  Each device independently constructs a local submap following three sequential steps.
  \ti{1)} The device captures images and estimates per-frame poses $\mathbf{T}^{S}_{i} \in SE(3)$ in a local session frame $S$ using any available odometry source (\eg VIO, visual SLAM, SfM, or GPS), forming an ordered sequence of nodes.
  \ti{2)} Covisibility graph edges are added between nodes sharing sufficient visual overlap such as a minimum number of co-visible feature points, and odometry graph edges $(\mathbf{T}^{S_i}_{S_j}, \mathbf{\Sigma}^{S_i}_{S_j})$ are established between consecutive nodes as local pose constraints.
  \ti{3)} Traversability graph edges connect spatially adjacent nodes along the recorded traversal path, implicitly encoding the navigable regions of the environment.
  These submaps are subsequently merged into a unified global map through multi-session mapping.
  This framework is device-agnostic, accommodating smartphones (\eg ARKit on iPhones), AR glasses~\cite{engel2023project}, GPS-enabled $360^\circ$ cameras (\eg Insta360), ground vehicles~\cite{wei2024fusionportablev2,zheng2024fast}, and geo-referenced repositories~\cite{anguelov2010google}, enabling high-efficiency map growth across both spatial and temporal scales.}

\subsubsection{Geometric Foundation Models for Stereo Reconstruction}
\label{sec:recap_dust3r}
\bt{
  \methodname employs MASt3R~\cite{leroy2024grounding}, a 3D GFM, for on-demand stereo reconstruction.
  Given a pair of uncalibrated RGB images $\mathbf{I}_1, \mathbf{I}_2 \in \mathbb{R}^{w \times h \times 3}$, the network $f_{\bm{\theta}}(\cdot)$ jointly predicts dense pointmaps and per-pixel confidence maps:
  \begin{equation}
    \label{equ:dust3r_network}
    \left(\mathbf{X}^{1,1}, \mathbf{C}^{1,1}\right),\ \left(\mathbf{X}^{2,1}, \mathbf{C}^{2,1}\right) = f_{\bm{\theta}}(\mathbf{I}_1,\mathbf{I}_2),
  \end{equation}
  where $\mathbf{X}^{1,1}, \mathbf{X}^{2,1} \in \mathbb{R}^{w \times h \times 3}$ are the predicted pointmaps of $\mathbf{I}_1$ and $\mathbf{I}_2$ expressed in the coordinate frame of $\mathbf{I}_1$, and $\mathbf{C}^{1,1}, \mathbf{C}^{2,1} \in \mathbb{R}^{w \times h \times 1}$ are the corresponding confidence maps measuring prediction reliability.
  Based on the predicted pointmaps and confidence maps, we explicitly recover camera intrinsics, relative pose, and 2D feature correspondences using the procedures described in~\cite{leroy2024grounding}, which serve as inputs to metric pose estimation in Sec.~\ref{sec:refinement}.}
% While the technical details of MASt3R can be found in \cite{leroy2024grounding}, our focus is on its integration within the collaborative localization.}

\subsection{Hierarchical Multi-Session Mapping}
\label{sec:collaborative_mapping}

\bt{
This section mainly describes the methodology in aligning several individual submaps into a single, globally consistent reference frame, enabling the multi-session map merging.
Devices upload local query maps ($\mathcal{M}^{Q}$) to a server, which iteratively integrates each into the reference map ($\mathcal{M}^{R}$), yielding a progressively unified map ($\mathcal{M}^{R'}$).
Within a submap, the odometry graph encodes relative pose constraints $(\mathbf{T}^{i}_{j}, \mathbf{\Sigma}^{i}_{j})$ between consecutive nodes; merging two submaps additionally requires inter-submap loop-closure factors, each a relative constraint $(\mathbf{T}^{R_{i}}_{Q_{j}}, \mathbf{\Sigma}^{R_{i}}_{Q_{j}})$ between a reference node $\mathbf{n}^{R}_{i}$ and a query node $\mathbf{n}^{Q}_{j}$.
We estimate these factors in two steps (Fig.~\ref{fig:pipeline_hierarchical_localization}):
topological localization identifies geographically proximate and visually similar candidate pairs, and metric localization estimates the relative transformation $(\mathbf{T}^{R_{i}}_{Q_{j}}, \mathbf{\Sigma}^{R_{i}}_{Q_{j}})$ for each.}

% \begin{figure}[t]
%   \begin{center}
%     \includegraphics[width=0.95\linewidth]{Imgs/explanation_vpr-crop}
%   \end{center}
%   \caption{Three common types of trajectories in collaborative mapping and difference matrix with valid loops for the trajectory $3$}
%   \label{fig:traj_crowdsourcing}
% \end{figure}

\begin{figure}
  \centering
  \subfigure[Overlapping Cases]
  {\label{fig:explanation_vpr_traj}\centering\includegraphics[width=0.16\textwidth]{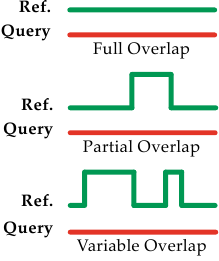}}
  \hspace{0.1cm}
  \subfigure[Difference Matrix with Loops]
  {\label{fig:explanation_vpr_d_matrix}\centering\includegraphics[width=0.31\textwidth]{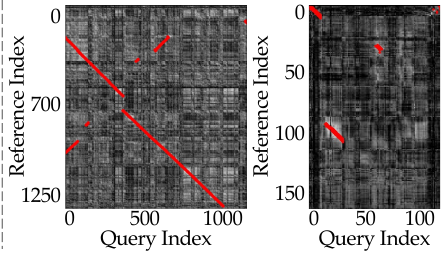}}
  \vspace{-0.3cm}
  \caption{\bt{Trajectory overlap patterns in multi-session mapping. (a) Three common cases: full overlap, partial overlap, and variable overlap. (b) Pairwise difference matrices with detected loop closures (red dots): the Oxford RobotCar dataset~\cite{maddern20171} shows dense, regular overlaps, while our dataset shown in Sec.~\ref{sec:exp_datasets} exhibits sparse, irregular overlaps, which motivates the proposed flexible sequence-based loop closure approach.}}
  \label{fig:explanation_vpr}
  \vspace{-0.3cm}
\end{figure}

\subsubsection{Topological Localization}
\label{sec:topo_localization}
\bt{
  Vision-based topological localization grows increasingly challenging as spatio-temporal scale expands: larger environments widen the search space and amplify perceptual aliasing, while temporal gaps introduce appearance variations that heighten ambiguity.
  A robust approach must achieve high precision (rejecting false matches) and high recall (retaining true ones).
  To this end, it combines global descriptor matching, sequence-based matching for temporal consistency, and GV for false-positive rejection.}

\bt{
  First, we extract a discriminative global descriptor (\ie a high-dimensional vector) for each image using a pre-trained VPR model \cite{yin2025general}.
  Under the assumption that geographically proximate images are visually similar, the cosine similarity between descriptors serves as a place-match score.
  However, single-frame descriptor matching is fragile: limited descriptor discriminability, appearance changes, and perceptual aliasing all contribute to false positives.
  We address this through sequence-based matching, which accumulates per-frame descriptor similarity scores along a trajectory to derive a more robust overall match score.
  SeqSLAM~\cite{milford2012seqslam} assumes continuous, heavily overlapping trajectories with near one-to-one alignments, which breaks down where trajectories span full, partial, or non-contiguous overlap (Fig.~\ref{fig:explanation_vpr}); forcing a single continuous path then causes missed or spurious matches.}

  \bt{We formulate sequence-based matching as a shortest-path search in a difference matrix $\mathbf{Diff}\in\mathbb{R}^{N\times M}$, where $\mathbf{Diff}(i,j)$ is the cosine distance between reference descriptor $\mathbf{d}_{R_i}$ and query descriptor $\mathbf{d}_{Q_j}$.
  We propose a DP-based method that supports two path-expansion operations ($[i,j]\rightarrow[i+k,j+1]$): \textit{1) in-sequence matching} ($k=V$) for continuous segments, and \textit{2) out-of-sequence jumping} ($|k| \geq \Delta$) to bridge trajectory gaps.
  An accumulated cost matrix $\mathbf{C}(i,j)$ is updated via:
  \begin{equation}
    \label{equ:dp_sequential_matching}
    \begin{aligned}
      \mathbf{C}(i, j) \leftarrow \min[\,  & \mathbf{C}(i, j),                            \\
                                           & \mathbf{C}(i-V, j-1) + \mathbf{Diff}(i, j),  \\
      \min_{|k-i|\ge\Delta}\,              & \mathbf{C}(k, j-1) + \mathbf{Diff}(i, j) + \lambda],
    \end{aligned}
  \end{equation}
  where $\lambda$ is a penalty for jump operations.
  Alg.~\ref{alg:dp_sequential_matching} summarizes this procedure. The optimal path is recovered by backtracking from the minimum-cost entry in the final column $\mathbf{C}(:,M)$, yielding a set of matched node pairs $\mathcal{P}_{\mathrm{SM}}=\{(\mathbf{n}^{R}_{i}, \mathbf{n}^{Q}_{j})\}$. 
  Since sequence matching does not guarantee per-pair precision, we implement a GV module~\cite{yu2024gv} to reject false positives. GV validates each pair by running RANSAC to estimate a fundamental matrix from keypoint correspondences; a pair is accepted into the final set $\mathcal{P}_{\mathrm{GV}}$ only if the inlier count exceeds a threshold.}

\begin{algorithm}[t]
    \caption{DP-based Sequential Matching}
    \label{alg:dp_sequential_matching}
    \footnotesize
    \begin{algorithmic}[1]
        \State \textbf{Input:} Difference matrix $\mathbf{Diff}\in\mathbb{R}^{N\times M}$
        \State \hspace{3.0em}Velocity set $\mathbb{V}=\{V_{min}, \cdots, V_{max}\}$
        \State \hspace{3.0em}Jump threshold $\Delta$; Penalty $\lambda$
        \State \textbf{Output:} Accumulated cost matrix $\mathbf{C}\in\mathbb{R}^{N\times M}$
        \State Initialize $\mathbf{C}(i,j)\gets\infty,\ \forall i,j$; \Comment{\tb{$i$: reference index; $j$: query index}}
        \State Set $\mathbf{C}(i,1)=\mathbf{Diff}(i,1)$
        \State Set $\mathcal{Q}\gets\{(i,1)\mid i=1,\dots,N\}$
        \For{$j = 2$ to $M$}
        \State $\mathcal{Q}'\gets\emptyset$
        \For{each $(i, V)\in \mathcal{Q}$}
        \State $i' \gets i + V$ \Comment{\tb{In-sequence matching}}
        \If{$i'\leq N$ and $\mathbf{C}(i,j-1)+\mathbf{Diff}(i',j)<\mathbf{C}(i',j)$}
        \State $\mathbf{C}(i',j) \gets \mathbf{C}(i,j-1)+\mathbf{Diff}(i',j)$; \ $\mathcal{Q}'\gets\mathcal{Q}'\cup\{(i',V)\}$
        \EndIf
        \For{$k\in [1,i-\Delta]\cup[i+\Delta,N]$} \Comment{\tb{Out-of-sequence jumping}}
        \State $Cost\gets\mathbf{C}(i,j-1)+\mathbf{Diff}(k,j)+\lambda$
        \If{$Cost<\mathbf{C}(k,j)$}
        \State $\mathbf{C}(k,j)\gets Cost$; \ $\mathcal{Q}'\gets\mathcal{Q}'\cup\{(k,V')\mid V'\in\mathbb{V}\}$
        \EndIf
        \EndFor
        \EndFor
        \State $\mathcal{Q}\gets\mathcal{Q}'$
        \EndFor
    \end{algorithmic}
\end{algorithm}

\subsubsection{Metric Localization}
\label{sec:refinement}

\bt{Given a verified node pair in $\mathcal{P}_{\mathrm{GV}}$, we estimate the 6-DoF relative transformation $(\mathbf{T}^{R_i}_{Q_j}, \mathbf{\Sigma}^{R_i}_{Q_j})$.
The sparse, landmark-free topometric map poses two challenges:
\ti{a)} sparse nodes often have limited visual overlap, and
\ti{b)} the absence of pre-built 3D structure, which typically provides a strong prior for pose estimation~\cite{sarlin2019coarse}.
To address both, we leverage MASt3R~\cite{leroy2024grounding} to perform on-demand dense local reconstruction.
For a given pair $(\mathbf{n}^{R}_{C_{i}}, \mathbf{n}^{Q}_{C_{j}})$, we retrieve additional reference nodes from $\mathcal{G}^{R}_{C}$ with high covisibility to $\mathbf{n}^{R}_{C_{i}}$, then jointly optimize local scene geometry and all camera poses using predicted pointmaps, poses, and scale factors as initialization.
The optimization further incorporates confidence map calibration to suppress outliers.
Accuracy improves with more reference nodes at the cost of additional computation; as shown in Sec.~\ref{sec:exp_metric_level_localization}, even with few references our method outperforms prevalent SfM-based approaches.}

\begin{figure}[t]
  \begin{center}
    \includegraphics[width=0.93\linewidth]{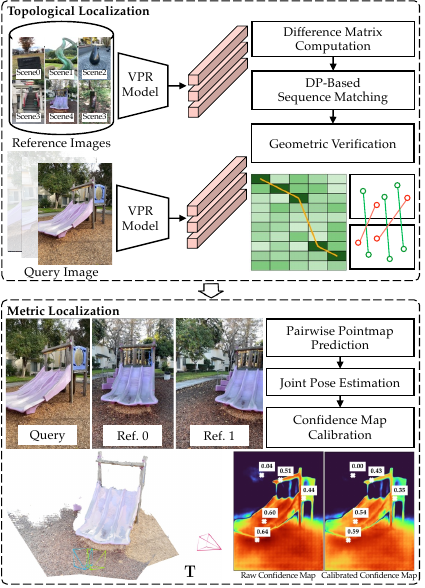}
  \end{center}
  \vspace{-0.3cm}
  \caption{\bt{Pipeline of the first two stages of the hierarchical visual localization. (a) Topological localization matches query and reference descriptors via DP-based sequence matching and geometric verification. (b) Metric localization predicts pairwise pointmaps and jointly optimizes camera poses and scene geometry; confidence map calibration suppresses low-confidence regions (\eg uncovered by references) more strongly ($0.44\rightarrow0.35$) than high-confidence ones ($0.64\rightarrow0.59$), yielding the $6$-DoF loop-closure factor for PGO.}}
  \label{fig:pipeline_hierarchical_localization}
  \vspace{-0.3cm}
\end{figure}

\paragraph{\bt{Joint Pose Estimation}}
constructs a local connectivity graph $\mathcal{G}(\mathcal{V}, \mathcal{E})$ over a small keyframe set $\mathcal{I}$ and jointly optimizes all involved camera poses and local scene geometry for multi-view consistency, taking MASt3R's feed-forward predictions as input.
  Note that this optimization is strictly local: $\mathcal{I}=\{\mathbf{I}^{R}_1, \dots, \mathbf{I}^{R}_{N-1}, \mathbf{I}^{Q}\}$ contains only $N-1$ reference keyframes retrieved near the query and the query image itself, with $N$ typically set to 2--3.
  Pairwise predictions $f_{\bm{\theta}}(\mathbf{I}_i, \mathbf{I}_j)$ are computed for all pairs in $\mathcal{I} \times \mathcal{I}$, forming $\mathcal{G}$ where nodes $\mathcal{V}$ are images and edges $\mathcal{E}$ connect pairs with sufficient visual overlap measured by the confidence map.
  All frames in $\mathcal{I}$ are then jointly optimized for consistent local pose and geometry estimates.
  Initialization proceeds via the maximum-spanning tree of $\mathcal{G}$ weighted by confidence, yielding initial focal lengths $\{f_{v}\}_{v=1}^{N}$, scale factors $\{\sigma^{(i,j)}\in \mathbb{R}^{+}\}$, and poses $\{\mathbf{T}^{(i,j)} \in SE(3)\}$.
  Reference poses $\{\mathbf{T}^{R}_i\}$ are held fixed; the query pose $\mathbf{T}^{Q}$ and remaining parameters are optimized by minimizing the 3D-3D alignment error between a unified local pointmap $\bm{\chi}$ and the transformed pairwise pointmaps $\mathbf{X}$:
  \begin{equation}
    \begin{aligned}
      \mathbf{T}^{Q*}, \{\bm{\chi}, \sigma\}^{*} =
      \underset{\mathbf{T}^{Q}, \{\bm{\chi}, \sigma\}}{\arg\min}
       & \sum_{(i,j)\in\mathcal{E}}
      \sum_{v\in (i,j)}
      \sum_{p=1}^{hw}               \\
       & \mathbf{C}^{v,i}_{p}
      \| \bm{\chi}^{v}_{p} - \sigma^{(i,j)} \mathbf{T}^{(i,j)} \mathbf{X}^{v,i}_{p} \|,
    \end{aligned}
    \label{equ:global_alignment}
  \end{equation}
  where $\mathbf{C}^{v,i}_{p}$ is the confidence weight for point $p$ of view $v$ in pair $(i,j)$.
  The local pointmap $\bm{\chi}$ is re-parameterized via depth back-projection:
  \begin{equation}
    \bm{\chi}^{v}_{p}
    =
    \mathbf{T}_{v}\mathbf{K}_{v}^{-1}\mathbf{D}_{v,p}[u_{p}, v_{p}, 1]^{\top}
    =
    \mathbf{T}_{v}\frac{\mathbf{D}_{v,p}}{f_{v}}[u_{p}', v_{p}', 1]^{\top},
  \end{equation}
  where $\mathbf{K}_{v}$ and $\mathbf{T}_{v}$ are the intrinsics and extrinsics of view $v$, $\mathbf{D}_{v,p}$ is the depth at pixel $p$, and $(u_p', v_p')$ are principal-point-normalized coordinates.
  Substituting into Eq.~\eqref{equ:global_alignment} yields the expanded objective:
  \begin{equation}
    \label{equ:global_alignment_expand}
    \begin{aligned}
       & \mathbf{T}^{Q*}, \{\mathbf{D}, f, \sigma\}^{*}
      =
      \underset{\mathbf{T}^{Q}, \{\mathbf{D}, f, \sigma\}}{\arg\min}
      \sum_{(i,j)}
      \sum_{v}
      \sum_{p}
      \\
       & \ \ \ \ \ \ \ \ \ \ \ \mathbf{C}^{v,i}_{p}
      \| \mathbf{T}_{v}\frac{\mathbf{D}_{v,p}}{f_{v}}[u_{p}', v_{p}', 1]^{\top} - \sigma^{(i,j)} \mathbf{T}^{(i,j)} \mathbf{X}^{v,i}_{p} \|.
    \end{aligned}
  \end{equation}

\bt{
  The pairwise pose $\mathbf{T}^{(i,j)}$ and the per-view pose $\mathbf{T}_{v}$ are parameterized independently for greater optimization flexibility.
  Problem \eqref{equ:global_alignment_expand} is solved via several hundred gradient-descent iterations and converges within seconds on a standard GPU.
  As a byproduct, scale-aware pointmaps and camera intrinsics are recovered, enabling consistent local scene reconstruction.
  Because the joint optimization couples pointmap, pose, and scale estimation into a single objective, the quality of the recovered reconstruction directly reflects the reliability of the estimated query pose, where we exploit this property in the subsequent confidence map calibration.}

\paragraph{\bt{Confidence Map Calibration}}
addresses the overconfidence of MASt3R's feed-forward predictions~\cite{lu2025lora3d}, which assign high weights to poorly predicted pointmaps and thereby bias the optimization in Eq.~\eqref{equ:global_alignment_expand}.
  We address this by calibrating the confidence maps at each iteration, dynamically coupling each weight to the actual per-point residual via the Geman-McClure robust kernel.
  This reformulates the objective as an iteratively re-weighted least-squares (IRLS) problem:
\begin{equation}
  \label{equ:global_alignment_irls}
  \begin{aligned}
    \mathbf{T}^{Q*}, \{\mathbf{D}, f, \sigma\}^{*}
     & =
    \underset{\mathbf{T}^{Q}, \{\mathbf{D}, f, \sigma\}}{\arg\min}
    \sum_{(i,j)}
    \sum_{v}
    \sum_{p}
    \mathbf{W}^{v,i}_{p}\|\mathbf{e}^{v,i}_{p}\|, \\
    \mathbf{W}^{v,i}_{p}
     & =
    \frac{\mathbf{C}^{v,i}_{p}}{(1 + \|\mathbf{e}^{v,i}_{p}\|/\mu)^{2}},
  \end{aligned}
\end{equation}
\bt{where $\mu$ controls outlier rejection (smaller values yield stronger suppression), and the per-point residual is
$\|\mathbf{e}^{v,i}_{p}\| = \|\mathbf{T}_{v}\frac{\mathbf{D}_{v,p}}{f_{v}}[u_{p}', v_{p}', 1]^{\top} - \sigma^{(i,j)} \mathbf{T}^{(i,j)} \mathbf{X}^{v,i}_{p}\|$.
Since pose estimation quality is inherently coupled with reconstruction quality, the calibrated confidence map (CCM) at the final iteration serves as a reliable indicator of pose accuracy (Fig.~\ref{fig:pipeline_hierarchical_localization}).
We therefore define an approximate pose covariance as $\bm{\Sigma}^{R_i}_{Q_j}=\diag[(\overline{\mathbf{W}}_{i,i}\cdot\overline{\mathbf{W}}_{i,j})^{-2}]$, where $\overline{\mathbf{W}}_{i,i}$ and $\overline{\mathbf{W}}_{i,j}$ are the mean CCM weights for the reference-reference and reference-query image pairs, respectively.
In Sec.~\ref{sec:optimization}, this covariance is used to select high-confidence node pairs $\mathcal{P}_{\mathrm{CCM}}$ as loop-closure factors, filtering outliers prior to PGO.}

\subsubsection{Pose Graph Optimization}
\label{sec:optimization}
\bt{The high-confidence loop-closure factors from $\mathcal{P}_{\mathrm{CCM}}$ link the query and reference submaps.
A unified pose graph is then constructed and a global PGO problem is solved, jointly optimizing all submap poses in the world frame $W$ and cumulative odometry drift via inter-submap loop constraints, ensuring geometric consistency of the merged map.
The PGO minimizes a robust nonlinear least-squares objective over all poses $\mathcal{T}= \{\mathbf{T}^{R}_i\} \cup \{\mathbf{T}^{Q}_j\}$:
\begin{equation}
  \label{equ:pgo_formulation}
  {\mathcal{T}}^{*} =  \underset{\mathcal{T}}{\arg\min}
  \sum_{(i,j) \in \mathcal{E}}
  \rho\left(
  \|\log[(\bar{\mathbf{T}}^{i}_{j})^{-1}(\mathbf{T}_{i})^{-1}\mathbf{T}_{j}]\|_{\bm{\Sigma}^{i}_{j}}^2
  \right),
\end{equation}
where $\mathcal{E}$ includes odometry edges and inter-submap loop-closure edges;
$\bar{\mathbf{T}}^{i}_{j}$ is the measured relative transformation between nodes $i$ and $j$ with covariance $\bm{\Sigma}^{i}_{j}$;
$\rho(\cdot)$ is a robust kernel (\eg Huber);
$\log(\cdot): SE(3) \to \mathfrak{se}(3)$ is the Lie algebra logarithm; and
$\|\mathbf{x}\|_{\bm{\Sigma}}^2 = \mathbf{x}^\top \bm{\Sigma}^{-1}\mathbf{x}$ is the squared Mahalanobis distance.
This is solved using GTSAM~\cite{frank2022gtsam} with the Levenberg-Marquardt solver, yielding optimized poses $\mathcal{T}^{*}$ that integrate the query submap into the reference map.}

\subsection{Cross-Device Localization}
\label{sec:crowdsourcing}

\bt{A key advantage of the topometric map is that its nodes are raw images rather than 3D landmarks, which makes the representation inherently device-agnostic: any camera that produces an RGB image can contribute a node.
This section describes how we realize this potential in practice, extending multi-session map merging to data captured by heterogeneous devices under uncontrolled, in-the-wild conditions.
First, \ti{device heterogeneity}: consumer cameras span a wide range of optics, from fisheye lenses on AR glasses to equirectangular panoramas from $360^\circ$ cameras, none of which are directly compatible with the perspective-image assumption of feed-forward 3D GFMs (\eg MASt3R~\cite{leroy2024grounding}).
Second, \ti{perceptual degradation}: non-expert users produce imagery with motion blur, inconsistent exposure, and sensor noise that, if left unfiltered, corrupt both descriptor matching and metric pose estimation.
We address each in turn.}

\begin{figure}[t]
    \centering
    \subfigure[Equirectangular to perspective projection]{
        \includegraphics[width=0.99\linewidth]{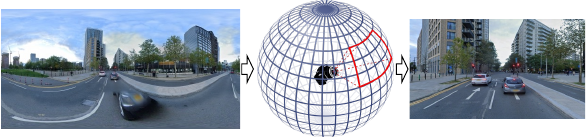}
        \label{fig:pano_to_pers}
    }
    % \hspace{-0.5cm}
    \subfigure[Images with their IQA scores. The left pair are low-quality due to motion blur and low illumination, respectively]{
        \includegraphics[width=0.99\linewidth]{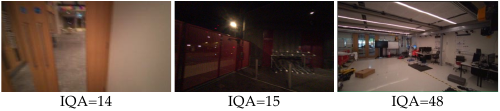}
        \label{fig:image_quality}
    }
    \caption{\bt{Cross-device localization strategies. (a) A panoramic equirectangular image from Google street view is reprojected into a synthetic perspective view with user-defined FoV and intrinsics. (b) Image quality assessment (IQA) scores identify low-quality frames for filtering prior to map construction.}}
    \label{fig:cross_device_strategy}
    \vspace{-0.4cm}
\end{figure}

\bt{\tb{\ti{Perspective Rectification}} converts non-perspective inputs into undistorted perspective images compatible with our metric localization pipeline.
    For consumer devices (\eg smartphones, AR glasses), we apply an undistortion module that uses manufacturer-provided intrinsics to rectify raw images into distortion-free views.
    For panoramic sources such as Google Street View, images are stored as equirectangular projections and cannot be used directly; we convert them by sampling a virtual pinhole view via spherical projection, typically centered along the travel direction, yielding perspective images with a user-defined FoV and known virtual intrinsics.
    As shown in Fig.~\ref{fig:pano_to_pers}, this conversion produces inputs that are fully compatible with our pipeline.
    All images are then cropped to a uniform resolution matching the reference images.}

\bt{\tb{\ti{Image Quality Filtering}} suppresses perceptually degraded nodes that would otherwise corrupt descriptor matching or metric pose estimation.
    Crowdsourced imagery exhibits variable quality from motion blur, occlusions, and sensor noise, degrading localization reliability.
    We apply MUSIQ~\cite{ke2021musiq}, a non-reference IQA method that predicts a quality score $q\in[0, 100]$ per image; those below a minimum threshold are filtered before submap construction, and the same score $q$ is reused as the image quality factor in node culling, where it contributes to the information contribution score that governs long-term map maintenance.
    Fig.~\ref{fig:image_quality} shows representative IQA scores under day and night conditions, and Fig.~\ref{fig:exp_datasets} visualizes the full score distribution across our dataset.}
\subsection{Lifelong Map Maintenance}
\label{sec:method_data_pruning}

\bt{As submaps are continuously merged, the map grows in both spatial coverage and node count.
While our topometric map is inherently more compact than dense metric or landmark-based alternatives, unconstrained growth still raises storage and retrieval costs, and nodes from dynamically changed regions degrade localization over time.
SLAM systems typically bound map size by keyframe selection on geometric criteria such as spatial overlap, parallax~\cite{qin2018vins}, or feature-tracking stability~\cite{mur2015orb}, which neglect two factors critical for lifelong operation: perceptual quality (\eg motion blur) and temporal relevance (\eg dynamically changed regions).
Our key insight is that a node's utility is not fixed at insertion but evolves as the environment changes and new observations arrive.
We therefore introduce a probabilistic node culling strategy that evaluates each node's information contribution (IC) from three complementary factors: image quality, information gain relative to neighboring nodes, and temporal difference.
Nodes with low IC are culled, keeping the map compact and up-to-date without sacrificing coverage.}

\subsubsection{Node Culling}
\bt{The culling procedure operates after each PGO step and consists of forward and backward passes applied exclusively to the high-confidence metric-localization pairs $\mathcal{P}_{\mathrm{CCM}}$.
The IC of each node is evaluated from three factors:
  \begin{itemize}[leftmargin=0.7cm]
    \item \ti{Image Quality (IQ):} captures perceptual reliability via the IQA score $q$. Nodes with low $q$ (\eg due to motion blur or poor illumination) contribute unreliable observations.
    \item \ti{Temporal Difference (TD):} captures temporal relevance via the elapsed time $\Delta\tau = \tau_{Q} - \tau_{R}$ between a query node and its matched reference node. Larger $\Delta\tau$ indicates greater risk of appearance or structural mismatch due to environmental dynamics.
    \item \ti{Information Gain (IG):} captures geometric novelty of a query node $\mathbf{n}^{Q}$ relative to a reference node $\mathbf{n}^{R}$. The pointmap $\mathbf{X}^{R}_{Q}$ predicted by the feed-forward 3D GFM is projected onto $\mathbf{I}_{R}$; the IG score is $g = n / (wh)$, where $n$ is the number of projected points falling outside the image bounds of $\mathbf{I}_{R}$, and $w, h$ are its dimensions.
  \end{itemize}
}

\bt{
As illustrated in Fig.~\ref{fig:data_pruning} (left), the culling procedure distinguishes between two node roles.
The forward pass evaluates the incoming query node $\mathbf{n}^{Q}$, assessing whether it provides sufficient new information to justify retention.
The backward pass re-evaluates existing reference nodes $\{\mathbf{n}^{R}_{i}\}$ connected to $\mathbf{n}^{Q}$, checking whether they remain informative given the newly merged observations.
To minimize computation, IC is calculated only for nodes within $\mathcal{P}_{\mathrm{CCM}}$, and culling is triggered immediately after each submap merge.
The forward culling probability for $\mathbf{n}^{Q}$ and the backward culling probability for each reference node $\mathbf{n}^{R}_{i}$ are:}
\begin{align}
  P_{FW}(\mathbf{n}^{Q})     &= P_{IQ}(q_{Q}) \cdot P_{IG}(g_{Q}) \cdot P_{TD}(\tau_{Q} - \tau_{R}), \label{equ:forward_operation} \\
  P_{BW}(\mathbf{n}^{R}_{i}) &= P_{IG}(g_{R_{i}}) \cdot P_{TD}(\tau_{R_{i}} - \tau_{Q}),              \label{equ:backward_operation}
\end{align}
\bt{where $P_{IQ}(\cdot)$ and $P_{IG}(\cdot)$ are sigmoid functions mapping their respective scores to $[0,1]$, and $P_{TD}(\cdot)$ is an exponential decay function.
Nodes whose culling probability falls below a predefined threshold are removed from the map.
Fig.~\ref{fig:data_pruning} (right) illustrates a representative scenario in which a node is culled or replaced upon the arrival of a higher-IC candidate.}

\begin{figure}[t]
  \begin{center}
    \includegraphics[width=0.92\linewidth]{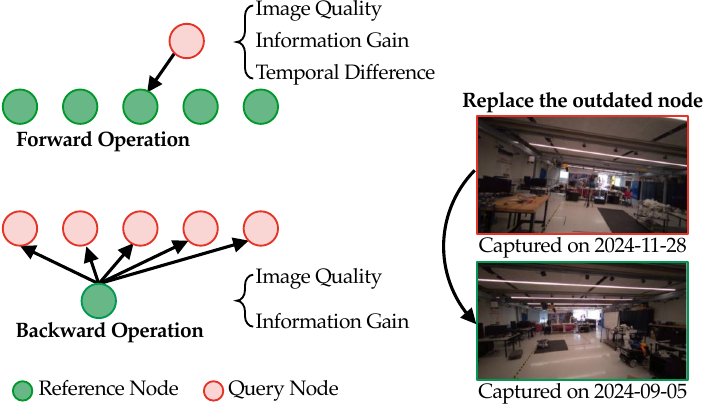}
  \end{center}
  \vspace{-0.3cm}
  \caption{\bt{Node culling for lifelong map maintenance. Each node in $\mathcal{P}_{\mathrm{CCM}}$ is scored by its information contribution (IC), combining image quality, information gain, and temporal recency. Forward and backward passes compare adjacent nodes: the tail node is culled if its IC falls below the threshold (left). A real-world example shows an outdated node (2024-09-05) replaced by a higher-IC node from a later session (2024-11-28) (right).}}
  \label{fig:data_pruning}
  \vspace{-0.4cm}
\end{figure}

\subsubsection{Edge Updating}
\label{sec:method_connectivity_augmentation}
\bt{Submap construction in Sec. \ref{sec:submap_construction} produces edges only between temporally consecutive nodes, yielding a sparse chain topology that fails to capture the true spatial connectivity of the environment.
Nodes that are geographically proximate but temporally distant, such as those at revisited locations, lack direct edges, which limits both localization recall and path planning efficiency.
We therefore augment the covisibility graph by scanning each node's spatial neighbors and adding edges between pairs whose covisibility exceeds a minimum keypoint-match threshold, applying the same scan to the traversability graph.
Performed during per-session submap construction, this adds shortcuts between geographically close but temporally distant nodes, improving global path planning via the shortest-path algorithm.
Whenever map updates occur, such as node culling or submap merging, edges in the affected regions are re-evaluated to keep connectivity consistent with the current node set.
More principled traversability estimation methods~\cite{shah2023gnm} could further refine edge quality, which we leave as future work.}

% For each node, we identify spatially adjacent nodes within a distance threshold and add a traversability edge between each pair whose Euclidean distance $d_{ij} = \|\mathbf{t}^W_i - \mathbf{t}^W_j\|$ falls below a threshold $d_{\max}$.
% Each added edge is assigned a traversability cost $v_{T_{ij}} = d_{ij}$, encoding physical path length as the default motion-feasibility proxy.

\section{Leveraging \methodname for Visual Navigation}
\label{sec:vnav_system}

\bt{
Because the multi-layer topometric map of Sec.~\ref{sec:methodology} stores an image at every node, it is directly actionable for image-goal visual navigation; we therefore build a navigation system on \methodname using existing open-source frameworks \cite{jiao2025litevloc,zhang2020falco}.
The system inherits the hierarchical structure of the above multi-session localization: topological localization queries the covisibility graph for efficient candidate retrieval, after which metric localization computes the 6-DoF query pose relative to the retrieved reference node.
This deployment validates that the topometric representation alone suffices for autonomous navigation, requiring no additional map augmentation.}

\subsection{Hierarchical Global Visual Localization}
\label{sec:visual_localization}

\bt{
The robot estimates its global pose relative to the map through a two-stage hierarchical pipeline.
The multi-session metric localization in Sec.~\ref{sec:refinement} performs offline multi-view joint optimization and is not suitable for real-time operation.
For online navigation, we instead adopt a lightweight alternative: feature matching via MASt3R~\cite{leroy2024grounding} followed by PnP pose solving~\cite{gao2003complete}.
Since the topometric map stores no pre-built 3D structure, establishing 2D-3D correspondences for PnP requires an online depth source; we therefore target platforms equipped with RGB-D sensors or stereo cameras, which are widely available on modern navigation robots.
This reflects a design trade-off: lighter offline storage at the cost of requiring an online depth source.}

\begin{figure}[t]
  \begin{center}
    \includegraphics[width=0.97\linewidth]{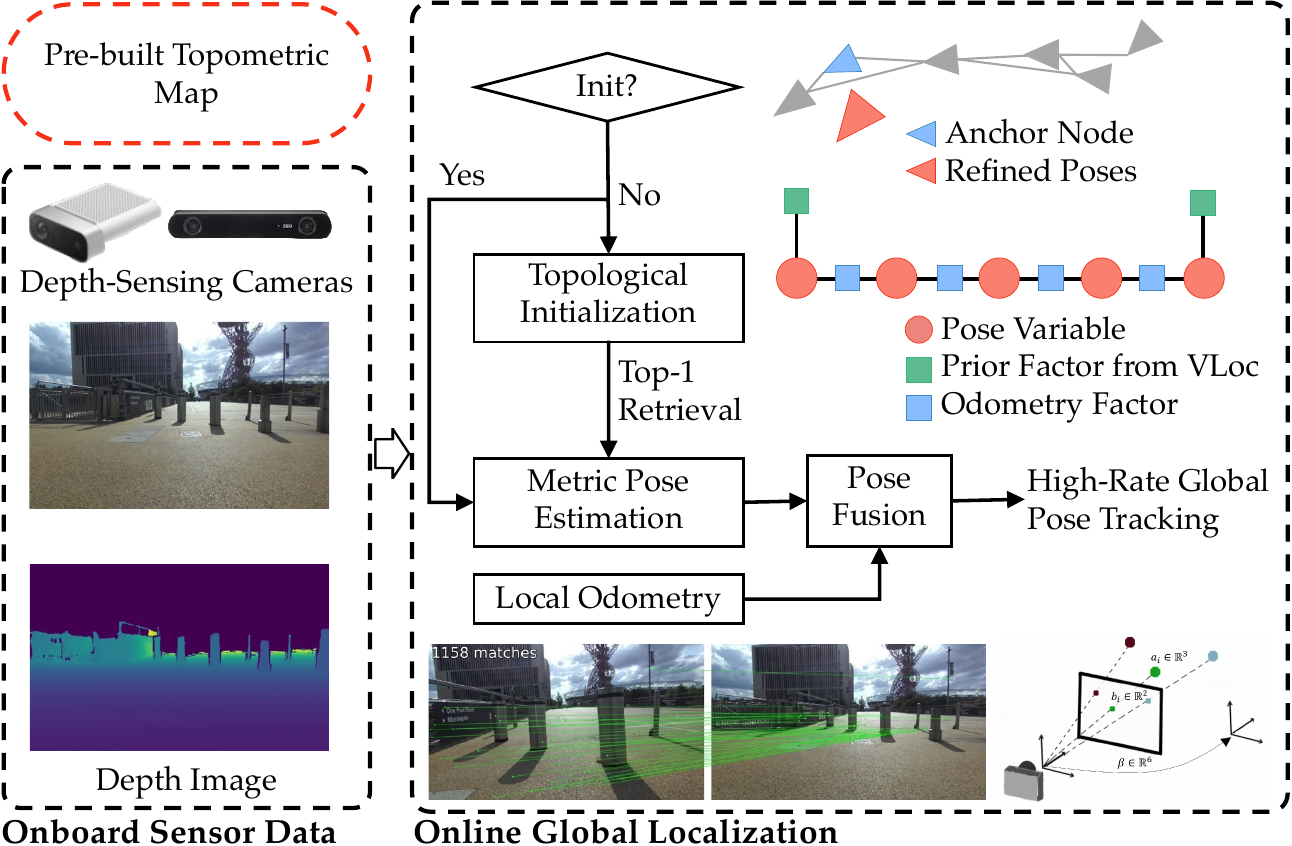}
  \end{center}
  \vspace{-0.3cm}
  \caption{\bt{Hierarchical localization pipeline built on the \methodname topometric map. Topological initialization identifies the anchor node via VPR. Metric pose estimation computes the $6$-DoF query pose using feature matching and PnP. Pose fusion combines absolute localization estimates with local odometry via PGO to produce high-rate, drift-corrected poses for image-goal navigation.}}
  \label{fig:pipeline_vloc}
  \vspace{-0.3cm}
\end{figure}

\subsubsection{Topological Initialization}
\label{sec:topo_initialization}
\bt{
Topological initialization is required at mission start, after pose tracking is lost due to insufficient feature correspondences, or following a system restart.
As the map can span large-scale environments with thousands of nodes, and external global localization sources such as GNSS is not always available, the robot must determine its initial position purely from visual appearance.
It does so by performing VPR: a global descriptor of the current observation is compared against all map nodes to identify the most visually similar candidates.
The top-$K$ candidates are retrieved by selecting the map nodes with the highest descriptor similarity scores, and a geometric verification step re-ranks them to identify the single most confident match, which is designated as the \ti{anchor node}.}

\subsubsection{Metric Localization}
\label{sec:loc_metric_localization}
\bt{Given the anchor node, the robot retrieves spatially local nodes from the covisibility graph and selects the most visually similar one to the current query image $\mathbf{I}^Q$ as the reference frame $\mathbf{n}^R$.
Feature correspondences between $\mathbf{I}^Q$ and $\mathbf{I}^{R}$ are established, and each matched query pixel is back-projected into 3D using the depth measurement from the RGB-D sensor, yielding 3D-2D correspondences between the query frame and the reference image.
The relative pose $\mathbf{T}^{R}_{Q}$ is then recovered via PnP~\cite{gao2003complete} with RANSAC~\cite{fischler1981random}, and the global query pose follows as $\mathbf{T}^W_Q = \mathbf{T}^W_{R} \mathbf{T}^{R}_{Q}$, where $\mathbf{T}^W_{R}$ is the reference node pose stored in the map.}

\subsubsection{Pose Fusion}
\label{sec:pose_fusion}
\bt{The metric localization above is susceptible to domain gaps, viewpoint variations, texture-less scenes, depth noise, and processing latency.
We formulate pose fusion as a PGO problem analogous to Eq.~\eqref{equ:pgo_formulation}, combining absolute pose priors for global anchoring with relative motion constraints from local odometry (\eg wheel encoders) to suppress drift; the PGO triggers on each new estimate, with poses propagated by odometry between updates.}

%%%%%%%%%%%%%%%%%%%%%%%%%%%%%%%%%%%%%
\subsection{Closed-Loop Visual Navigation}
\label{sec:navigation}

\bt{A key advantage of storing images at map nodes is that the navigation goal can be specified as any image in the map, without coordinate input or semantic labeling.
The robot identifies the goal node via the topological initialization and then plans a path through the traversability graph using Dijkstra's algorithm, yielding a waypoint sequence $P = \{\mathbf{n}_S, \dots, \mathbf{n}_i, \dots, \mathbf{n}_G\}$ whose edge costs encode physical traversability.
At each step, the robot advances toward the next waypoint $\mathbf{n}_{i+1}$ using a local planner for obstacle avoidance, while the global localization pipeline continuously anchors its pose to the map.
At runtime, the three layers play complementary roles: the covisibility graph supports localization, the odometry graph maintains global consistency, and the traversability graph supports planning, all within a single lightweight topometric map.}

\section{Experiments}
\label{sec:experiments}

\bt{
This section comprehensively evaluates \methodname, covering localization, multi-session map merging, cross-device mapping, node culling, and visual navigation.
We begin with Sec.~\ref{sec:exp_setup}, which describes the shared implementation, datasets, and metrics used throughout, and then organize the remaining experiments around the following questions:
\begin{itemize}
\item \textbf{Q1.} How well do the topological and metric stages of the proposed hierarchical visual localization pipeline perform (Sec.~\ref{sec:exp_localization})?
\item \textbf{Q2.} How accurately does \methodname merge multi-session submaps, and how does it compare to a multi-session SLAM baseline (Sec.~\ref{sec:exp_map_merging_homo})?
\item \textbf{Q3.} Can \methodname integrate data from heterogeneous devices into a globally consistent map (Sec.~\ref{sec:exp_colla_mapping})?
\item \textbf{Q4.} How does the node culling strategy affect map size and localization accuracy over time (Sec.~\ref{sec:exp_keyframe_selection})?
\item \textbf{Q5.} \bt{Does multi-session merging extend and update the navigable space and shorten planned paths, and can the resulting topometric map thereby support reliable image-goal visual navigation (Sec.~\ref{sec:experiment_vnav})?}
\end{itemize}}

\subsection{\bt{Experimental Setup}}
\label{sec:exp_setup}

\subsubsection{\bt{Implementation Details}}
\bt{
  The proposed multi-session mapping is implemented in Python. All learning-based models are used with their publicly released pre-trained weights without additional fine-tuning. We next describe the submap construction, the place recognition and navigation modules, and the datasets and metrics shared across experiments.}

\paragraph{\bt{Submap Construction Settings}}
\label{sec:exp_submap_construction}
\bt{
  Following Sec.~\ref{sec:submap_construction}, each device builds an individual submap from keyframes selected along its trajectory.
  A new keyframe is triggered when the displacement from the previous one exceeds a translational threshold of $3.9$m or a rotational threshold of $60^{\circ}$.
  This threshold stays small enough for keyframes to remain spatially dense, so a query image can retrieve a nearby anchor with sufficient overlap for metric localization, yet large enough to limit redundancy and keep the map compact.
  % If consecutive keyframes are too far apart, a query image may fail to find a good database candidate and yield poor metric pose estimates.
  We further assume that consecutive keyframes are locally traversable, so traversability edges are added directly between them along the recorded trajectory.}

\begin{figure}[t]
  \centering
  \includegraphics[width=0.93\linewidth]{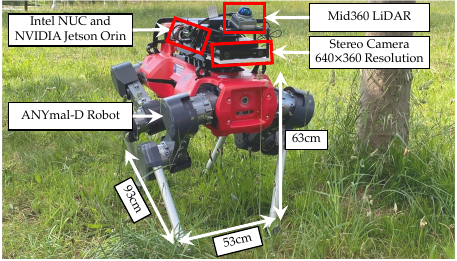}
  \vspace{-0.1cm}
  \caption{\bt{Hardware platform for real-world navigation experiments: an ANYmal-D legged robot equipped with onboard stereo cameras and an embedded computing unit, used to validate the full visual navigation pipeline in indoor and outdoor environments.}}
  \label{fig:exp_robot}
  \vspace{-0.33cm}
\end{figure}

\paragraph{\bt{Place Recognition Models}}
\bt{
  A unified VPR model serves both topological localization during multi-session map merging and global localization during navigation.
  We adopt the pre-trained ResNet-$18$ CosPlace \cite{berton2022rethinking}, which has $11$M parameters and produces a $256$-D descriptor.
  It offers the best accuracy-efficiency trade-off in our evaluation of Sec.~\ref{sec:exp_topo_localization}, thanks to its classification-based training scheme.
  The architecture is also compatible with stronger VPR models such as EigenPlaces \cite{berton2023eigenplaces}, AnyLoc \cite{keetha2023anyloc}, and MegaLoc \cite{berton2025megaloc} for higher recall when needed.}

\paragraph{\bt{Navigation System and Robots}}
\bt{
  Our visual navigation system is evaluated in both simulation and on a real robot.
  In simulation, we use a two-wheeled differential-drive robot in the photo-realistic Matterport3D environment \cite{chang2017matterport3d}, which provides $10$Hz RGB and depth inputs.
  Three scenes covering home and office settings, denoted Env$0$-$2$ in Tab.~\ref{tab:experiment_vnav_simu}, are selected for testing\footnote{Sourced from \url{https://aspis.cmpt.sfu.ca/scene-toolkit/scans/matterport3d/houses}. The scene IDs, ordered by increasing size, are 17DRP5sb8fy, EDJbREhghzL, and B6ByNegPMK.}.
  Local odometry comes from ICP on depth-image point clouds, fused with $1$Hz vision-based global pose estimates to output poses at the odometry rate.
  Local planning uses the learning-based iPlanner \cite{yang2023iplanner} from a single depth image.
  Simulation experiments run on a desktop PC with an Intel i$9$ CPU and an NVIDIA RTX $4090$ GPU.}

\bt{
  For real-world deployment, we use the ANYmal-D quadruped robot shown in Fig.~\ref{fig:exp_robot}.
  Local odometry is provided by a $20$Hz proprioceptive state estimator \cite{bloesch2017state} that fuses IMU, joint-encoder, and contact data, and a front-facing ZED$2$ stereo camera ($640\times360$) supplies RGB images and depth maps up to $15$m.
  Local planning uses the model-based Falco \cite{zhang2020falco}, operating on a local 3D map built on-the-fly from a single-frame point cloud.
  Computation is split across two onboard computers: an Intel NUC handles local planning and path tracking, while an NVIDIA Jetson Orin runs the global planner and the visual localization algorithm.}

\subsubsection{\bt{Datasets}}
\label{sec:exp_datasets}
\bt{
  We evaluate our system on a our self-collected dataset complemented by several public datasets.
  The data supports both localization and map merging across diverse scales and conditions, while the public datasets enable comparison with baseline methods and simulated cross-device merging.}

\paragraph{\bt{Self-Collected Dataset}}
\label{sec:exp_datasets_self_collected}
\bt{
  We construct our self-collected dataset using Meta Project Aria glasses \cite{engel2023project}, which integrate two wide-angle grayscale cameras, a front-facing RGB camera, a consumer-grade IMU, and GPS.
  Spanning two countries over $3.5$ months, the dataset comprises $35$ sequences and over $18.7$km of trajectories, organized into three geographic regions R$0$-R$2$ detailed in Tab.~\ref{tab:exp_regions}.
  Five users with varied behaviors contributed, introducing wide viewpoint variations (\eg forward- and backward-facing motion), appearance changes, and irregular trajectories.
  Ground truth (GT) poses are generated by Meta's cloud-based SLAM service \cite{engel2023project} in a global frame, whose city-scale accuracy has been independently validated \cite{krishnan2025benchmarking}.
  All images are anonymized by blurring identifiable information such as faces, and a dataset overview is shown in Fig.~\ref{fig:exp_datasets}.
  Following the Map-Free benchmark \cite{arnold2022map}, the sequences are post-processed to evaluate both the coarse-to-fine localization and the full map-merging pipeline.}

\begin{table}[t]
    \centering
    \caption{Geographic Regions in the Self-Collected Dataset.}
    \renewcommand\arraystretch{1.1}
    \renewcommand\tabcolsep{5.0pt}
    \footnotesize
    \begin{tabular}{cccc}
        \toprule[0.03cm]
        \textbf{Region} & \textbf{Scale} & \textbf{Environment}      & \textbf{Key Challenge}   \\
        \midrule[0.03cm]
        R$0$            & Small          & Vineyard                  & Repetitive structure \\
        R$1$            & Medium         & Campus                    & Day and night        \\
        \multirow{2}{*}{R$2$} & \multirow{2}{*}{Large} & Urban: buildings, parks, & Day and night           \\
                              &                        & roads, shopping centers  & Large viewpoint changes \\
        \bottomrule[0.03cm]
    \end{tabular}
    \label{tab:exp_regions}
\end{table}

\begin{itemize}[leftmargin=0.7cm]
  \item \tb{Topological Localization:} \bt{From R$2$ we manually select $4$ reference sequences and, for each, the spatially and visually overlapping query sequences, giving $26$ query sequences in total.}
  \item \tb{Metric Localization:} \bt{We randomly select $57$ query images per scene from R$2$. Each query is paired with reference images within $5$m that have verified visual overlap, measured by matched keypoints. Queries span varying perspectives and times of day, with GT relative transformations provided.}
  \item \tb{Map Merging:} \bt{We use all of R$0$-R$2$, splitting each sequence into segments of at most $300$m to simulate short-term multi-user capture, yielding $68$ segments. Camera intrinsics, extrinsics, timestamps, local VIO poses, and GT poses are provided.}
\end{itemize}

\paragraph{\bt{Public Datasets}}
\bt{
  Metric localization is additionally evaluated on Map-Free \cite{arnold2022map} with $65$ validation scenes and GZ-Campus \cite{jiao2025litevloc} with $30$ scenes.
  For qualitative cross-device localization and mapping, we augment R$1$ with three iPhone $16$ Pro sequences captured using the kit \cite{liu2024marvin} and one vehicle-mounted camera sequence from \cite{wei2024fusionportablev2}, and enrich R$2$ with Google Street View panoramas spanning $2012$-$2024$.
  We further use the $360$Loc dataset \cite{huang2024360loc} for quantitative map merging, leveraging its GT poses and simulating cross-device queries by projecting its panoramic images into perspective views with varying intrinsics and extrinsics.
  Further data-processing details and examples are provided in the \textbf{Suppl. Mat.} \cite{anonymous2025opennavmap_supp}.}

\begin{figure}[t]
  \centering
  \includegraphics[width=0.99\linewidth]{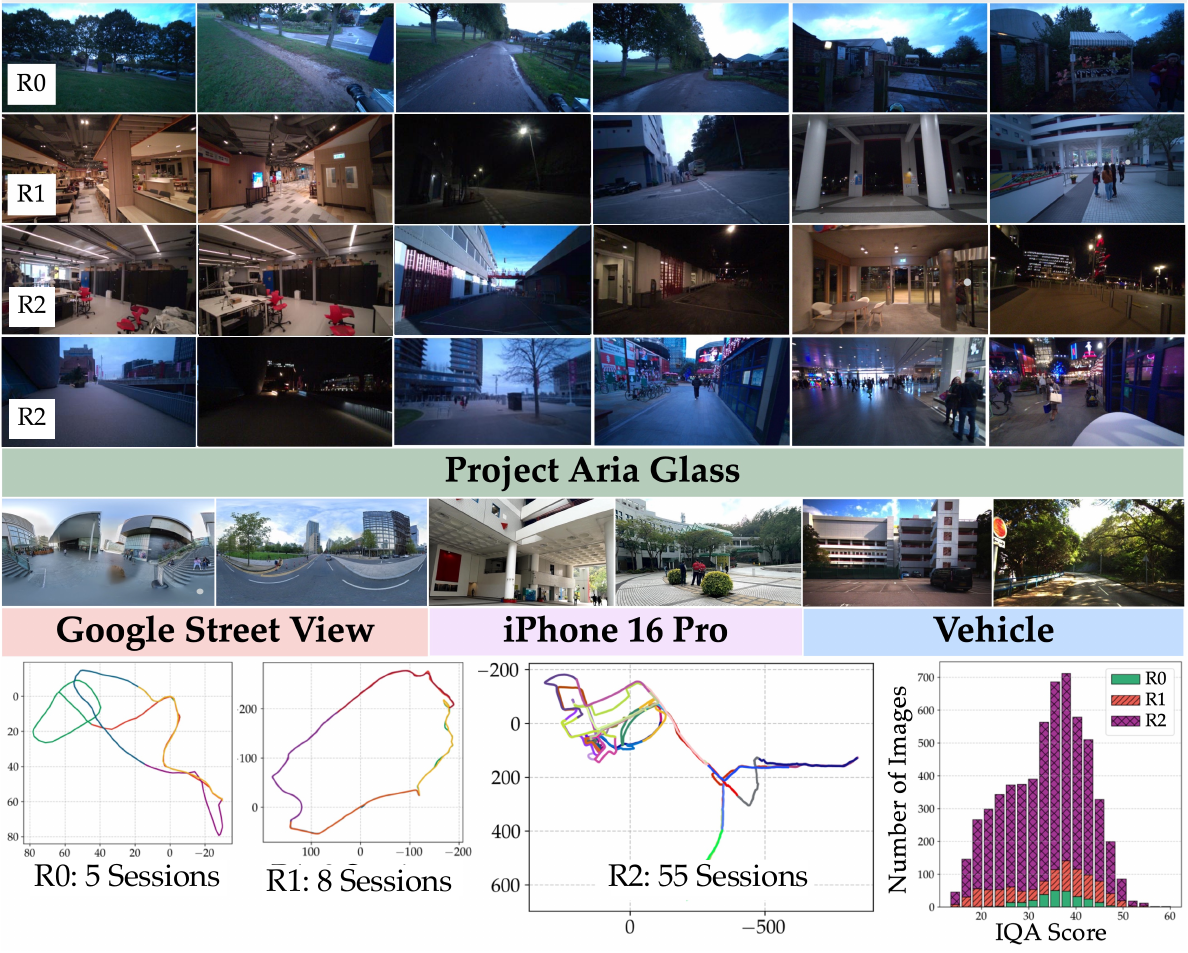}
  \vspace{-0.3cm}
  \caption{Overview of our self-collected dataset and trajectories captured by various devices. The data collected using Aria glasses covers diverse environments, including offices, campuses, shopping centers, and vineyards, across two countries, spanning approximately $3.5$ months, $35$ sequences, and over $18.7$km of traversed distance.}
  \label{fig:exp_datasets}
  \vspace{-0.3cm}
\end{figure}

\subsubsection{\bt{Evaluation Metrics}}
\label{sec:exp_metrics}

\bt{
  We evaluate the system with metrics tailored to each task: topological localization, metric localization, map merging, and visual navigation.}

\paragraph{\bt{Topological Localization}}
\bt{
  Topological localization tests whether a query node matches a reference at the same place; we use retrieval metrics with a match deemed correct within a $[7.5\text{m}, 75^{\circ}]$ tolerance, set to reflect practical navigation requirements.
  \tb{Recall@1} measures the ability to find valid places, \ie among query nodes with a correct match in the reference map, the fraction ranked first; it is a common metric for VPR.
  \tb{Precision@1} measures the reliability of the top-1 prediction, \ie among query nodes for which a match is returned, the fraction that are correct; since many queries lack a true match, rejecting such false positives is crucial for stability.
  \tb{Average@1} is the mean of the two.}

\paragraph{\bt{Metric Localization}}
\bt{
  We report two metrics for relative pose estimation \cite{arnold2022map}.
  \tb{Precision}@$[1\text{m},10^{\circ}]$ is the fraction of queries whose pose error lies within the threshold, reflecting estimation accuracy.
  \tb{AUC}@$[1\text{m},10^{\circ}]$ is the area under the precision-confidence curve, reflecting how well the confidence score separates correct from incorrect estimates.}

\paragraph{\bt{Map Merging}}
\bt{
  This process couples topological localization, metric localization, PGO, and node culling.
  We assess the global pose accuracy of the merged map by the \tb{ATE} in RMSE, which captures the overall drift and alignment of the estimated poses against GT.
  We further provide qualitative results for merging across heterogeneous devices.}

\paragraph{\bt{Visual Navigation}}
\bt{
  In simulation, we quantify performance by two standard metrics: the \tb{total time} to reach the goal and the executed \tb{path length}.
  Together they expose the trade-offs between our fully vision-based system and conventional approaches that exploit privileged information, such as GT poses and detailed local maps, for planning and execution.}
\subsection{\bt{Exp 1: Topological and Metric Localization}}
\label{sec:exp_localization}

\bt{
This experiment answers \textbf{Q1} by separately benchmarking the topological and metric stages of the proposed localization pipeline.
We present the results in order of increasing difficulty: topological localization for place retrieval, followed by metric localization for precise $6$-DoF pose estimation.}
% , and within each stage we progress from single-scene scenarios to large-scale multi-session settings.}

\subsubsection{\bt{Topological Localization}}
\label{sec:exp_topo_localization}

We evaluate our multi-stage topological localization, comprising VPR, sequence matching, and GV, on the self-collected dataset.
We use MASt3R \cite{leroy2024grounding} as the 2D feature matcher for the GV stage and compare against SeqSLAM \cite{milford2012seqslam} with a sequence length of $20$.
Four SoTA VPR models are evaluated: \tb{AnyLoc} \cite{keetha2023anyloc}, \tb{NetVLAD} \cite{gordo2017end}, \tb{CosPlace} \cite{berton2022rethinking}, and \tb{EigenPlaces} \cite{berton2023eigenplaces}, whose global descriptor dimensions are $49152$, $4096$, $256$, and $256$, respectively.

\begin{figure*}[t]
  \centering
  \includegraphics[width=0.98\linewidth]{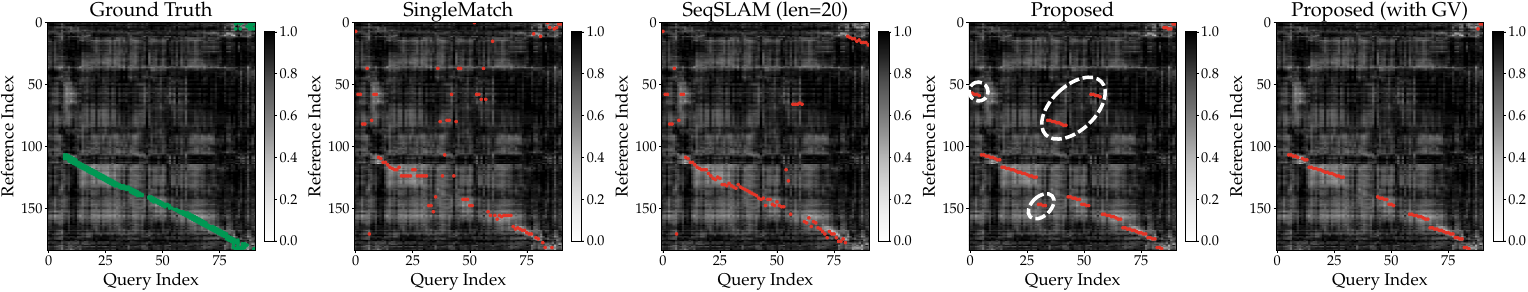}
  \caption{\bt{Difference matrix between reference sequence ID~$=1$ (captured at 2024/12/04, day) and a query (captured at 2024/12/04, night), where darker cells denote larger VPR descriptor distances. We compare the matched paths from the GT, SingleMatch, SeqSLAM, our DP-based matching, and DP with GV. Our DP-based matching produces more sequence-coherent detections, and GV removes false positives (\ie indicated in white circles).}}
  \label{fig:exp_dmatrix_example}
  \vspace{-0.3cm}
\end{figure*}

\begin{table}[t]
  \centering
  \begin{threeparttable}
    \caption{Results of topological localization on our self-collected dataset (R$2$) with metrics: Precision@$1$ \rt{$[\%]$} and Recall@$1$ \rt{$[\%]$} within a threshold of $[7.5\text{m}, 75^{\circ}]$. The \rt{\tb{first}} and \bt{\tb{second}} best results are highlighted in red and blue, respectively.}
    \renewcommand\arraystretch{1.00}
    \renewcommand\tabcolsep{0.9pt}
    \scriptsize
    \begin{tabular}{c|c|c|ccc|ccc|ccc|ccc}
      \toprule[0.03cm]
      \multicolumn{3}{c}{\tb{Ref. Sequence ID}} &
      \multicolumn{3}{c}{$\mathbf{0}$}          & \multicolumn{3}{c}{$\mathbf{1}$} & \multicolumn{3}{c}{$\mathbf{2}$} & \multicolumn{3}{c}{$\mathbf{3}$}                                                                                                                                                \\
      \cmidrule{1-15}
      \multicolumn{3}{c}{$N_{Pos}:N_{Query}$}   & \multicolumn{3}{c}{$838:999$}    & \multicolumn{3}{c}{$208:479$}    & \multicolumn{3}{c}{$235:547$}    & \multicolumn{3}{c}{$187:215$}                                                                                                                \\
      \midrule[0.03cm]
      \midrule[0.03cm]
      \multirow{1}{*}{\textbf{VPR}}             &
      \multirow{1}{*}{\textbf{Matching}}        &
      \multirow{1}{*}{\textbf{GV}}              &
      \textbf{Prec.}                            & \textbf{Rec.}                    & \tb{Avg.}                        &
      \textbf{Prec.}                            & \textbf{Rec.}                    & \tb{Avg.}                        &
      \textbf{Prec.}                            & \textbf{Rec.}                    & \tb{Avg.}                        &
      \textbf{Prec.}                            & \textbf{Rec.}                    & \tb{Avg.}                                                                                                                                                                                                          \\
      \midrule[0.03cm]
      \multirow{6}{*}{\rotatebox{90}{NetVLAD~\cite{arandjelovi2015netvlad}}}
                                                 & SingMatch                        & $\times$                         & $76$                             & $61$                          & $68$           & $18$ & $28$ & $23$           & $16$ & $26$ & $21$           & $67$  & $31$ & $49$           \\
                                                 & SeqSLAM                          & $\times$                         & $75$                             & $56$                          & $66$           & $21$ & $36$ & $29$           & $21$ & $36$ & $29$           & $78$  & $55$ & $67$           \\
                                                 & DP (Ours)                        & $\times$                         & $81$                             & $80$                          & $\bt{\bm{80}}$ & $32$ & $63$ & $48$           & $21$ & $35$ & $28$           & $82$  & $69$ & $\bt{\bm{75}}$ \\
      \cmidrule{2-15}
                                                 & SingMatch                        & $\checkmark$                     & $97$                             & $59$                          & $78$           & $79$ & $26$ & $53$           & $85$ & $26$ & $55$           & $98$  & $30$ & $64$           \\
                                                 & SeqSLAM                          & $\checkmark$                     & $97$                             & $54$                          & $76$           & $93$ & $32$ & $\bt{\bm{62}}$ & $88$ & $35$ & $\rt{\bm{62}}$ & $99$  & $51$ & $75$           \\
                                                 & DP (Ours)                        & $\checkmark$                     & $98$                             & $77$                          & $\rt{\bm{88}}$ & $88$ & $58$ & $\rt{\bm{73}}$ & $84$ & $32$ & $\bt{\bm{58}}$ & $98$  & $66$ & $\rt{\bm{82}}$ \\
      \midrule[0.03cm]

      \multirow{6}{*}{\rotatebox{90}{CosPlace~\cite{berton2022rethinking}}}
                                                 & SingMatch                        & $\times$                         & $77$                             & $64$                          & $71$           & $19$ & $30$ & $24$           & $20$ & $34$ & $27$           & $69$  & $35$ & $52$           \\
                                                 & SeqSLAM                          & $\times$                         & $76$                             & $60$                          & $68$           & $20$ & $33$ & $26$           & $20$ & $33$ & $27$           & $79$  & $58$ & $69$           \\
                                                 & DP (Ours)                        & $\times$                         & $80$                             & $77$                          & $79$           & $24$ & $41$ & $32$           & $28$ & $51$ & $40$           & $80$  & $64$ & $72$           \\
      \cmidrule{2-15}
                                                 & SingMatch                        & $\checkmark$                     & $96$                             & $63$                          & $\bt{\bm{80}}$ & $75$ & $30$ & $52$           & $92$ & $34$ & $63$           & $99$  & $34$ & $66$           \\
                                                 & SeqSLAM                          & $\checkmark$                     & $96$                             & $58$                          & $77$           & $82$ & $30$ & $\bt{\bm{56}}$ & $88$ & $32$ & $\bt{\bm{60}}$ & $100$ & $56$ & $\bt{\bm{78}}$ \\
                                                 & DP (Ours)                        & $\checkmark$                     & $97$                             & $75$                          & $\rt{\bm{86}}$ & $78$ & $40$ & $\rt{\bm{59}}$ & $92$ & $48$ & $\rt{\bm{70}}$ & $97$  & $62$ & $\rt{\bm{79}}$ \\
      \midrule[0.03cm]

      \multirow{6}{*}{\rotatebox{90}{EigenPlaces~\cite{berton2023eigenplaces}}}
                                                 & SingMatch                        & $\times$                         & $77$                             & $63$                          & $70$           & $19$ & $30$ & $25$           & $20$ & $34$ & $27$           & $72$  & $41$ & $57$           \\
                                                 & SeqSLAM                          & $\times$                         & $76$                             & $61$                          & $68$           & $23$ & $39$ & $31$           & $21$ & $35$ & $28$           & $79$  & $57$ & $68$           \\
                                                 & DP (Ours)                        & $\times$                         & $81$                             & $80$                          & $\bt{\bm{80}}$ & $24$ & $41$ & $32$           & $25$ & $44$ & $35$           & $77$  & $52$ & $65$           \\
      \cmidrule{2-15}
                                                 & SingMatch                        & $\checkmark$                     & $96$                             & $61$                          & $79$           & $73$ & $30$ & $51$           & $85$ & $33$ & $59$           & $96$  & $40$ & $68$           \\
                                                 & SeqSLAM                          & $\checkmark$                     & $96$                             & $58$                          & $77$           & $83$ & $38$ & $\rt{\bm{60}}$ & $83$ & $34$ & $\bt{\bm{58}}$ & $97$  & $55$ & $\rt{\bm{76}}$ \\
                                                 & DP (Ours)                        & $\checkmark$                     & $98$                             & $77$                          & $\rt{\bm{87}}$ & $73$ & $38$ & $\bt{\bm{56}}$ & $82$ & $40$ & $\rt{\bm{61}}$ & $98$  & $50$ & $\bt{\bm{74}}$ \\
      \midrule[0.03cm]

      \multirow{6}{*}{\rotatebox{90}{AnyLoc~\cite{keetha2023anyloc}}}
                                                 & SingMatch                        & $\times$                         & $80$                             & $79$                          & $80$           & $31$ & $60$ & $46$           & $32$ & $63$ & $48$           & $83$  & $74$ & $78$           \\
                                                 & SeqSLAM                          & $\times$                         & $77$                             & $65$                          & $71$           & $32$ & $60$ & $46$           & $32$ & $63$ & $48$           & $82$  & $71$ & $77$           \\
                                                 & DP (Ours)                        & $\times$                         & $82$                             & $86$                          & $84$           & $31$ & $59$ & $45$           & $36$ & $74$ & $55$           & $86$  & $92$ & $\rt{\bm{89}}$ \\
      \cmidrule{2-15}
                                                 & SingMatch                        & $\checkmark$                     & $95$                             & $77$                          & $\bt{\bm{86}}$ & $75$ & $57$ & $\bt{\bm{66}}$ & $81$ & $63$ & $72$           & $93$  & $66$ & $80$           \\
                                                 & SeqSLAM                          & $\checkmark$                     & $96$                             & $62$                          & $79$           & $80$ & $53$ & $\rt{\bm{67}}$ & $83$ & $60$ & $\bt{\bm{72}}$ & $98$  & $65$ & $81$           \\
                                                 & DP (Ours)                        & $\checkmark$                     & $96$                             & $82$                          & $\rt{\bm{89}}$ & $74$ & $54$ & $64$           & $80$ & $68$ & $\rt{\bm{74}}$ & $94$  & $85$ & $\rt{\bm{89}}$ \\
      \bottomrule[0.03cm]
    \end{tabular}
    \label{tab:results_self_collected}
  \end{threeparttable}
  \vspace{-0.3cm}
\end{table}

\bt{
  Tab.~\ref{tab:results_self_collected} reports the topological localization results, where $N_{Pos}$ is the number of queries with positive loops and $N_{Query}$ is the total number of query images, so a lower $N_{Pos}/N_{Query}$ ratio means more queries lack any positive loop.
  CosPlace offers the best accuracy-efficiency trade-off, matching AnyLoc and surpassing NetVLAD with a compact $256$-D descriptor.
  We attribute this to its classification-based training objective, which scales to massive datasets, and its large-margin cosine loss, which yields discriminative yet lightweight embeddings.
Post-processing with sequence matching and GV yields at least $10$ points of precision gain across all reference sequences.
Classical SeqSLAM assumes a near-continuous, heavily overlapping trajectory that forms a clear diagonal band in the difference matrix.
This rarely holds in crowdsourced multi-session data, whose overlap follows three patterns (Fig.~\ref{fig:explanation_vpr}): full, partial with gaps, and minimal or non-contiguous.
In the latter two, the valid loop-closure path is discontinuous, so forcing a single continuous path makes SeqSLAM miss valid matches or accumulate spurious ones.
Our DP-based approach instead supports two path-expansion operations, in-sequence matching for continuous segments and out-of-sequence jumping to bridge trajectory gaps.
The benefit is evident in the results: our method attains higher recall on Reference $0$, whose non-contiguous overlap defeats SeqSLAM's fragmented path, while the precision gain is largest on References $1$ and $2$, where GV filters the spurious candidates introduced by jump operations, improving precision by up to $60\%$.}

\bt{As illustrated in Fig.~\ref{fig:exp_dmatrix_example} for reference sequence ID $1$, our DP-based method recovers a more sequence-coherent path than SingleMatch and SeqSLAM.
However, the difference matrix reveals several challenges: not all queries have a valid reference correspondence, not all valid query-reference pairs exhibit small descriptor distances, and some invalid pairs yield deceptively close distances. The subsequent GV stage is thus designed to reject these false positives. 
For space reasons, the \textbf{Suppl.~Mat.}~\cite{anonymous2025opennavmap_supp} visualizes the GT and estimated matching pairs for all reference sequences.}

\begin{figure}[t]
  \centering
  \subfigure[Map-Free dataset ($65$ scenes)]
  {\label{fig:exp_rpe_mapfree}\centering\includegraphics[width=0.99\linewidth]{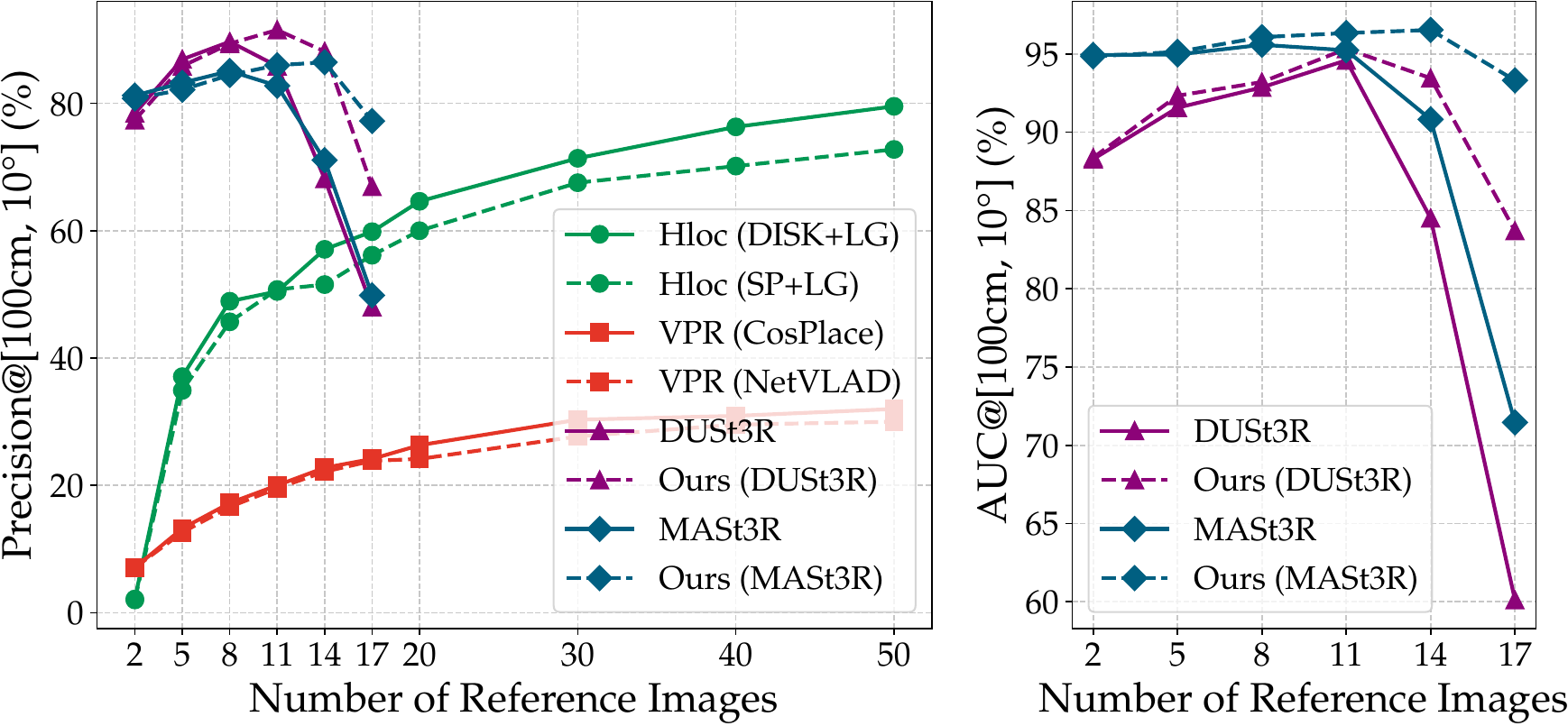}}
  \subfigure[GZ-Campus dataset ($30$ scenes)]
  {\label{fig:exp_rpe_hkustgzcampus}\centering\includegraphics[width=0.99\linewidth]{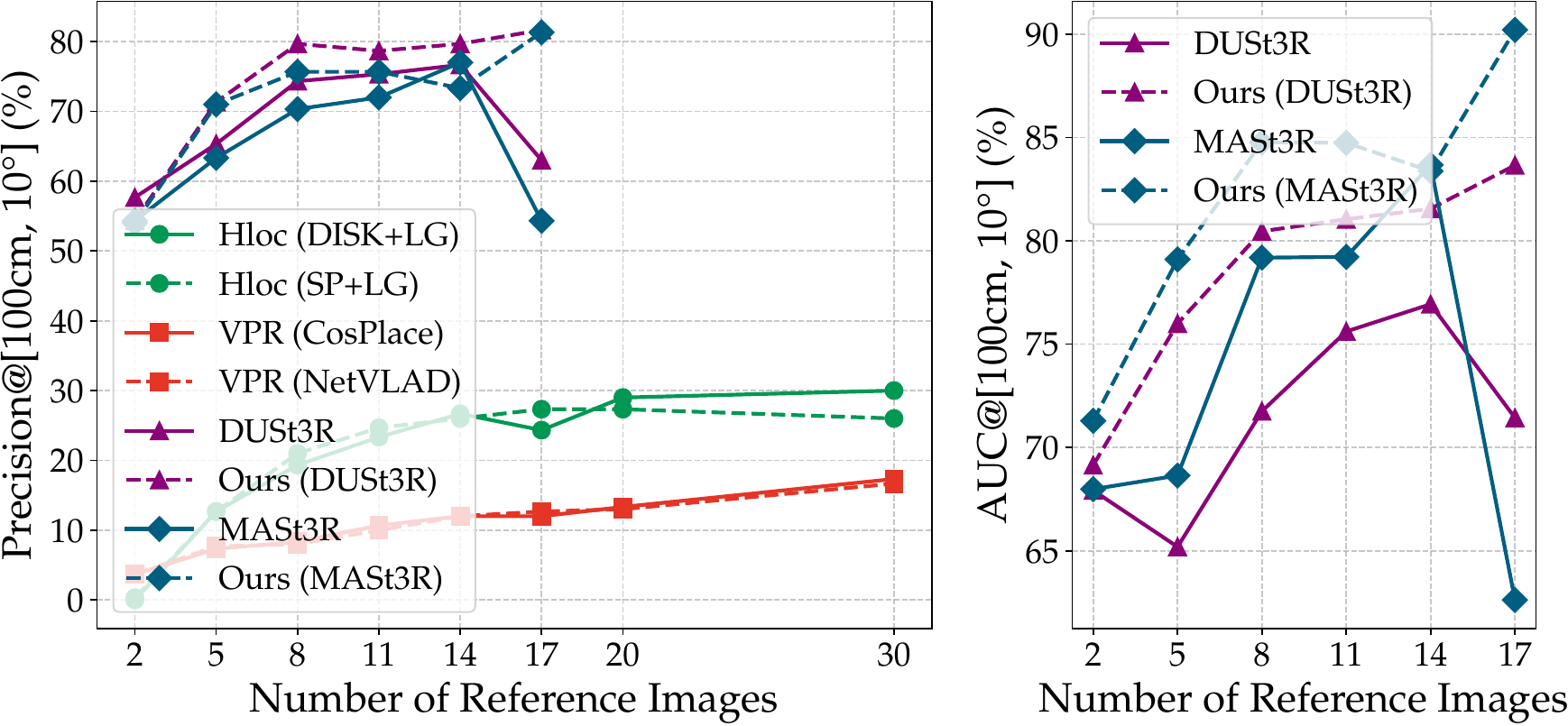}}
  % \subfigure[$360$Loc dataset]
  % {\label{fig:exp_rpe_360loc}\centering\includegraphics[width=0.99\linewidth]{Imgs/experiment/rpe_estimation/360loc_rpe_results-crop}}
  \subfigure[Our self-collected dataset ($57$ scenes)]
  {\label{fig:exp_rpe_uclcampus}\centering\includegraphics[width=0.99\linewidth]{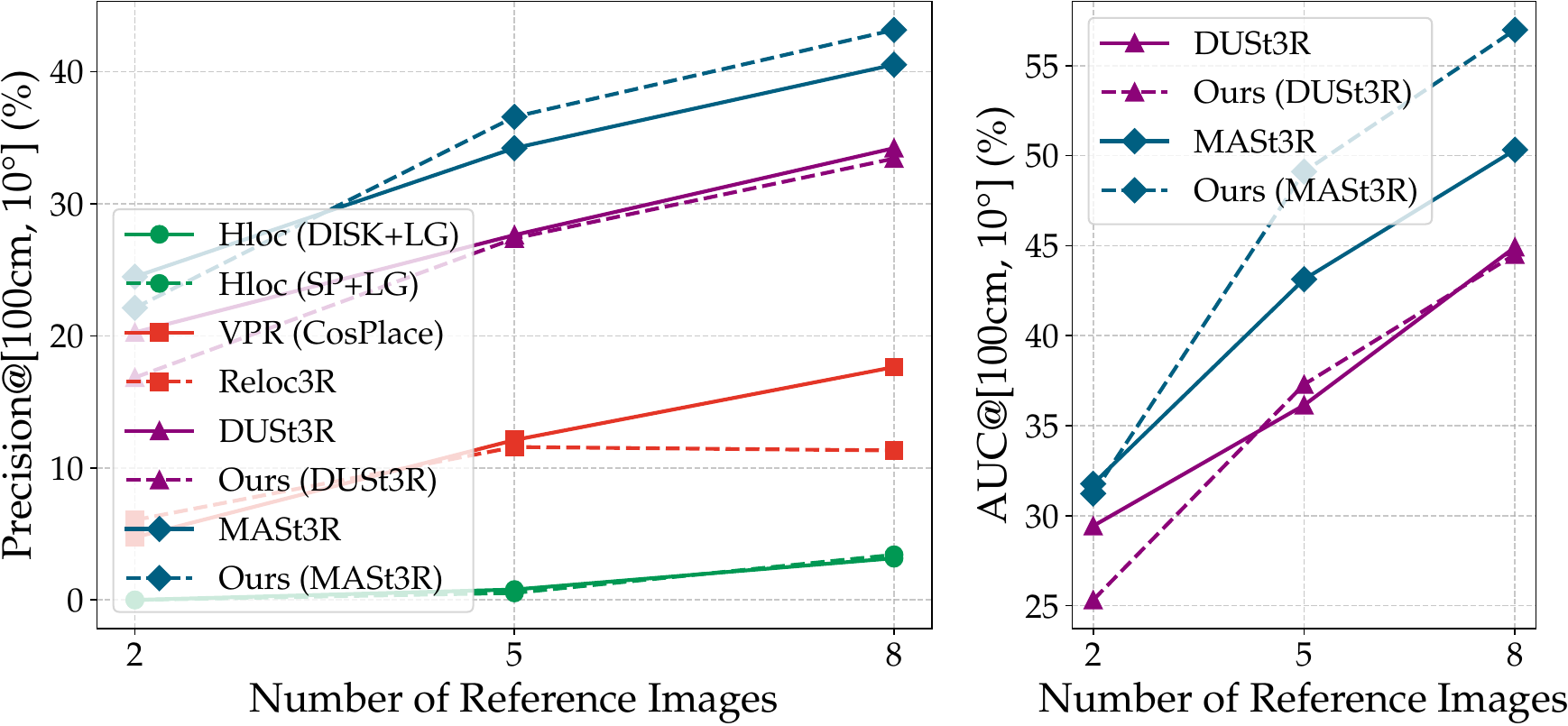}}
  \caption{\rt{Evaluation of metric localization performance across three datasets. We report Precision $[100cm, 10^{\circ}](\%)$ and Area Under the Curve (AUC) as a function of the number of reference images $N$.
      The range of $N$ varies across datasets, this is because the maximum number of all available views in each scene is different.
      To ensure statistical robustness, the performance for each $N$ is averaged over $10$ randomized selections of reference groups across all scenes in the dataset. Note that AUC curves are plotted exclusively for DUSt3R- and MASt3R-based methods (including ours), as other methods do not output explicit estimation confidence scores.}}
  \label{fig:exp_rpe_estimation}
  \vspace{-0.3cm}
\end{figure}

\begin{figure}[t]
  \centering
  \subfigure[Map-Free Dataset]
  {\label{fig:exp_rpe_mapfree_viz}\centering\includegraphics[width=0.995\linewidth]{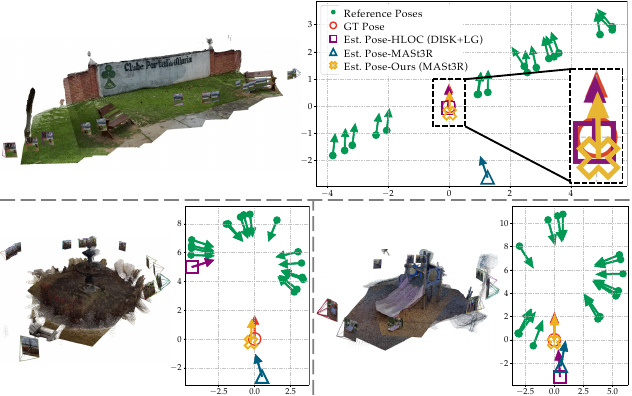}}
  \subfigure[GZ-Campus Dataset]
  {\label{fig:exp_rpe_hkustgzcampus_viz}\centering\includegraphics[width=0.402\linewidth]{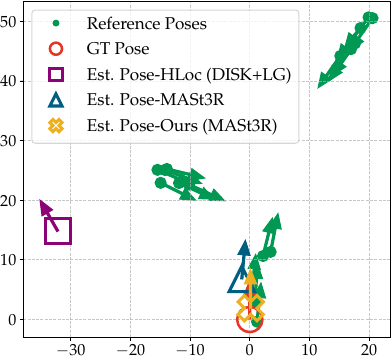}}
  \subfigure[Our Dataset]
  {\label{fig:exp_rpe_uclcampus_viz}\centering\includegraphics[width=0.583\linewidth]{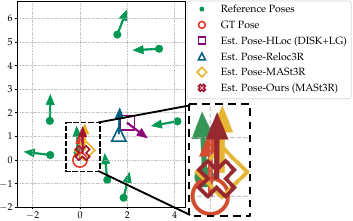}}
  \caption{Sample results of metric localization on three datasets. The accompanying local reconstructions are generated as by-products using both reference images (provided GT poses) and query images. The pose distributions of reference images vary from different datasets.}
  \label{fig:exp_rpe_estimation_viz}
  \vspace{-0.3cm}
\end{figure}

%%%%%%%%%%%%%%%%%%%%%%%%%%%%%%%%%%%%%%%%%%%%
%%%%%%%%%%%%%%%%%%%%%%%%%%%%%%%%%%%%%%%%%%%%
%%%%%%%%%%%%%%%%%%%%%%%%%%%%%%%%%%%%%%%%%%%%
\subsubsection{\bt{Metric Localization}}
\label{sec:exp_metric_level_localization}

Metric localization entails estimating the 6-DoF relative pose of a query image with respect to a set of reference images. We conduct evaluations across three diverse datasets that pose distinct challenges:
\begin{itemize}
  \item \textbf{Map-Free} \cite{arnold2022map}: Represents tourist-centric scenarios characterized by object-focused imagery, long-term temporal changes, and extreme viewpoint variations ($>90^{\circ}$).
  \item \textbf{GZ-Campus} \cite{jiao2025litevloc} and \textbf{Self-Collected}: Capture daily routine routes in large-scale environments, introducing challenges such as structural ambiguities, repetitive textures, and large spatial baselines.
\end{itemize}

\bt{We randomly sample $10$ groups of $N$ reference images per scene and average results over all $M$ scenes, reducing selection bias and ensuring statistically robust accuracy and AUC. Several SoTA methods serve as baselines:}
\begin{itemize}[leftmargin=0.7cm]
  \item \textit{Landmark-based Methods:} We evaluate \tb{HLoc} pipelines configured with different keypoint extractors: \tb{HLoc (DISK+LG)} \cite{tyszkiewicz2020disk} and \tb{HLoc (SuperPoint+LG)} \cite{detone2018superpoint}, both utilizing LightGlue \cite{lindenberger2023lightglue} for matching. These methods rely on COLMAP \cite{schoenberger2016sfm} to construct an explicit SfM map and then localize a query image within the map \cite{sarlin2019coarse}.
  \item \textit{VPR-based Topological Methods:} We employ \tb{VPR (CosPlace-$\bm{256}$)} and \tb{VPR (NetVLAD-$\bm{4096}$)}. They approximate metric localization by assigning the pose of the top-1 retrieved reference image to the query.
  \item \textit{Pose Regression Methods:} We include \tb{Reloc3R} \cite{dong2025reloc3r}, which regresses relative pose between reference images and the query image via a neural network, followed by motion averaging for the final estimate.
  \item \textit{3D GFM-based Methods:} We compare against \tb{DUSt3R} \cite{wang2024dust3r} and \tb{MASt3R} \cite{leroy2024grounding} utilizing global optimization. Our method, denoted as \tb{Ours (DUSt3R)} and \tb{Ours (MASt3R)}, distinguishes itself by introducing a confidence map calibration mechanism to refine the optimization. Unlike the first category, these latter three categories (VPR, Regression, and GFM) operate without storing explicit 3D environment geometry.
\end{itemize}

\bt{
  Fig.~\ref{fig:exp_rpe_estimation} reports pose estimation accuracy across three datasets, where we plot the AUC for DUSt3R- and MASt3R-related methods since they output estimation confidence.
  On Map-Free, our method attains over $80\%$ precision with only two reference images per query, whereas HLoc (DISK+LG) needs more than $50$ for comparable precision and degrades rapidly below five views due to insufficient triangulation.
  The other two datasets are more challenging: as shown in Fig.~\ref{fig:exp_rpe_estimation_viz}, GZ-Campus reference poses are distributed irregularly, some lying over $50$m from the query.
  Even so, our method consistently outperforms HLoc-based approaches and benefits more from additional views in the $5$-$11$ range, while other landmark-free baselines fail to match this accuracy.}

\bt{
  Confidence map calibration both enhances optimization and quantifies estimate reliability, and our method achieves higher precision and AUC than DUSt3R and MASt3R in most scenarios.
  Notably, once views reach $14$ or more on Map-Free and GZ-Campus, all 3D GFM-based methods decline, likely from higher outlier rates in pointmap prediction~\cite{tang2025mv} or insufficient convergence, exposing limitations of DUSt3R-series methods.
  Our calibration mitigates this degradation more effectively than the baselines, and the AUC analysis confirms the CCM is a more reliable indicator than uncalibrated confidence, making it effective for rejecting false matches.}

%%%%%%%%%%%%%%%%%%%%%%%%%%%%%%%%%%%%%%%%%%%%%%%%%%%%%%%%%%
\begin{table}[]
  \centering
  \caption{Theoretical map size per image and relative ratio of baselines compared to ours.}
  \renewcommand\arraystretch{1.00}
  \renewcommand\tabcolsep{4.5pt}
  \footnotesize
  \begin{tabular}{c|c|c}
    \toprule[0.03cm]
    \tb{Methods}   & \tb{Parameters}            & \tb{Map Size} $[\text{MB}]$
    \\
    \midrule[0.03cm]
    Ours           & $H=512, W=288, C\in(0, 1]$ & $0.423CN$                        \\
    Hloc (DISK+LG) & $M=5000, D=128$            & $1.22N$ ($\frac{2.89}{C}\times$) \\
    Hloc (SP+LG)   & $M=4096, D=256$            & $2N$ ($\frac{4.66}{C}\times$)    \\
    \bottomrule[0.03cm]
    \multicolumn{3}{l}{*$N$: Number of reference images for a map.}
    % \multicolumn{3}{l}{$h$, $w$, $c$: height, width, and compression ratio of the image}\\
    % \multicolumn{3}{l}{$m$, $d$: number of features and the dimensions of the feature descriptor; $n$ is the number of reference images.} \\
  \end{tabular}
  \label{tab:exp_rpe_the_map_size}
  \vspace{-0.3cm}
\end{table}
%%%%%%%%%%%%%%%%%%%%%%%%%%%%%%%%%%%%%%%%%%%%%%%%%%%%%%%%%%

\begin{table}[t]
  \centering
  \caption{\rt{Theoretical and empirical map sizes for different examples.}}
  \renewcommand\arraystretch{1.00}
  \renewcommand\tabcolsep{3.3pt}
  \footnotesize
  \begin{tabular}{c|c|c|c}
    \toprule[0.03cm]
    \tb{Dataset} & \multirow{2}{*}{\tb{Method}} & \tb{Theoretical}           & \tb{Empirical}                           \\
    \tb{Scene}   &                              & \tb{Size} $[\text{MB}]$           & \tb{Size} $[\text{MB}]$                         \\
    \midrule[0.03cm]
    \multirow{3}{*}{\makecell[c]{Map-Free                                                                               \\ s$00524$, $N=17$}}
                 & Ours                         & $7.2C$                     & $1.3\ (C\approx 0.18)$                   \\
                 & HLoc (DISK+LG)               & $20.7$                     & $22.4$                                   \\
                 & HLoc (SP+LG)                 & $34.0$                     & $12.3$                                   \\

    \midrule
    \multirow{3}{*}{\makecell[c]{GZ-Campus                                                                              \\ s$00039$, $N=30$}}
                 & Ours                         & $12.7C$                    & $1.8\ (C\approx0.14)$                    \\
                 & HLoc (DISK+LG)               & $36.6$                     & $39.7$                                   \\
                 & HLoc (SP+LG)                 & $60.0$                     & $21.1$                                   \\

    \midrule
    \multirow{2}{*}{\makecell[c]{Self-Collected                                                                         \\ \bt{R$2$}, $N=4273$}}
                 & \multirow{2}{*}{Ours}        & \multirow{2}{*}{$1802.7C$} & \multirow{2}{*}{$213.0\ (C\approx0.12)$} \\
                 &                              &                            &                                          \\
    \bottomrule[0.03cm]
  \end{tabular}
  \label{tab:exp_rpe_map_size_theory_vs_real}
  \vspace{-0.3cm}
\end{table}

\subsubsection{\bt{Map Size Analysis}}
\label{sec:exp_rl_map_size}
Following \cite{brachmann2023accelerated}, we estimate the theoretical map size to assess the compactness of our representation.
Our method stores only resized RGB images, whereas HLoc-based maps additionally require a 3D point cloud, per-point feature descriptors, covisibility information, and a visual vocabulary.
\bt{
  Considering only feature descriptors, the largest portion of the map, with $N$ reference images, $M$ the maximum features per image, and $D$-dimensional descriptors at half-precision ($2$ bytes per dimension), the map size is $N \times 2DM$ bytes.
  For landmark-free maps, stored JPEG images dominate, giving $N \times 3WHC$ bytes, where $W$ and $H$ are the image dimensions and $C$ is the average JPEG compression ratio.
  Tab.~\ref{tab:exp_rpe_the_map_size} reports theoretical sizes for our method and HLoc-based approaches, and Tab.~\ref{tab:exp_rpe_map_size_theory_vs_real} compares theoretical and empirical sizes for three scenes.
  Our method requires far less storage.
  HLoc's empirical size falls below its theoretical estimate due to sparse feature extraction in texture-less scenes, at the cost of localization precision.
  Finally, while 3D GFMs introduce a fixed model-weight overhead (\eg $2.6$GB for MASt3R), this cost is independent of scale, making our solution more efficient as the map grows.}

%%%%%%%%%%%%%%%%%%%%%%%%%%%%%%%%%%%%%%%%%%%%
%%%%%%%%%%%%%%%%%%%%%%%%%%%%%%%%%%%%%%%%%%%%
%%%%%%%%%%%%%%%%%%%%%%%%%%%%%%%%%%%%%%%%%%%%
\subsection{\bt{Exp 2: Single-Device Multi-Session Map Merging}}
\label{sec:exp_map_merging_homo}

%%%%%%%%%%%%%%%%%%%%%%%%%%%%%%%%%%%%%%%%%%%%%%%%%%%%%%%%%%
% R0: vineyard, R1: hkust_campus, R2: ucl_campus_aria
\begin{table}[t]
  \centering
  \caption{\bt{ATE of landmark-based baselines and \methodname on R$0$-R$2$-InOrder under three threshold groups.}}
  \renewcommand\arraystretch{1.00}
  \renewcommand\tabcolsep{0.7pt}
  \scriptsize
  \begin{tabular}{c|c|cc|c|c|c}
    \toprule[0.03cm]
    \multirow{4}{*}{\textbf{Data}} & \multirow{4}{*}{\textbf{Method}}
      & \multicolumn{2}{c|}{\textbf{Threshold}}
      & \multirow{4}{*}{\textbf{\makecell[c]{Trans. \\ ATE $[\text{m}]$}}}
      & \multirow{4}{*}{\textbf{\makecell[c]{Rot.\\ ATE $[$\text{deg}$]$}}}
      & \multirow{4}{*}{\textbf{\makecell[c]{Missing \\ Sessions}}} \\
    \cmidrule[0.01cm]{3-4}
    & & \textbf{\makecell[c]{GV Min \\ Inliers}} & \textbf{\makecell[c]{PnP Min \\ Inliers}} & & & \\
    \midrule[0.03cm]
    \multirow{7}{*}{\makecell[c]{R$0$-InOrder \\  $4$ sessions}}
      & \multirow{3}{*}{\makecell[c]{HLoc-COLMAP \\ (NetVLAD+DISK+LG)}}
        & $300$ & $70$ & $1.75$ & $1.52$ & $0$ \\
      & & $400$ & $110$ & $3.33$ & $2.88$ & $0$ \\
      & & $500$ & $150$ & $3.18$ & $1.71$ & $0$ \\
    \cmidrule[0.01cm]{2-7}
      & \multirow{3}{*}{\makecell[c]{HLoc-COLMAP \\ (NetVLAD+SP+LG)}}
        & $100$ & $25$ & $23.28$ & $68.71$ & $0$ \\
      & & $120$ & $35$ & $0.24$ & $0.45$ & $1$ \\
      & & $150$ & $50$ & $0.26$ & $0.49$ & $1$ \\
    \cmidrule[0.01cm]{2-7}
      & Proposed & $100$ & -- & $0.47$ & $0.70$ & $0$ \\
    \midrule[0.03cm]
    \multirow{7}{*}{\makecell[c]{R$1$-InOrder \\ $7$ sessions}}
      & \multirow{3}{*}{\makecell[c]{HLoc-COLMAP \\ (NetVLAD+DISK+LG)}}
        & $300$ & $70$ & $0.79$ & $0.36$ & $4$ \\
      & & $400$ & $110$ & $0.78$ & $0.36$ & $4$ \\
      & & $500$ & $150$ & $0.77$ & $0.36$ & $4$ \\
    \cmidrule[0.01cm]{2-7}
      & \multirow{3}{*}{\makecell[c]{HLoc-COLMAP \\ (NetVLAD+SP+LG)}}
        & $100$ & $25$ & $1.31$ & $0.69$ & $4$ \\
      & & $120$ & $35$ & $0.85$ & $0.40$ & $4$ \\
      & & $150$ & $50$ & $1.11$ & $0.51$ & $4$ \\
    \cmidrule[0.01cm]{2-7}
      & Proposed & $100$ & -- & $1.18$ & $0.44$ & $0$ \\
    \midrule[0.03cm]
    \multirow{7}{*}{\makecell[c]{R$2$-InOrder \\ $55$ sessions}}
      & \multirow{3}{*}{\makecell[c]{HLoc-COLMAP \\ (NetVLAD+DISK+LG)}}
        & $300$ & $70$ & $104.95$ & $55.49$ & $21$ \\
      & & $400$ & $110$ & $61.08$ & $58.25$ & $23$ \\
      & & $500$ & $150$ & $33.87$ & $16.04$ & $26$ \\
    \cmidrule[0.01cm]{2-7}
      & \multirow{3}{*}{\makecell[c]{HLoc-COLMAP \\ (NetVLAD+SP+LG)}}
        & $100$ & $25$ & $90.97$ & $67.88$ & $23$ \\
      & & $120$ & $35$ & $8.06$ & $7.19$ & $30$ \\
      & & $150$ & $50$ & $3.50$ & $30.07$ & $36$ \\
    \cmidrule[0.01cm]{2-7}
      & Proposed & $100$ & -- & $1.51$ & $1.71$ & $1$ \\
    \bottomrule[0.03cm]
  \end{tabular}
  \label{tab:exp_collaborative_baseline}
  \vspace{-0.3cm}
\end{table}
%%%%%%%%%%%%%%%%%%%%%%%%%%%%%%%%%%%%%%%%%%%%%%%%%%%%%%%%%%

%%%%%%%%%%%%%%%%%%%%%%%%%%%%%%%%%%%%%%%%%%%%%%%%%%%%%%%%%%
\begin{table}[t]
  \centering
  \caption{ATE across different scenarios in our self-collected dataset.}
  \renewcommand\arraystretch{1.00}
  \renewcommand\tabcolsep{5.0pt}
  \scriptsize
  \begin{tabular}{c|c|c|c|c|c}
    \toprule[0.03cm]
    \textbf{Data} & \textbf{\makecell[c]{Distance                                     \\Time Spans}}             & \textbf{Shuffle}
                  & \textbf{\makecell[c]{Translational                                \\ ATE $[\text{m}]$}}              & \textbf{\makecell[c]{Rotational \\ ATE $[\text{deg}]$}} & \textbf{\makecell[c]{Missing \\ Sessions}}                  \\
    \midrule[0.03cm]
    R$0$-InOrder  & \multirow{3}{*}{\makecell[c]{$0.6$km                              \\$6$mins}}
                  & N                                      & $0.47$ & $0.70$          & $0$ \\
    R$0$-$0$      &                                        & Y      & $0.65$ & $0.86$ & $0$ \\
    R$0$-$1$      &                                        & Y      & $0.27$ & $0.50$ & $0$ \\
    \midrule[0.03cm]
    R$1$-InOrder  & \multirow{3}{*}{\makecell[c]{$2.5$km                              \\$18$hours}}
                  & N                                      & $1.18$ & $0.44$          & $0$ \\
    R$1$-$0$      &                                        & Y      & $2.39$ & $1.15$ & $0$ \\
    R$1$-$1$      &                                        & Y      & $1.27$ & $0.41$ & $0$ \\
    \midrule[0.03cm]
    R$2$-InOrder  & \multirow{10}{*}{\makecell[c]{$15.7$km                            \\$110$days}}
                  & N                                      & $1.51$ & $1.71$          & $1$ \\
    R$2$-$0$      &                                        & Y      & $2.30$ & $1.10$ & $0$ \\
    R$2$-$1$      &                                        & Y      & $2.87$ & $1.52$ & $0$ \\
    R$2$-$2$      &                                        & Y      & $2.30$ & $1.56$ & $0$ \\
    R$2$-$3$      &                                        & Y      & $1.93$ & $2.06$ & $0$ \\
    R$2$-$4$      &                                        & Y      & $1.78$ & $0.98$ & $0$ \\
    R$2$-$5$      &                                        & Y      & $2.05$ & $1.52$ & $0$ \\
    R$2$-$6$      &                                        & Y      & $1.80$ & $1.35$ & $0$ \\
    R$2$-$7$      &                                        & Y      & $1.72$ & $0.79$ & $0$ \\
    R$2$-$8$      &                                        & Y      & $2.58$ & $1.00$ & $0$ \\
    % \rt{G$2$-WO-Calib} & & N &$73.015$ & $28.046$ \\
    % \rt{G$2$-WO-Pruning} & & N &$148.206$ & $50.697$ \\
    \bottomrule[0.03cm]
    \multicolumn{6}{l}{\rt{*Distance and time spans are totals over all sessions in each region.}}
  \end{tabular}
  \label{tab:exp_collaborative_mapping}
  \vspace{-0.3cm}
\end{table}
%%%%%%%%%%%%%%%%%%%%%%%%%%%%%%%%%%%%%%%%%%%%%%%%%%%%%%%%%%

\bt{
This experiment addresses \textbf{Q2} by evaluating the accuracy and robustness of \methodname in merging multi-session submaps into a globally consistent map.
The pipeline incrementally aligns incoming submaps into a global coordinate frame through four sequential stages: topological localization, metric localization, PGO, and node culling.
Experiments are conducted on our self-collected dataset spanning $15.7$km over $110$ days across three geographic regions (R$0$--R$2$) of increasing scale and temporal diversity.
We evaluate two ordering protocols: \textit{InOrder}, which processes sessions chronologically and benefits from spatial overlap between consecutive submaps, and \textit{Shuffled}, which randomizes session order to simulate real crowdsourced deployment where temporal proximity does not guarantee spatial overlap.
Each shuffled trial is denoted R$i$-$j$, the $j$-th random ordering of region R$i$.}

\subsubsection{\bt{Comparison with Landmark-Based Methods}}
\label{sec:exp_orbslam3}

\begin{figure*}[t]
  \centering
  \subfigure[$10$ submaps]
  {\label{fig:exp_colla_10}\centering\includegraphics[width=0.194 \linewidth]
    {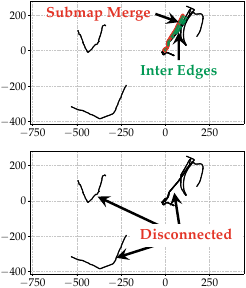}
  }
  \hspace{-0.30cm}
  \subfigure[$22$ submaps]
  {\label{fig:exp_colla_22}\centering\includegraphics[width=0.193\linewidth]
    {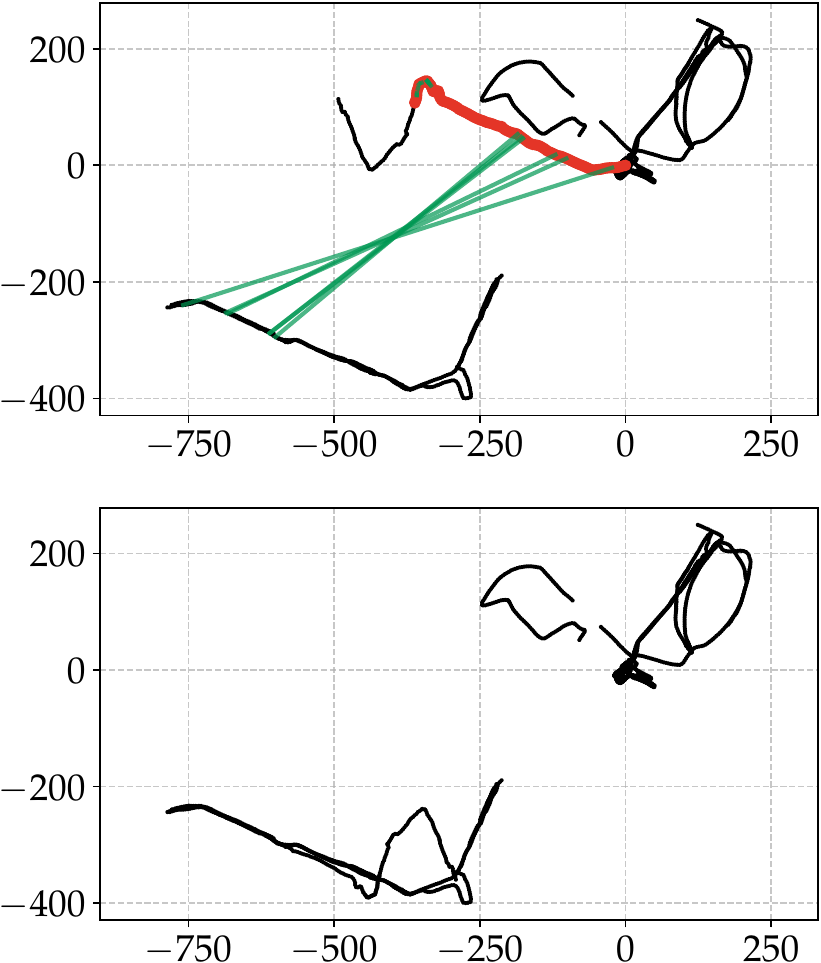}
  }
  \hspace{-0.30cm}
  \subfigure[$27$ submaps]
  {\label{fig:exp_colla_27}\centering\includegraphics[width=0.193\linewidth]
    {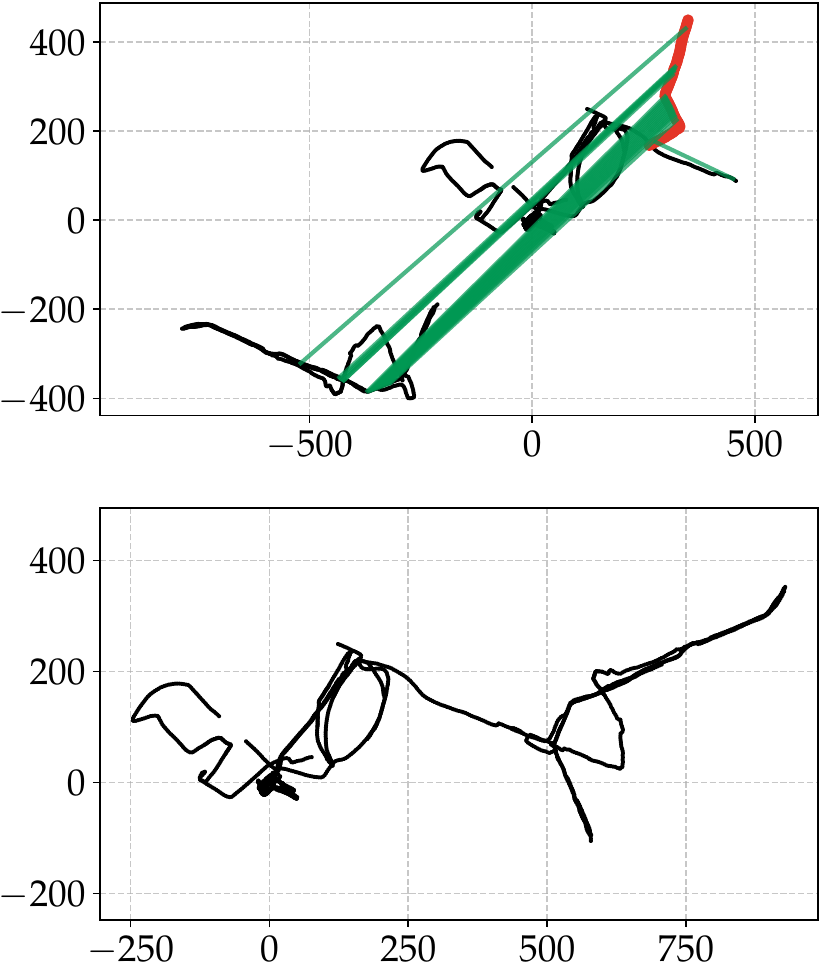}
  }
  \hspace{-0.30cm}
  \subfigure[$36$ submaps]
  {\label{fig:exp_colla_36}\centering\includegraphics[width=0.193\linewidth]
    {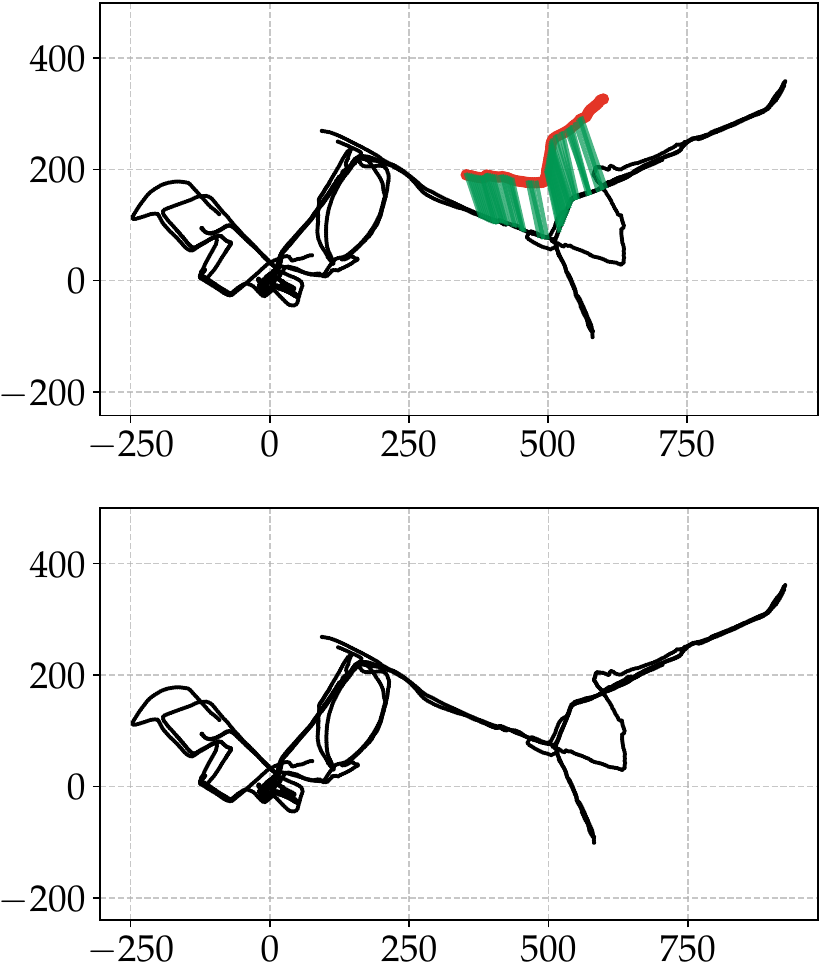}
  }
  \hspace{-0.30cm}
  \subfigure[$55$ submaps]
  {\label{fig:exp_colla_55}\centering\includegraphics[width=0.193\linewidth]
    {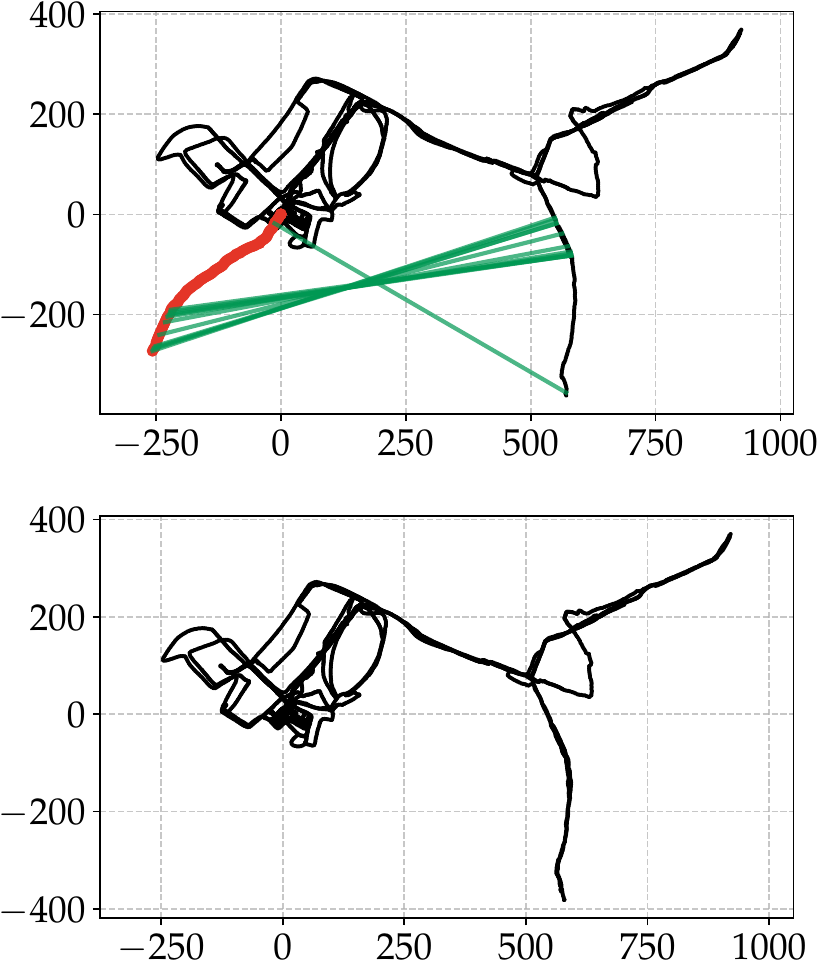}
  }
  \caption{Incremental map merging with submaps added in random order. Disconnected submaps, as in (a), do not affect pose graph optimization. Example shown uses data from R$2$-$4$ (top: before merging, bottom: after merging). Green lines indicate reliable loop closures to establish inter-submap edges.}
  \label{fig:exp_colla_mapping}
  \vspace{-0.3cm}
\end{figure*}

\begin{figure}[t]
  \centering
  \includegraphics[width=0.92\linewidth]{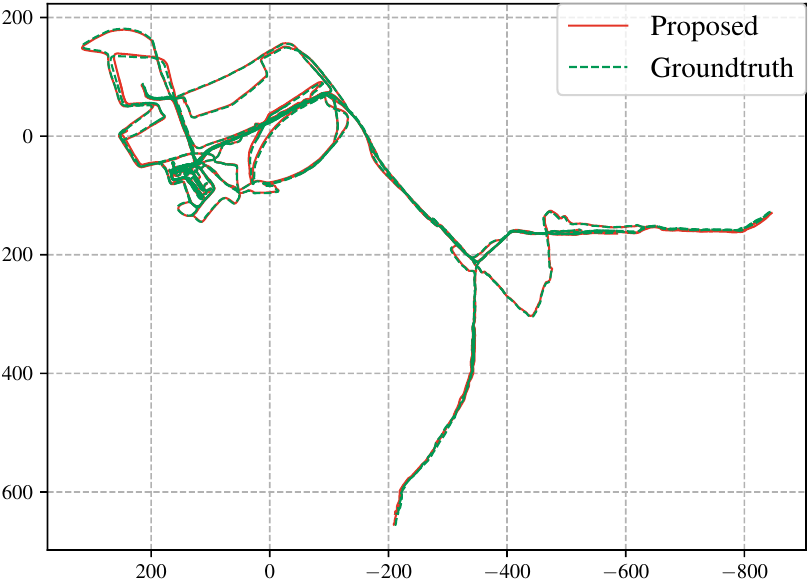}
  \caption{Estimated trajectory of R$2$-$4$ from our multi-session mapping system and GT trajectory from Aria glass-provided SLAM service.}
  \label{fig:exp_colla_mapping_est_gt}
  \vspace{-0.3cm}
\end{figure}

\begin{figure}[t]
  \centering
  \includegraphics[width=0.96\linewidth]{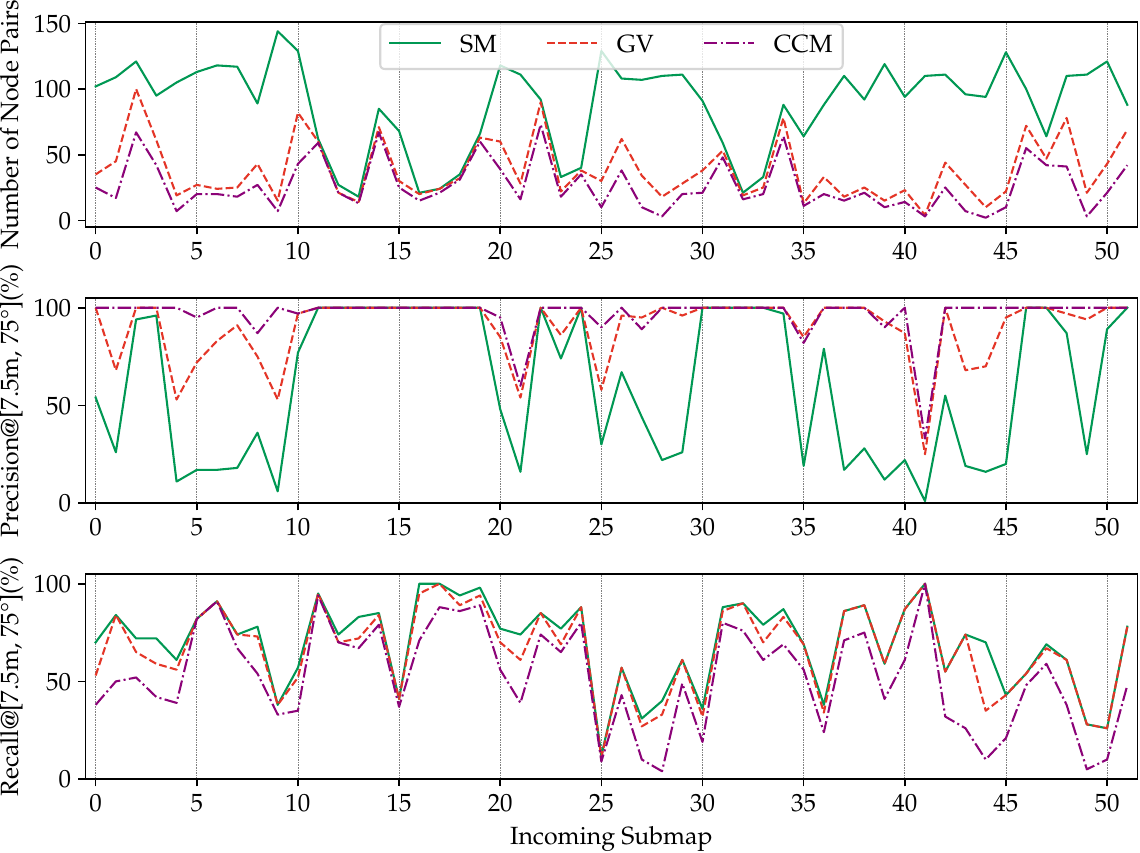}
  \caption{\bt{Precision-recall analysis on the R2-InOrder sequence. For each incoming submap, we plot the number of candidate node pairs, the precision, and the recall at three successive stages: initial sequence matching (SM), geometric verification (GV), and calibrated confidence map (CCM), whose candidate sets are denoted $\mathcal{P}_{\mathrm{SM}}$, $\mathcal{P}_{\mathrm{GV}}$, and $\mathcal{P}_{\mathrm{CCM}}$, respectively. Both GV and CCM act as effective filters, substantially improving precision by rejecting false positives at only a marginal cost in recall.}}
  \label{fig:exp_colla_prec_recall}
  \vspace{-0.4cm}
\end{figure}

\bt{
Extending experiments from Sec.~\ref{sec:exp_metric_level_localization}, we compare our proposed multi-session mapping systems against landmark-based couterparts:
\textbf{HLoc-COLMAP (NetVLAD+SuperPoint+LG)} and \textbf{HLoc-COLMAP (NetVLAD+DISK+LG)}, both of which maintain an explicit SfM map storing camera poses, 2D feature descriptors, and 3D landmarks.
Each submap is reconstructed as an independent SfM model with VIO-initialized camera poses; 3D landmarks are triangulated and globally bundle-adjusted.
A denser keyframe threshold of $0.25$m (vs. $3.9$m in Sec.~\ref{sec:exp_setup}) ensures sufficient 3D coverage for cross-submap localization.
For merging, NetVLAD retrieves cross-submap candidates; correspondences established by SuperPoint+LightGlue or DISK+LightGlue are filtered by GV.
Each verified pair yields a 3D anchor: the matched camera center in the incoming frame and its PnP counterpart in the merged frame.
A rigid $6$-DoF alignment is estimated via Umeyama~\cite{umeyama1989point} with RANSAC and applied to fuse the incoming SfM model.}

\bt{
For both variants, we evaluate three inlier-threshold groups for GV and PnP, from loose to strict.
Looser thresholds retain more cross-submap connections but admit more false positives and may lead to larger errors; stricter ones suppress false positives at the cost of excluding low-overlap submaps.
Tab.~\ref{tab:exp_collaborative_baseline} reports ATE and the number of missing (unmerged) sessions for each threshold group.
On the smaller region R$2$, where the limited number of sessions constrains viewpoint and appearance variation, the baselines outperform \methodname owing to sufficient feature availability for reliable SfM triangulation.
On the larger, more challenging regions R$0$ and R$1$, however, the trend reverses: tightening the thresholds reduces ATE but progressively increases the number of discarded submaps, revealing a stricter accuracy-mapping completeness trade-off than \methodname.}

\subsubsection{\bt{Map Merging with Shuffled Submaps}}

\bt{
  Tab.~\ref{tab:exp_collaborative_mapping} reports ATE across InOrder and Shuffled trials; ATE fluctuates with session order, as it alters the factor-graph topology.
  R$2$ is especially challenging since loop closures may initially be absent due to temporal disconnects, illustrated in Fig.~\ref{fig:exp_colla_mapping} for trial R$2$-$4$; Fig.~\ref{fig:exp_colla_mapping_est_gt} overlays the estimated and Aria-SLAM ground-truth trajectories for this trial.
  Despite the heightened false-positive risk under random ordering, \methodname sustains consistent accuracy, with maximum ATE below $3.0$m and $2.1^{\circ}$ over $15.7$km, by applying GV and CCM with fixed thresholds across all trials.
  When a submap lacks valid loop closures, it is retained as a separate component and merged once reliable constraints are detected, as shown in Figs.~\ref{fig:exp_colla_22}-\ref{fig:exp_colla_27}.}
\bt{
  Fig.~\ref{fig:exp_colla_prec_recall} shows how geometric verification and CCM suppress false positives among candidate node pairs during R$2$-InOrder merging, reporting candidate count, precision, and recall.
  Geometric verification validates local feature matches, while CCM applies a stricter filter on metric localization quality.
  These filters occasionally reduce recall by rejecting geometrically valid but poorly estimated connections, which is nonetheless essential to prevent unreliable edges from degrading the subsequent PGO.}

\subsection{\bt{Exp 3: Cross-Device Multi-Session Map Merging}}
\label{sec:exp_colla_mapping}

% % translational ATE
% 360loc\_S00000\_atrium\_in&  0.548 
% 360loc\_S00000\_concourse\_in&  0.498 
% 360loc\_S00002\_hall\_in&  1.107 
% 360loc\_S00003\_piatrium\_in&  1.113 
% % rotational ATE
% 360loc\_S00000\_atrium\_in&  0.809 
% 360loc\_S00000\_concourse\_in&  0.904 
% 360loc\_S00002\_hall\_in&  2.998 
% 360loc\_S00003\_piatrium\_in&  0.989 

%%%%%%%%%%%%%%%%%%%%%%%%%%%%%%%%%%%%%%%%%%%%%%%%%%%%%%%%%%
\begin{table}[t]
  \centering
  \caption{ATE across different scenarios in the $360$Loc dataset.}
  \renewcommand\arraystretch{1.00}
  \renewcommand\tabcolsep{4.0pt}
  \footnotesize
  \begin{tabular}{c|c|c|c|c}
    \toprule[0.03cm]
    \textbf{Scene} & \textbf{\makecell[c]{Spatial                     Extent $[m]$}} & \textbf{\makecell[c]{Session           \\ Number}}
                   & \textbf{\makecell[c]{Translational                                                                       \\ ATE $[m]$}}              & \textbf{\makecell[c]{Rotational \\ ATE $[deg]$}}                   \\
    \midrule[0.03cm]
    Atrium         & $65\times 36$
                   & $5$                                                             & $0.548$                      & $0.809$ \\
    \midrule[0.01cm]
    Concourse      & $93\times 15$
                   & $4$                                                             & $0.498$                      & $0.904$ \\
    \midrule[0.01cm]
    Hall           & $105\times 52$
                   & $5$                                                             & $1.107$                      & $2.998$ \\
    \midrule[0.01cm]
    Piatrium       & $98\times 70$
                   & $4$                                                             & $1.113$                      & $0.989$ \\
    \bottomrule[0.03cm]
  \end{tabular}
  \label{tab:exp_crowdsourcing_mapping_360loc_ate}
  \vspace{-0.2cm}
\end{table}
%%%%%%%%%%%%%%%%%%%%%%%%%%%%%%%%%%%%%%%%%%%%%%%%%%%%%%%%%%

\begin{figure}[t]
  \centering
  \subfigure[Mapping R$1$ with $13$ sessions: $8$ (Aria) + $2$ (Vehicle) + $3$ (Phone)]
{\label{fig:exp_crowdsourcing_mapping_hkust}\centering\includegraphics[width=0.97\linewidth]{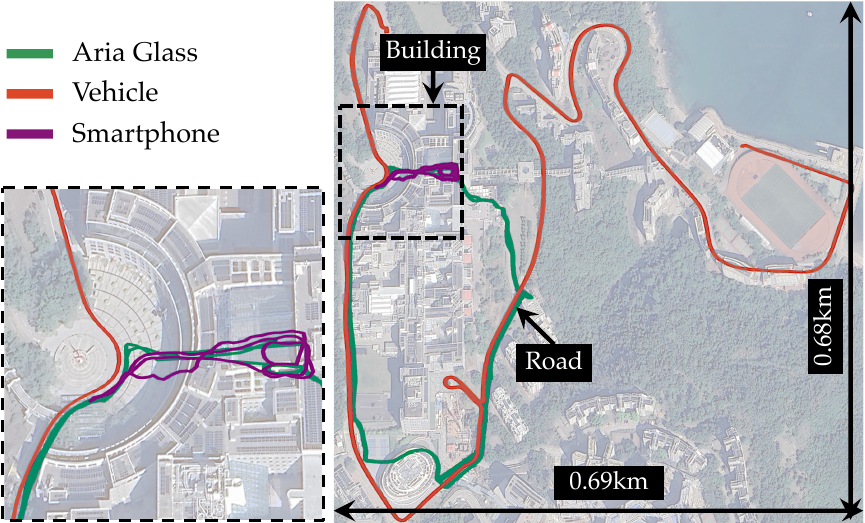}}
  \subfigure[Mapping R$2$ with $57$ sessions: $55$ (Aria) + $2$ (Google Street View)]{\label{fig:exp_crowdsourcing_mapping_ucl}\centering\includegraphics[width=0.97\linewidth]{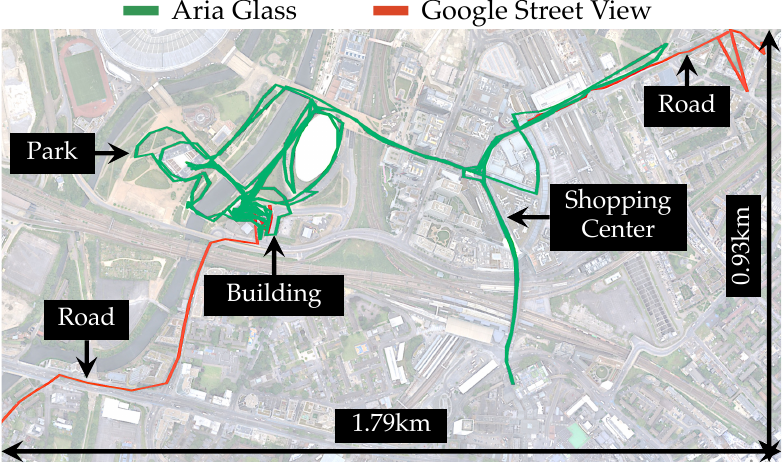}}
  \caption{\rt{Experiments on multi-session mapping with heterogeneous devices across two regions. The estimated map is rigidly transformed into the East-North-Up coordinate system and aligned onto the Google satellite image, enabled by the availability of GPS data for outdoor sequences.}}
  \label{fig:exp_crowdsourcing_mapping}
  \vspace{-0.3cm}
\end{figure}

\bt{
This experiment answers \textbf{Q3} by integrating data from heterogeneous devices into a globally consistent map.
The key distinction from Exp~$2$ is device heterogeneity: sessions are captured with cameras of differing optics, resolutions, and conditions, including fisheye AR glasses, smartphones, vehicle-mounted cameras, and panoramic sources.
Validating the landmark-free design under device heterogeneity is intrinsically hard: unlike landmark-based pipelines that initialize 3D landmarks offline, our system jointly solves local landmark initialization and relative pose estimation online in the cross-device setting; yet it recovers globally consistent inter-session geometry across devices, confirming that the topometric representation remains viable for crowdsourced mapping.}

\subsubsection{\bt{Results on $360$Loc Dataset}}
\bt{Map-merging accuracy, measured as ATE against GT, is reported in Tab.~\ref{tab:exp_crowdsourcing_mapping_360loc_ate}.
  % Fig.~\ref{fig:exp_crowdsourcing_mapping_360loc_} illustrates the resulting map which is merged from multi-session submaps.
  Within each scene, sessions are captured with different camera models, making the dataset well suited for cross-device localization and mapping.
  Accuracy is lower on the outdoor sequences (Hall, Piatrium) than on the indoor ones (Atrium, Concourse), indicating that the multi-session localization pipeline is more effective indoors.
  Due to space limits, the visualization of the merged map is provided in the \textbf{Suppl. Mat.}~\cite{anonymous2025opennavmap_supp}}

\subsubsection{\bt{Results on Self-Collected Dataset}}
Fig.~\ref{fig:exp_crowdsourcing_mapping} shows the resulting global maps for R$1$ and R$2$, overlaid on satellite imagery via GPS.
The preceding experiments reveal that relying on a single device type imposes complementary limitations.
Specialized devices such as the Aria glass capture rich, multi-sensor data but have restricted access.
Smartphones and public street-view services enable broad participation and flexible updates, whereas vehicle-mounted cameras support large-scale collection but are confined to vehicle-accessible areas.
Fusing these heterogeneous sources yields a more comprehensive, extensible map that overcomes these single-source limitations.
\bt{As shown in Fig.~\ref{fig:exp_path_planning}, the expanding map progressively extends the global path planning to new environments, highlighting the practical utility of our framework for lifelong updates.}

\subsection{\bt{Exp 4: Lifelong Map Maintenance}}
\label{sec:exp_keyframe_selection}

\begin{table}[t]
    \centering
    \renewcommand\arraystretch{1.0}
    \renewcommand\tabcolsep{2.3pt}
    \scriptsize
    \caption{\rt{Ablation study on node culling. IQA: image quality assessment, IG: information gain, and TD: temporal difference.}}
    \begin{tabular}{c|c|c|c|c|c|c|c}
        \toprule[0.03cm]
        \textbf{Operations}                    & \textbf{PreFilter} & \multicolumn{2}{c|}{\textbf{Forward}} & \multicolumn{2}{c|}{\textbf{Backward}} & \multicolumn{2}{c}{\textbf{Metrics}}                                                                   \\
        \midrule[0.03cm]
        \textbf{Factors}                       & \textbf{IQA}       & \textbf{IQA}                          & \textbf{IG}                            & \textbf{IG}                          & \textbf{TD} & \textbf{\makecell[c]{Translational                \\ ATE $[m]$$\downarrow$}} & \textbf{\makecell[c]{Rotational \\ ATE $[deg]$$\downarrow$}} \\
        \midrule[0.03cm]
        W.O. Node Culling             & $\times$           & $\times$                              & $\times$                               & $\times$                             & $\times$    & $0.105$                            & $0.337$      \\
        \midrule[0.01cm]
        \multirow{3}{*}{Node Culling} & \checkmark         & \checkmark                            & $\times$                               & $\times$                             & $\times$    & $\bm{0.099}$                       & $\bm{0.338}$ \\
                                               & \checkmark         & \checkmark                            & \checkmark                             & \checkmark                           & $\times$    & $0.108$                            & $0.344$      \\
                                               & \checkmark         & \checkmark                            & \checkmark                             & \checkmark                           & \checkmark  & $0.116$                            & $0.347$      \\
        \bottomrule[0.03cm]
    \end{tabular}
    \label{tab:exp_ablation_node_map_merging}
    % \vspace{-0.4cm}
\end{table}

\begin{figure}[t]
  \centering
  \subfigure[Evolution of the number of nodes in the map during map merging, where the above number indicates the incoming submap ID.]{
    \label{fig:exp_node_culling_map_merging_node_number}
    \includegraphics[width=0.98\linewidth]{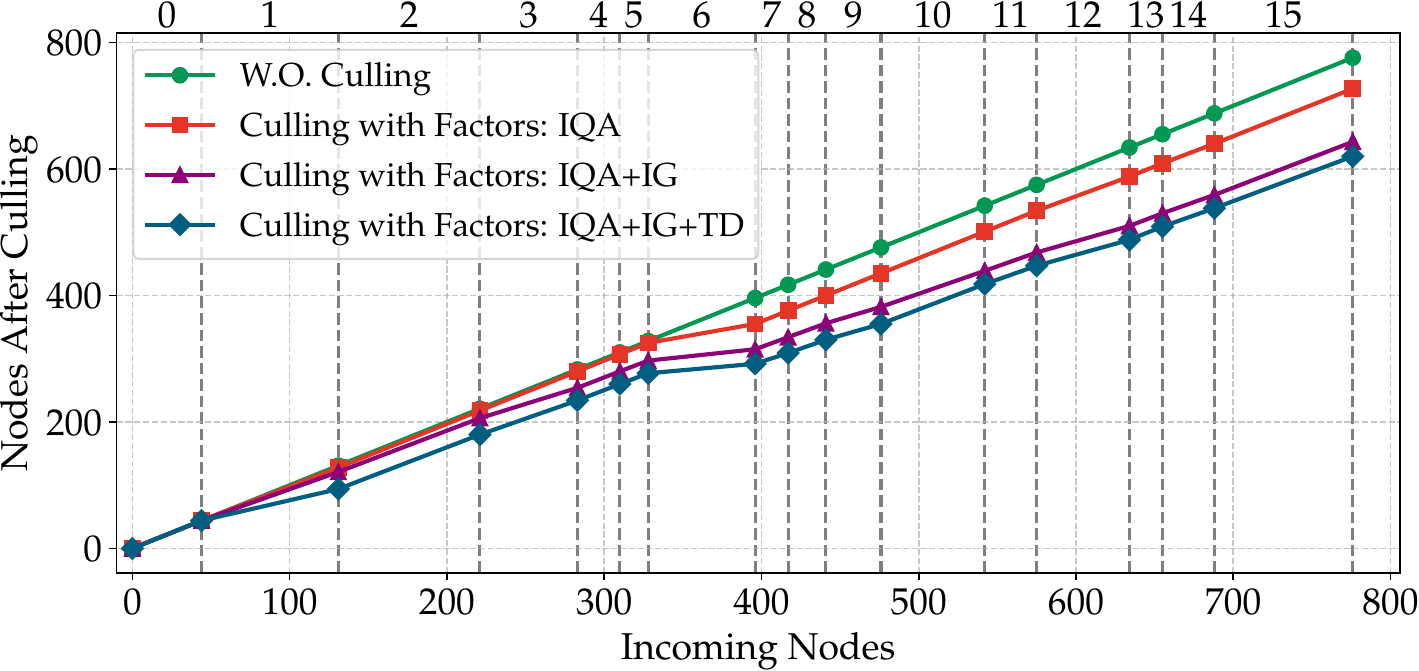}
  }
  \subfigure[Nodes are culled by specific operations.]{
    \label{fig:exp_node_culling_map_merging_node_operation}
    \includegraphics[width=0.98\linewidth]{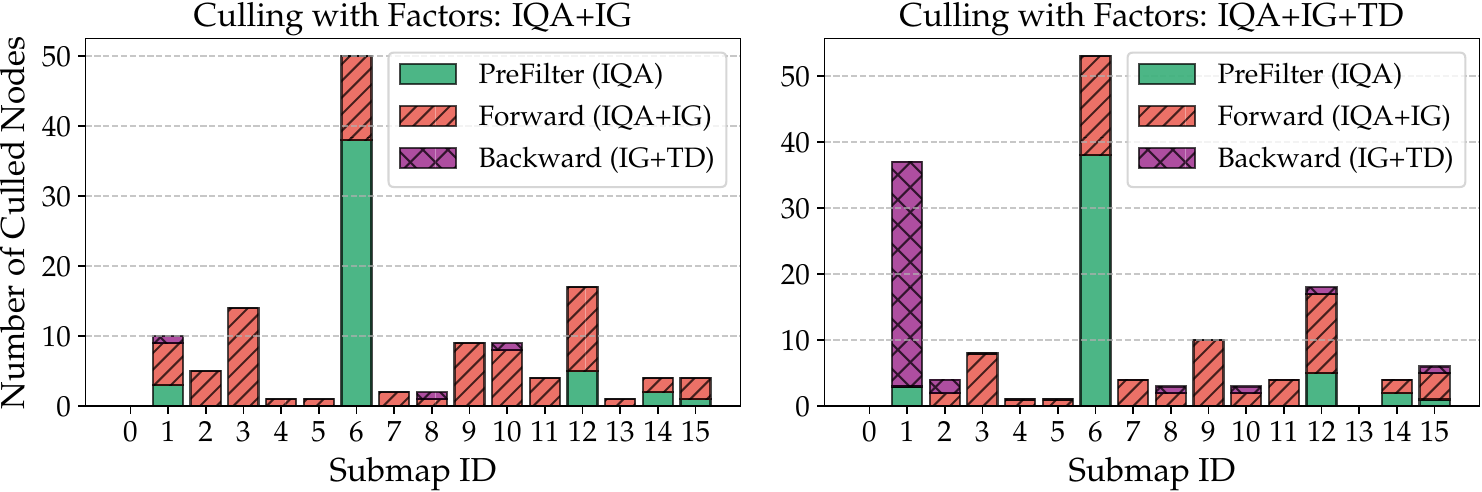}
  }
  \subfigure[Map without node culling.]{
    \label{fig:exp_node_culling_map_merging_node_operation}
    \includegraphics[width=0.485\linewidth]{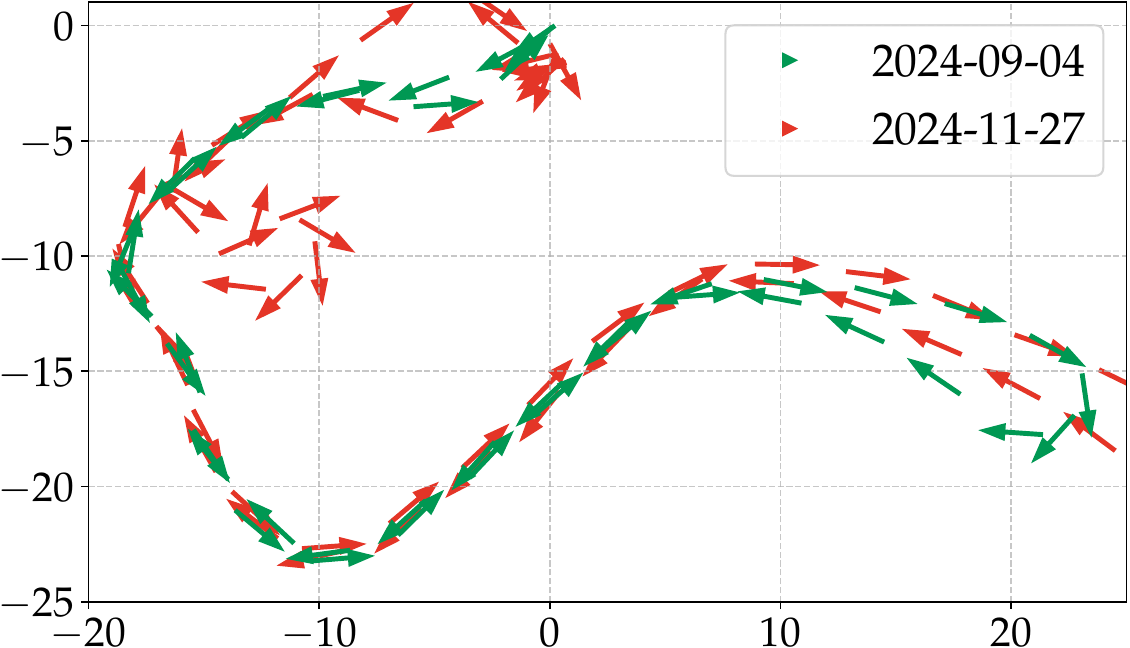}
  }
  \hspace{-0.05\linewidth}
  \subfigure[Map after node culling.]{
    \label{fig:exp_node_culling_map_merging_node_operation}
    \includegraphics[width=0.485\linewidth]{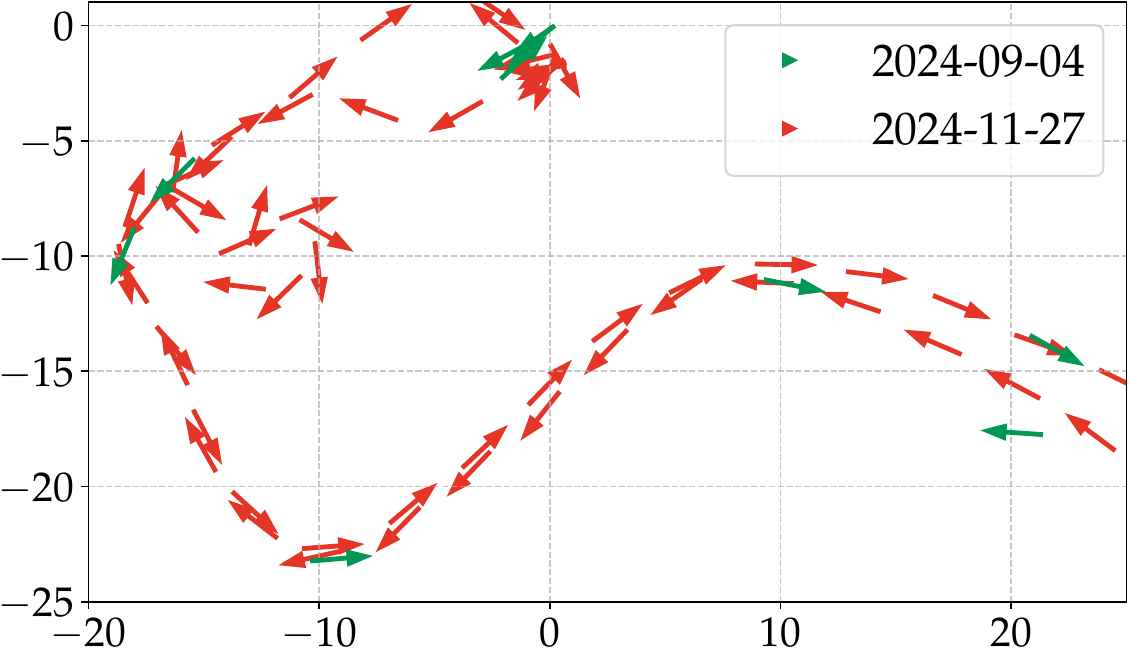}
  }
  \caption{\rt{Lifelong operation for the incremental map merging in Lab, where the dataset spans over three months. Nodes built with two-session maps are shown in (c) and (d), where the arrow direction indicates the forward direction.}}
  \label{fig:exp_node_culling_map_merging}
  \vspace{-0.3cm}
\end{figure}

\bt{
This experiment addresses \textbf{Q4}, assessing how node culling balances map size against localization accuracy over time.
We ablate the three IC factors cumulatively from most conservative to most aggressive: IQ alone (removes low-quality nodes), then adding IG (geometrically redundant nodes), then TD (outdated nodes), following the pipeline's sequential staging.
Two downstream tasks are evaluated, map-merging accuracy in terms of ATE and metric localization precision, to verify that culling preserves the map's utility for navigation.}

\subsubsection{\bt{Impact on Map Merging}}
\bt{
    To test node culling under lifelong operation, we merge the Lab sequences collected over three months. We choose these sequences from R$2$ because they revisit the same places under strong lighting and viewpoint changes, which is exactly when culling matters most. We report two metrics: the final ATE and the total node count.
    Tab.~\ref{tab:exp_ablation_node_map_merging} reports the ablation over the culling factors across the pre-filter, forward, and backward stages.
    Using IQ alone yields the minimum ATE, confirming that quality-based filtering effectively removes outlier-prone nodes.
    IG and TD act as more aggressive mechanisms: they incur a slight precision trade-off but enable substantial compression.
    As shown in Fig.~\ref{fig:exp_node_culling_map_merging}, the full strategy reduces map size by roughly $20\%$ with minimal accuracy loss.}

\begin{figure*}[t]
  \centering
  \subfigure[2024/09/04-2024/11/28]
  {\label{fig:exp_path_planning_0}\centering\includegraphics[width=0.1909\linewidth]
    {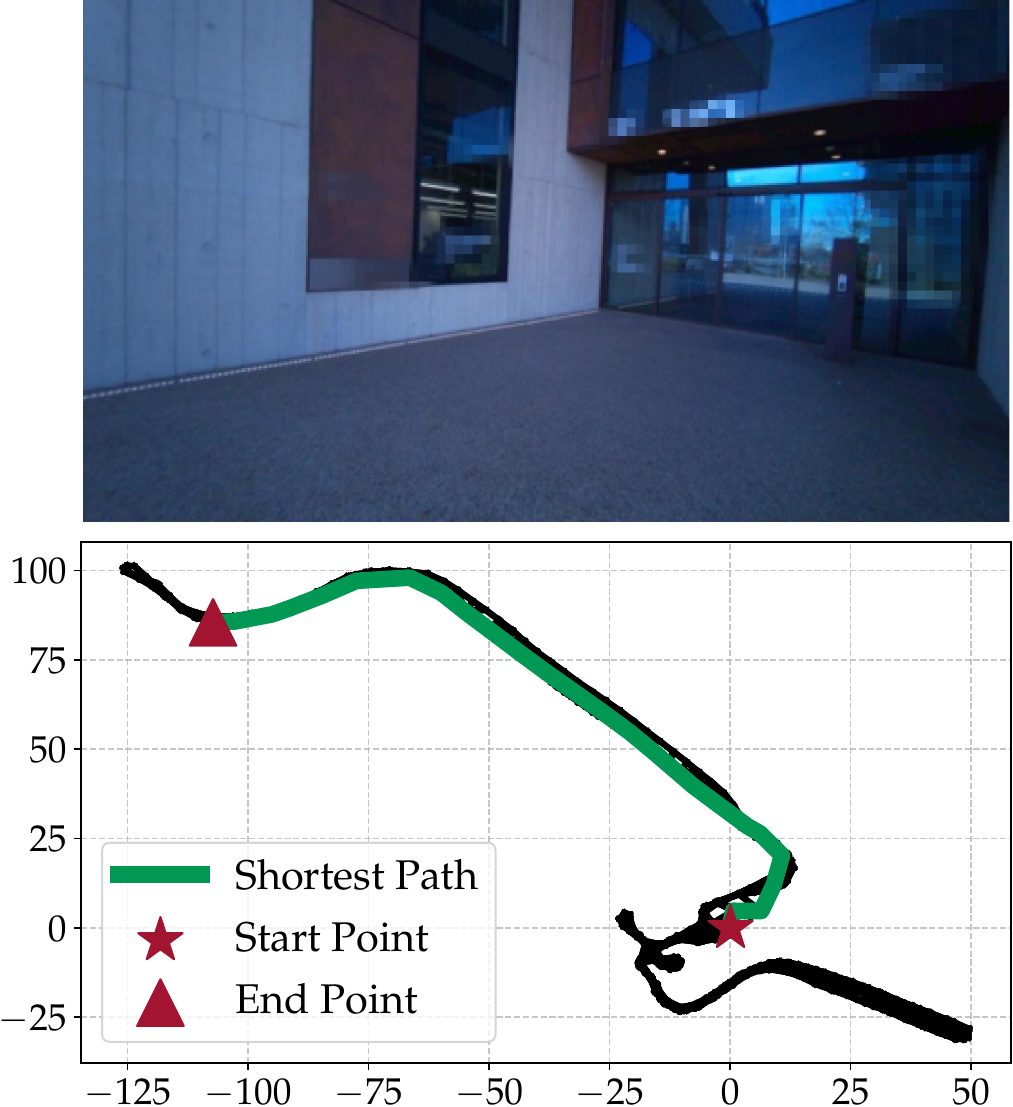}}
  \hspace{-0.17cm}
  \subfigure[2024/09/04-2024/12/04]
  {\label{fig:exp_path_planning_1}\centering\includegraphics[width=0.1880\linewidth]
    {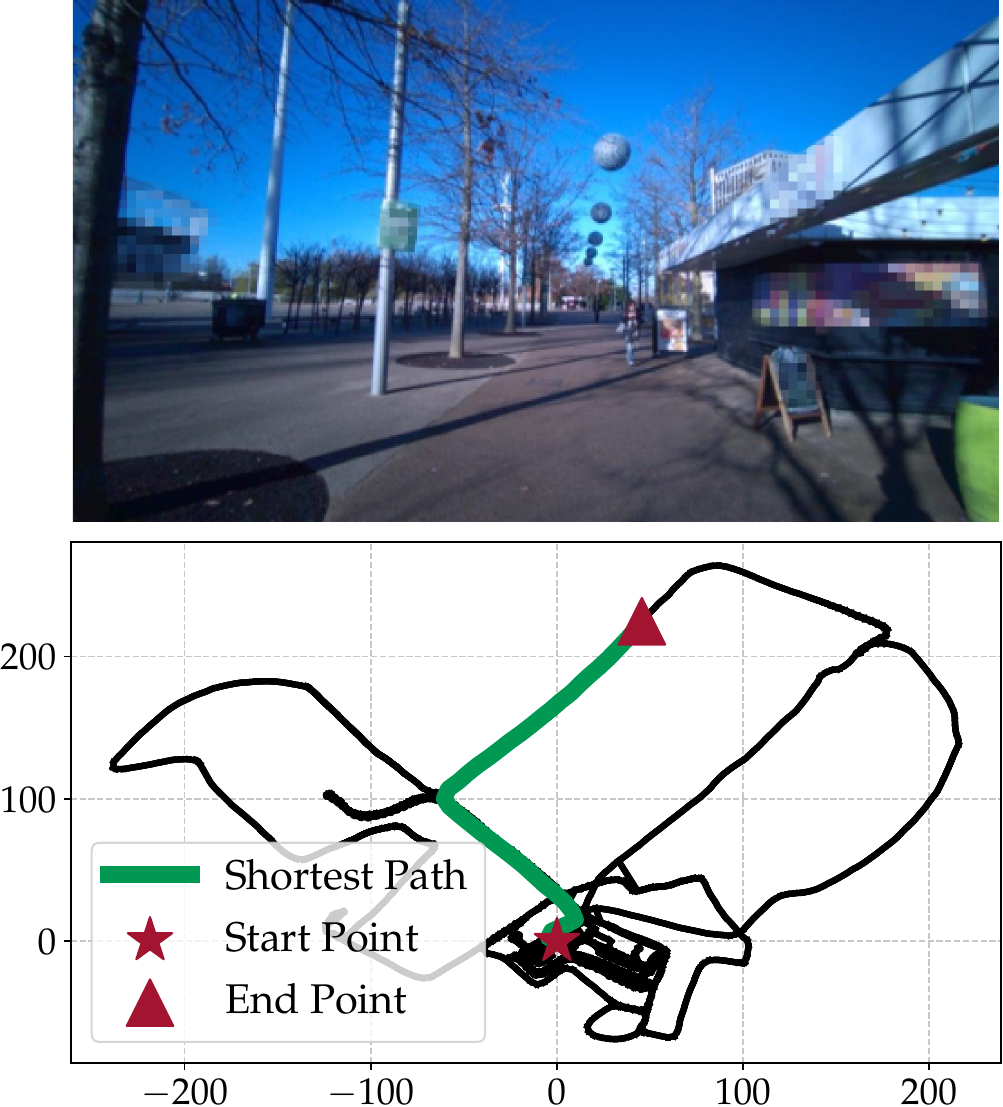}}
  \hspace{-0.17cm}
  \subfigure[2024/09/04-2024/12/23]
  {\label{fig:exp_path_planning_2}\centering\includegraphics[width=0.1928\linewidth]
    {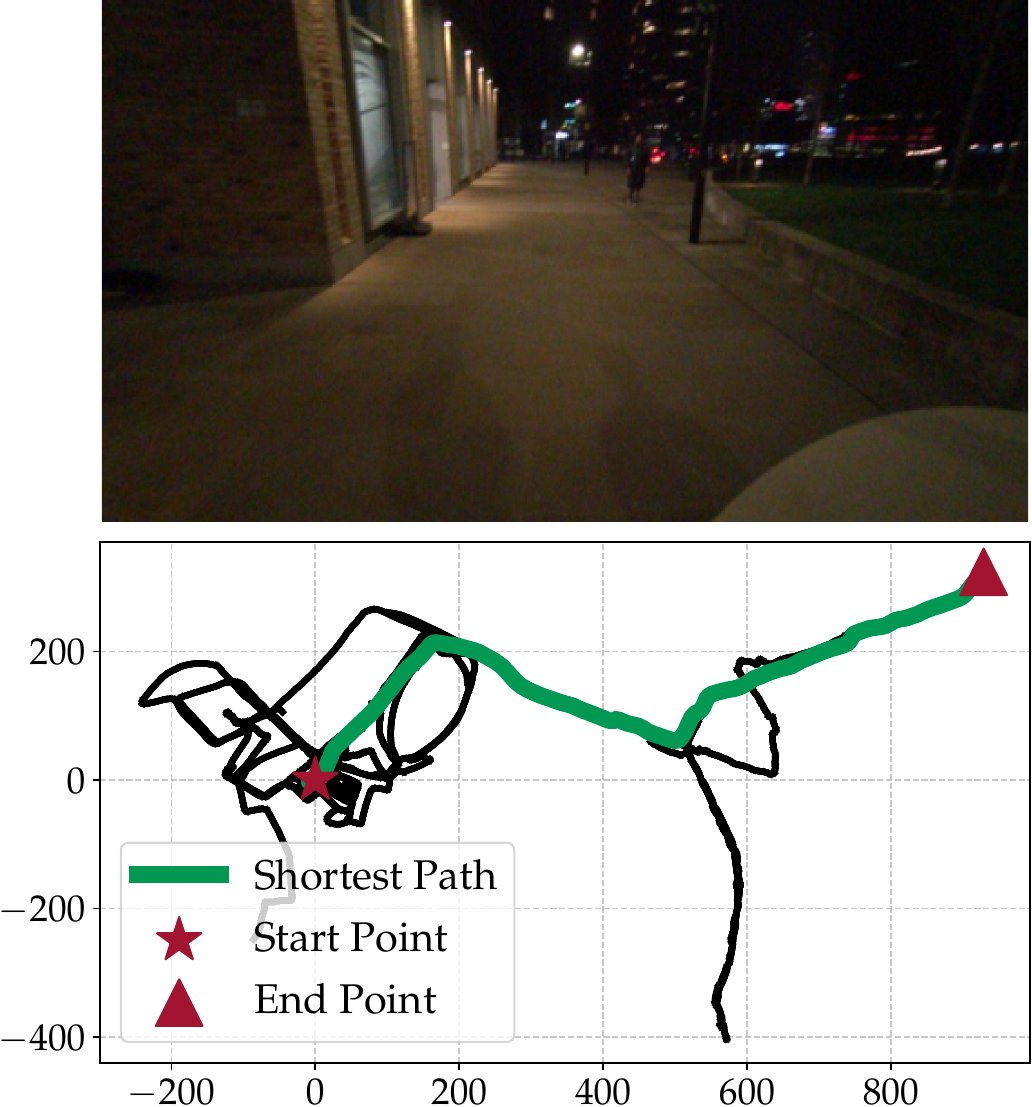}}
  \hspace{-0.17cm}
  \subfigure[2012/04/01-2024/12/23]
  {\label{fig:exp_path_planning_3}\centering\includegraphics[width=0.1959\linewidth]
    {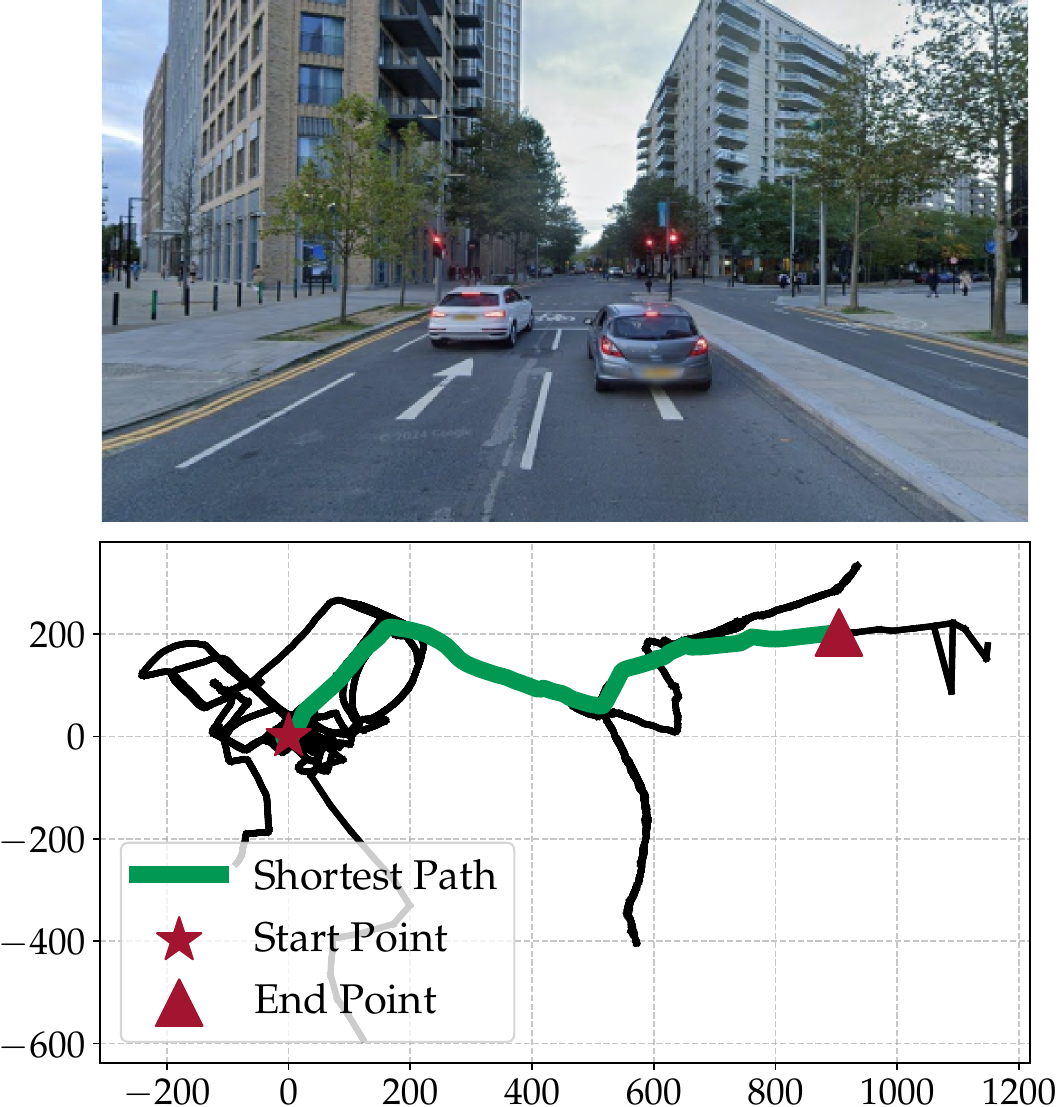}}
  \hspace{-0.17cm}
  \subfigure[2024/09/04-2024/12/04]
  {\label{fig:exp_path_planning_4}\centering\includegraphics[width=0.1832\linewidth]
    {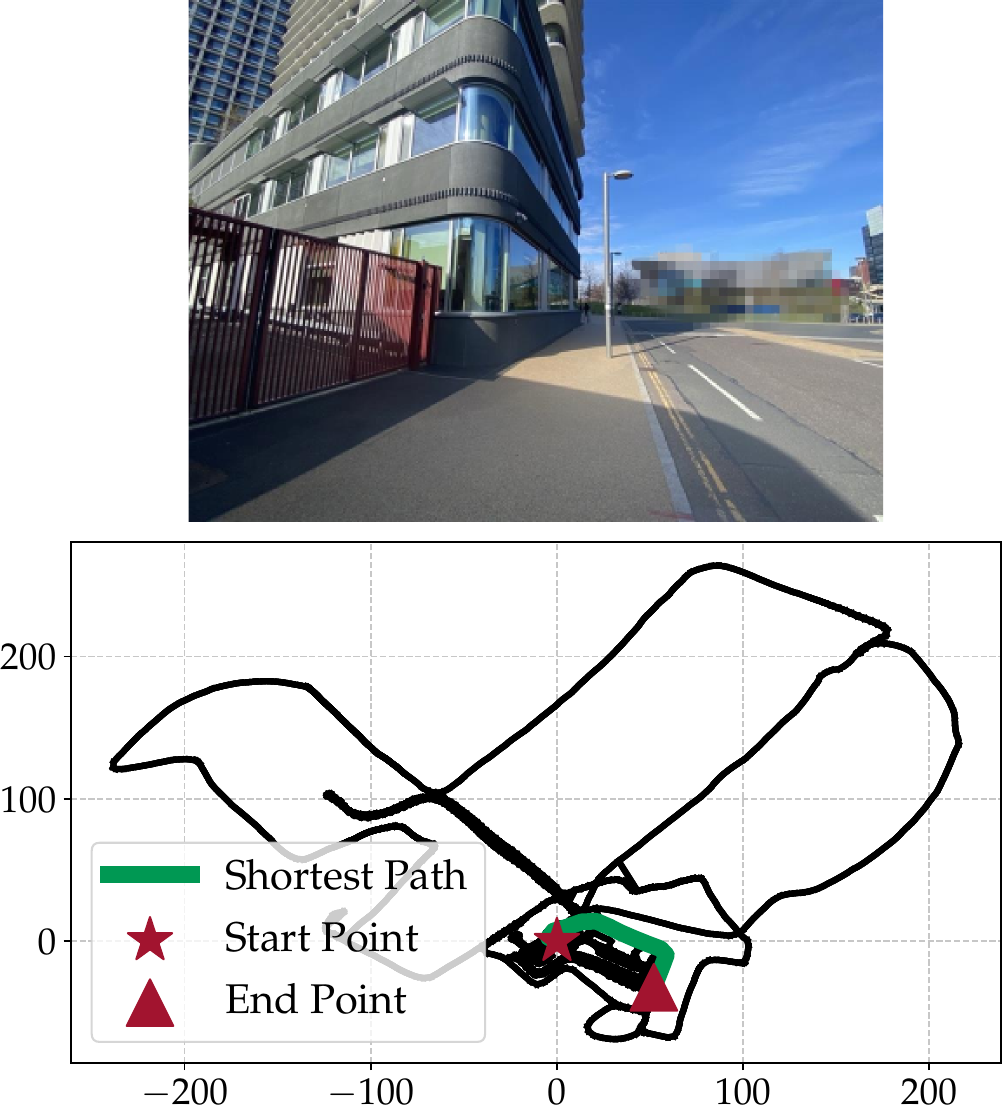}}
  \caption{\rt{Global path planning on the topometric map at different mapping stages (a)--(e). The caption for each subfigure indicates the time span of the accumulated map data. As the map expands with new submaps, shortest-path navigation becomes feasible to an increasing number of destinations. Goal Image Sources: (a)--(c) Aria glasses (frames selected from the raw data collection stream of the last submap), (d) Google Street View, and (e) a smartphone. Note that all paths originate from the same fixed starting point on the map.}}
  \label{fig:exp_path_planning}
  \vspace{-0.3cm}
\end{figure*}

\subsubsection{\bt{Impact on Global Localization}}
Culling is motivated by two goals: bounding map growth and pruning stale or redundant nodes. Yet indiscriminate node removal may discard geometrically critical viewpoints and degrade localization. We therefore assess how each culling strategy affects metric localization accuracy.
\bt{The IQ and TD factors are omitted here, as both vary little across the Map-Free and GZ-Campus datasets. We compare three strategies, the last being ours:}

\begin{itemize}
    \item \textbf{Pose Density}: A node is culled if its translational and rotational distances to the nearest existing node are below set thresholds. This method is computationally efficient but ignores all visual content.
    \item \textbf{2D Feature Matching}: A node is culled if it shares sufficient 2D inlier LoFTR feature correspondences \cite{sun2021loftr} with its visually closest neighbor.
    \item \textbf{3D Pointmap Overlap (Ours)}: A node is culled due to insufficient information gain relative to its visually most similar neighbor, as detailed in Sec.~\ref{sec:method_data_pruning}.
\end{itemize}

%%%%%%%%%%%%%%%%%%%%%%%%%%%%%%%%%%%%%%%%%%%%%%%%%%%%%%%%%%
\begin{table}[]
  \centering
  \caption{Pose estimation performance within the reference map.}
  \renewcommand\arraystretch{1.0}
  \renewcommand\tabcolsep{2.5pt}
  \scriptsize
  \begin{tabular}{c|c|c|c|c}
    \toprule[0.03cm]
    \multicolumn{5}{c}{\textbf{Map-Free Dataset} ($N_{Query}=7369$)}                                                                       \\
    \toprule[0.03cm]
    \textbf{Strategy} & \textbf{\makecell[c]{Node                                                                                          \\Number}}
                      & \textbf{\makecell[c]{Maximum                                                                                       \\$E_t[m]/E_r[deg]\downarrow$}}  & \textbf{\makecell[c]{Avg. Median \\$E_{t}[m]/E_{r}[deg]\downarrow$}}
                      & \makecell[c]{\textbf{Precision $[\%]$}                                                                             \\$[\mathbf{25}cm,\mathbf{5}^\circ]\uparrow$}      \\
    \midrule[0.03cm]
    Full KF Set       & $29629$                                & $0.36/6.81$           & $0.006/0.15$                   & $99.77$          \\
    \midrule[0.03cm]
    Pose Density      & $3691$                                 & $0.91/16.08$          & $0.018/\mathbf{0.28}$          & $96.40$          \\
    2D Feature        & $3803$                                 & $0.67/\mathbf{11.65}$ & $0.019/0.30$                   & $96.93$          \\
    3D Pointmap       & $3961$                                 & $\mathbf{0.61}/11.91$ & $\mathbf{0.017}/\mathbf{0.28}$ & $\mathbf{97.50}$ \\
    \midrule[0.03cm]

    \multicolumn{5}{c}{\textbf{GZ Campus Dataset} ($N_{Query}=476$)}                                                                       \\
    \toprule[0.03cm]
    \textbf{Strategy} & \textbf{\makecell[c]{Node                                                                                          \\Number}}
                      & \textbf{\makecell[c]{Maximum                                                                                       \\$E_t[m]/E_r[deg]\downarrow$}}
                      & \textbf{\makecell[c]{Avg. Median                                                                                   \\$E_{t}[m]/E_{r}[deg]\downarrow$}}
                      & \makecell[c]{\textbf{Precision $[\%]$}                                                                             \\$[\mathbf{25}cm,\mathbf{5}^\circ]\uparrow$}      \\
    \midrule[0.03cm]
    Full KF Set       & $1451$                                 & $1.47/1.65$           & $0.19/0.59$                    & $69.54$          \\
    \midrule[0.03cm]
    Pose Density      & $354$                                  & $2.05/4.16$           & $0.21/0.74$                    & $62.82$          \\
    2D Feature        & $720$                                  & $\mathbf{1.85}/2.26$  & $\mathbf{0.19}/\mathbf{0.64}$  & $\mathbf{69.12}$ \\
    3D Pointmap       & $720$                                  & $1.86/\mathbf{2.25}$  & $\mathbf{0.19}/\mathbf{0.64}$  & $\mathbf{69.12}$ \\
    \bottomrule[0.03cm]
  \end{tabular}
  \label{tab:pose_error_evaluation}
  \vspace{-0.3cm}
\end{table}
%%%%%%%%%%%%%%%%%%%%%%%%%%%%%%%%%%%%%%%%%%%%%%%%%%%%%%%%%%

For each scene, every strategy retains $75\%$ of the images as the reference map and uses the remaining $25\%$ as query frames.
Results, reported as Precision@$[25cm, 5^{\circ}]$, are summarized in Tab.~\ref{tab:pose_error_evaluation}.
\bt{Pose Density yields the strongest compression, cutting storage by $87.5\%$ (Map-Free) and $75.6\%$ (GZ-Campus), but its reduced viewpoint coverage raises the maximum positional error by $0.55m$.
    2D Feature Matching and ours reach comparable compression while attaining higher accuracy and smaller maximum errors, confirming that visually or geometrically informed culling preserves localization-critical viewpoints.
    Ours is thus adopted as it directly reuses the feed-forward pointmap prediction from the hierarchical localization pipeline.}
Still, any culling strategy may remove a viewpoint that is observed only rarely, leaving certain query frames with too little coverage and raising their localization error. On resource-constrained robots, an adaptive culling rate could trade a slightly larger map for more reliable localization.

%%%%%%%%%%%%%%%%%%%%%%%%%%%%%%%%%%%%%%%%%%%%%%%%
%%%%%%%%%%%%%%%%%%%%%%%%%%%%%%%%%%%%%%%%%%%%%%%%
%%%%%%%%%%%%%%%%%%%%%%%%%%%%%%%%%%%%%%%%%%%%%%%%

\subsection{\bt{Exp 5: Multi-Session Mapping for Navigation}}
\label{sec:experiment_vnav}

\bt{
This section mainly addresses \textbf{Q5}. It first quantifies how multi-session merging extends and updates the navigable space and shortens planned paths toward the goal, and then demonstrates that the resulting topometric map supports reliable image-goal visual navigation.}
% The evaluation proceeds from component to system level: we quantify the benefit of multi-session merging, assess full-system performance in simulation, and validate real-world deployment.}

% \input{doc/experiment_vloc}

\subsubsection{\bt{Benefits of Multi-Session Map Merging}}
\label{sec:exp_single_vs_multi}

\begin{figure}[t]
  \centering
  \includegraphics[width=0.99\linewidth]{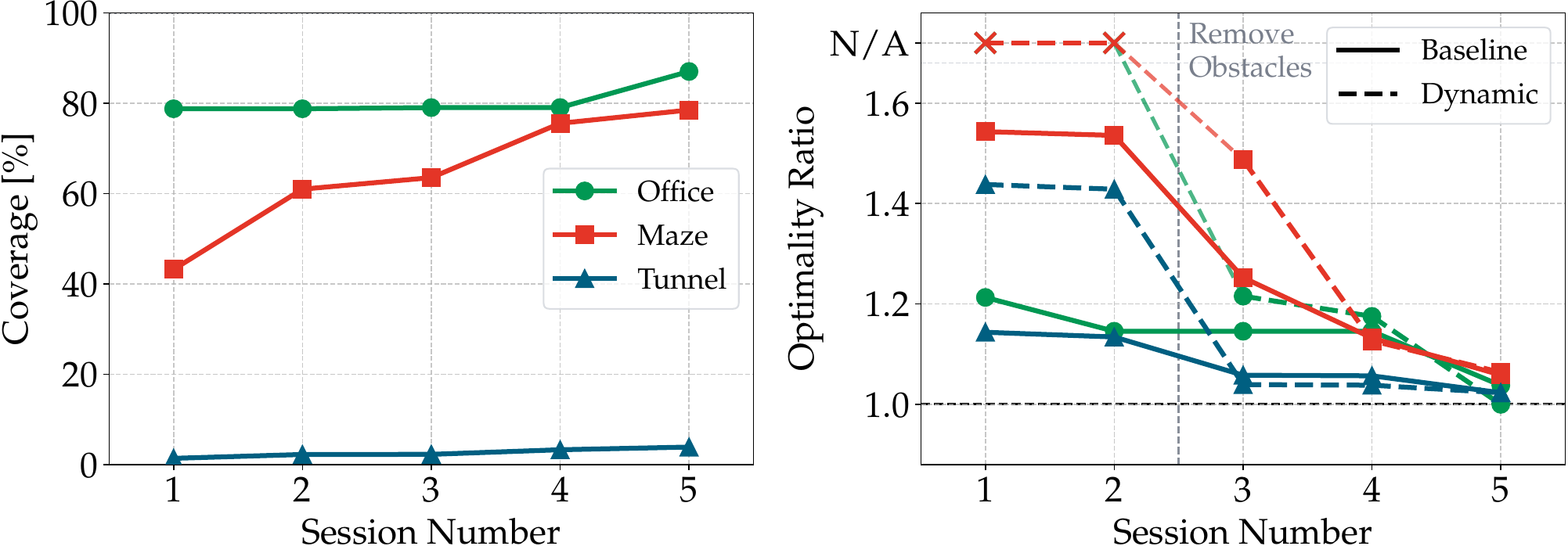}
  \caption{\bt{Multi-session mapping on three simulated environments of increasing scale: \textit{Office} ($432$m$^2$), \textit{Maze} ($1239$m$^2$), and \textit{Tunnel} ($82502$m$^2$). Across sessions, we report cumulative reachability and the topometric shortest-path ratio $r = \ell_{\text{topo}} / \ell_{\text{GT}}$ for the normal and the dynamic variant, where an obstacle blocks the route during sessions~$0$-$1$ and is removed afterward. While blocked, single-session maps either cannot reach the goal ($r = \infty$) or reach it only via a long detour (large $r$); merging the post-removal sessions restores direct reachability with $r$ close to $1$, showing that cumulative merging adapts to environmental change.}}
  \label{fig:exp_multisession_nav}
  \vspace{-0.3cm}
\end{figure}

\bt{
Merging multiple sessions improves navigation through two complementary mechanisms.
First, the metric coordinates of nodes enforce global metric consistency across all merged submaps.
Without it, inconsistent per-submap scale estimates mislead the distance-based planner, as a physically longer route may receive a smaller estimated length and be wrongly preferred.
% global consistency removes this error.
Second, each additional session expands the traversability graph with new routes, extending the planning horizon to goals unreachable from any single-session map.
We validate both mechanisms in three simulated environments, \textit{Office}, \textit{Maze}, and \textit{Tunnel}~\cite{zhang2024falcon,cao2022autonomous}.
In each environment, an agent driven by a frontier-based exploration strategy autonomously collects several mapping sessions; small randomization of its exploration yields a distinct trajectory per session, emulating independent crowdsourced traversals.
As sessions accumulate, merging incrementally raises reachability and improves the topometric shortest-path ratio (Fig.~\ref{fig:exp_multisession_nav}), defined as the planned path length over the GT shortest-path length.
In the dynamic variant, an obstacle blocks the route during sessions~$0$-$1$ and is removed from session~$2$ onward: single-session maps either cannot reach the goal or reach it only via a long detour, whereas the merged map recovers direct reachability and re-plans through the previously blocked path.
More details are in the \textbf{Suppl. Mat.}~\cite{anonymous2025opennavmap_supp}.}

\subsubsection{\bt{Full Simulated Navigation}}
\label{sec:exp_simu_vnav}
\bt{In simulation, the robot is sequentially assigned a series of image goals to reach.
Tab.~\ref{tab:experiment_vnav_simu} compares our system against a SoTA LiDAR-based navigation system \cite{zhang2020falco}, which pairs GT localization with Falco as the local planner.
Fig.~\ref{fig:exp_simu_vnav} visualizes the navigation route in Env$1$, together with the topometric map and the planned path.
Because the camera's limited field of view restricts how much of the surroundings the planner can observe at once, the robot generates suboptimal local paths and must rotate frequently to perceive more of the scene.
This lengthens the time to reach the goal but leaves the total path length largely unchanged, confirming that visual localization stays robust and accurate throughout the task.
Local planning also remains effective for collision avoidance without a pre-built dense map.}

%%%%%%%%%%%%%%%%%%%%%%%%%%%%%%%%%%%%%%%%%%%%%%%%%%%%%%%%%%
\begin{table}[t]
      \centering
      \caption{Navigation performance in simulated environments.}
      \renewcommand\arraystretch{1.0}
      \renewcommand\tabcolsep{4.0pt}
      \scriptsize
      \begin{tabular}{c|c|ccc}
            \toprule[0.03cm]
                                                   &                                    & \multicolumn{3}{c}{Navigation Time $[$s$]$ and Path Length $[$m$]$}                                                                                     \\
            \midrule
            \multirow{1}{*}{\textbf{Localization}} & \multirow{1}{*}{\textbf{Planning}} & \multicolumn{1}{c}{Env$0$ \rt{[home]}}                          & \multicolumn{1}{c}{Env$1$ \rt{[home]}} & \multicolumn{1}{c}{Env$2$ \rt{[office]}} \\

            \midrule[0.03cm]
            GT                                     & Falco                              & $40.3s/25.5m$                                                   & $42.7s/28.3m$                          & $266.9s/220.7m$                          \\
            Ours                                   & iPlanner                           & $50.0s/28.6m$                                                   & $52.9s/29.1m$                          & $280.3s/223.7m$                          \\
            \bottomrule[0.03cm]
      \end{tabular}
      \label{tab:experiment_vnav_simu}
      % \vspace{-0.3cm}
\end{table}
%%%%%%%%%%%%%%%%%%%%%%%%%%%%%%%%%%%%%%%%%%%%%%%%%%%%%%%%%%

\begin{figure}[t]
  \centering
  \includegraphics[width=0.95\linewidth]{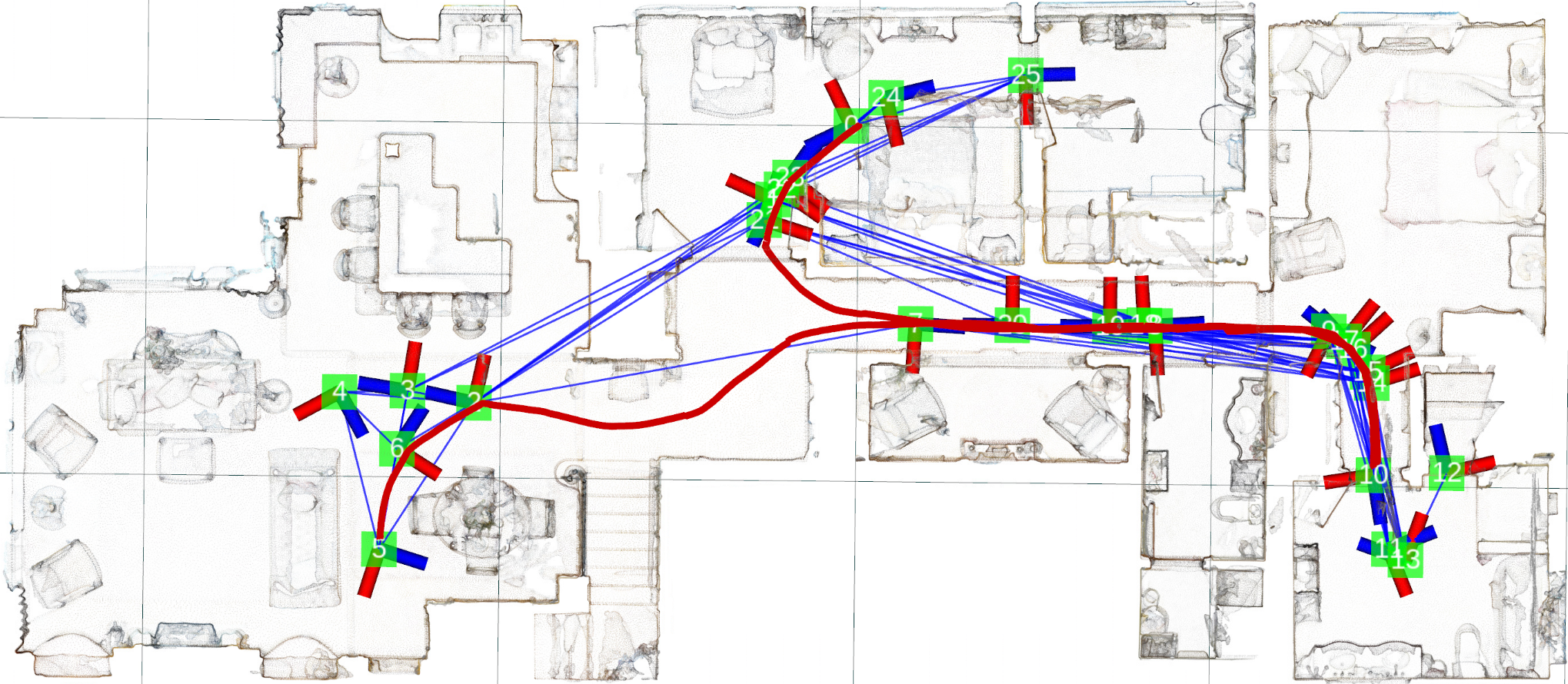}
  \caption{The \bt{visual navigation} system is evaluated in a simulated home environment (Env$1$), where the robot is tasked with sequentially navigating to a series of image goals. The traversed path is visualized as a \rt{red} trajectory, while the underlying topometric map is represented as a graph structure: nodes (\gt{green} squares) are connected by traversability edges (\bt{blue} lines).}
  \label{fig:exp_simu_vnav}
  \vspace{-0.3cm}
\end{figure}

\subsubsection{\bt{Real-World Navigation}}
\label{sec:exp_path_planning}
\bt{We evaluate the complete navigation system, comprising visual localization against the topometric map, global path planning on the traversability graph, and local motion planning, in three real-world environments:}
\textbf{Lab}, a hybrid indoor-outdoor laboratory setting;
\textbf{Bridge}, an outdoor route between two buildings that includes a bridge crossing; and
\textbf{Building}, a building-perimeter traversal with numerous turns.
The robot can only navigate inside the map, so we place it at an arbitrary position with an unknown starting pose and start autonomous navigation.
Each subsequent goal image is sent once the robot reaches the current goal.

\begin{figure*}[t]
  \centering
  \subfigure[First run]{\label{fig:exp_vnav_ops_lab_1}\centering\includegraphics[width=0.415\linewidth]{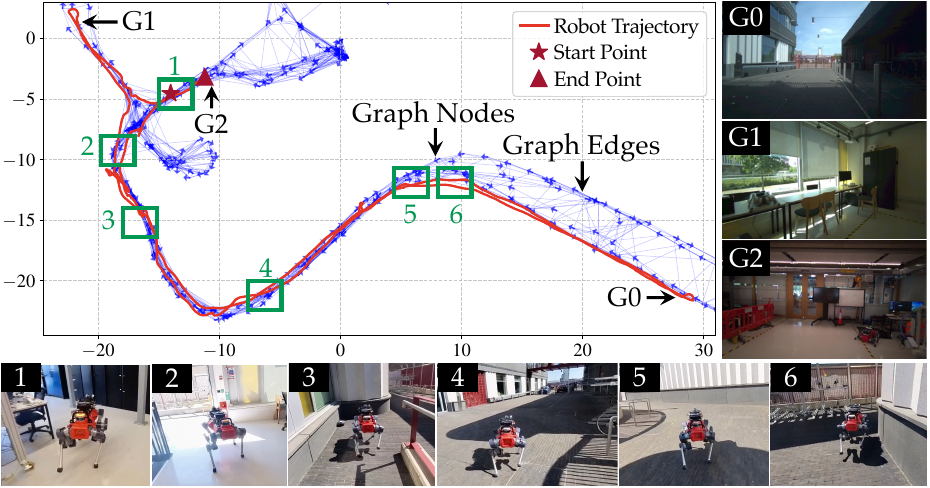}}
  \hspace{-0.2cm}
  \subfigure[Multiple runs at different times]{\label{fig:exp_vnav_ops_lab_multirun}\centering\includegraphics[width=0.555\linewidth]{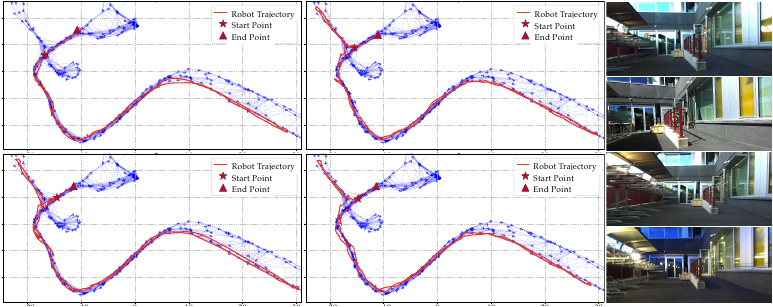}}
  \vspace{-0.3cm}
  \caption{
    \rt{Real-world experiment (Lab) with a quadruped robot. The topometric map structure comprises nodes (blue arrows) and traversable edges (blue lines). Global planning generates a shortest path on this map to provide subgoals for subsequent motion planning.
      The \rt{red} curve denotes the robot's trajectory estimated using our localization method.}
    (a) The robot is sequentially given a set of goal images. It begins inside a room, proceeds outdoors, follows a circular route, and returns to the starting location, covering a total distance of $160m$ in $312s$ at an average speed of $0.5m/s$. \gt{Green} boxes labeled $1$-$6$ highlight key planning events: ($1$ and $2$) lab areas, ($3$) a narrow passageway, ($4$ and $5$) outdoor paths exhibiting varying sunlight conditions, and ($6$) a right turn facing a wall.
    (b) The robot is instructed to repeat the navigation task at different times of day, ranging from daytime to nighttime.}
  \label{fig:exp_vnav_ops_lab}
  \vspace{-0.35cm}
\end{figure*}

\begin{figure}[t]
  \begin{center}
    \includegraphics[width=0.95\linewidth]{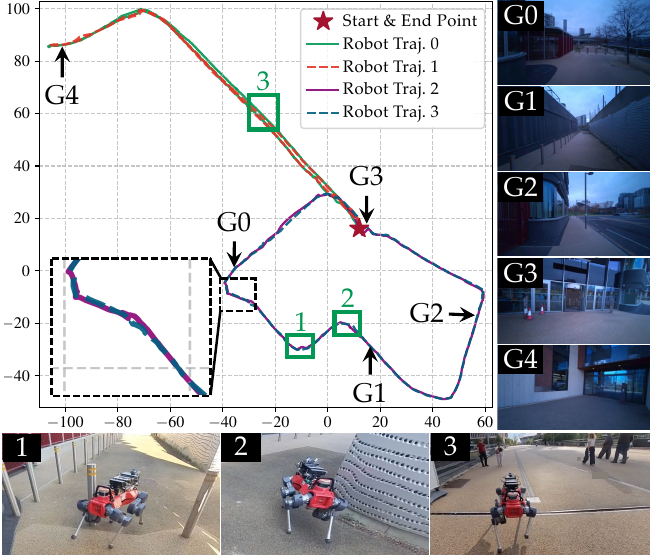}
  \end{center}
  \vspace{-0.25cm}
  \caption{\rt{Real-world experiments (Bridge and Building) demonstrate the robot's capability to autonomously navigate various regions, including a bridge and confined pathways with several obstacles. The robot is given five goal images as visual objectives for its navigation system in Buliding. Some traverses present shadow-light transition.}}
  \label{fig:exp_vnav_ops_around}
  \vspace{-0.4cm}
\end{figure}

In Lab, the robot executed three sequential image-goal tasks.
As shown in Fig.~\ref{fig:exp_vnav_ops_lab}, the system tracked its pose reliably despite varying outdoor lighting, and local perception and planning enabled obstacle avoidance, though the absence of a local map occasionally caused zig-zag motions near persistent obstacles such as walls.
Repeating the experiment four more times at different times of day, as shown in Fig.~\ref{fig:exp_vnav_ops_lab_1}, the robot navigated from the indoor lab to an outdoor path, autonomously passing through narrow doorways.
\bt{Trajectories along the same route, estimated by our visual localization module, stayed consistent even under nighttime illumination changes.
The $160$m route was completed in roughly $312$s (average speed $0.5$m/s), indicating robustness to sensor noise across indoor and outdoor lighting.}

In Bridge and Building (Fig.~\ref{fig:exp_vnav_ops_around}), we further challenged the system.
Bridge required traversing a span with sparse visual features.
\bt{
% Here, MASt3R was critical for reliable feature matching and relative pose estimation against the global map.
The low-rate global pose estimate is fused with high-rate local odometry to produce a high-frequency pose output, ensuring accurate navigation to the goal.}
Building involved complex turns demanding accurate localization for timely maneuvering, where odometry alone would accumulate drift and cause incorrect decisions.
The supplementary video shows the robot's behavior in these environments.

\section{Discussion}
\label{sec:discussion}

\subsection{Main Advantages}
\label{sec:discussion_advantages}

\bt{\methodname is a lightweight, scalable, appearance-based topometric mapping system: each node stores an RGB image, and edges encode covisibility, odometry, and traversability relationships supporting metric localization and path planning.
Its primary strengths are \ti{cross-device multi-session mapping} (fusing heterogeneous-device submaps across times and viewpoints), \ti{robust hierarchical localization} (coarse-to-fine pipeline resilient to viewpoint, sparsity, and appearance changes), and \ti{multi-modal extensibility} to goals beyond images, such as text or GPS.
Unlike conventional topological maps \cite{Dang2020,shah2023gnm} restricted to path planning, \methodname unifies localization and planning in a single topometric framework, with potential for scalable real-world deployment.}
% A current limitation is the reliance on a computationally intensive 3D GFM for online pointmap generation and optimization. Improving inference speed (\eg through network quantization and pruning) is therefore necessary to raise computational efficiency and broaden practical applicability.

\subsection{Limitations and Future Works}
\label{sec:discussion_limitations}
We further indicate several directions for future research beyond the proposed \methodname:
\begin{itemize}[leftmargin=15pt]
      \item \textit{Sequence-Based Matching:}
            \rt{Our DP-based sequence matching relies on traversal sequences for indexing rather than graph topology, limiting flexibility, and exhaustive searches become inefficient at scale. Future research should explore graph-based or location-indexed strategies exploiting graph topology for more flexible and efficient matching (\eg of nearby, non-sequential images).}

      \item \textit{Generalization of 3D Foundation Models:}
            \rt{Performance of 3D GFMs degrades under cross-device variations, illumination shifts, and appearance changes. As noted in~\cite{blum2025crocodl}, the cross-device domain gap is more challenging than single-device scenarios. Developing increasingly generalizable 3D GFMs \cite{wang2025vggt} thus remains an important direction.}

      \item \textit{Centralized Multi-Session Mapping:}
            The current centralized framework faces computational, storage, and communication bottlenecks at large scale.
            Decentralized frameworks using Gaussian belief propagation and distributed optimization could significantly enhance scalability and real-time adaptability.

      \item \textit{Conservative Node Culling:}
            Our culling strategy is conservative; ideally, retained nodes should scale with environmental complexity rather than traversal frequency. More aggressive adaptive methods that reduce redundancy without compromising accuracy are worth exploring.

      \item \textit{Navigation and Out-of-Map Recovery:}
            \bt{The system assumes operation within known map boundaries, lacking recovery for out-of-map scenarios (\eg robot kidnapping). Integrating autonomous exploration and active SLAM would enable recovery from incorrect initialization, improving autonomy and robustness.}
\end{itemize}

\section{Conclusion}
\label{sec:conclusion}

% This paper presented \methodname, a visual collaborative localization and multi-session mapping system tailored for scalable robot navigation.
% \methodname introduces a lightweight, structure-free topometric representation that leverages a 3D GFM within the robust collaborative localization pipeline.
% The introduced GV and CCM modules are demonstrably effective in rejecting false positives.
% We conducted extensive large-scale experiments to evaluate collaborative localization and map merging across diverse environments, ranging from indoor laboratories to urban settings, outperforming typical
% methods like HLoc using structure-based maps under challenging conditions such as large viewpoint changes.

% \rt{
%   Despite of the simplicity of the proposed map, it can support sub-meter accuracy in localization and planning.
%   To demonstrate its utility and scalability, the resulting topometric map was deployed on mobile robots, successfully facilitating $12$ autonomous image-goal navigation trials in both simulated and real-world scenarios.
%   These experiments confirm our central hypothesis: a map constructed from sparse observations can deliver competitive, and often superior, localization performance, while significantly simplifying map construction by obviating the need for a globally consistent 3lD model. To foster further research and collaboration, our code and datasets will be made publicly available.}

\bt{In this paper, we presented \methodname, a vision-based multi-session mapping system for scalable robot navigation that couples a landmark-free topometric representation with a 3D GFM, recovering local geometry on demand without maintaining a globally consistent 3D model. Geometric verification rejects topological false positives, and confidence map calibration suppresses forward-prediction errors of the GFM during metric pose estimation.
Extensive experiments across diverse environments showed that \methodname outperforms landmark-based baselines under large viewpoint variations and retains sub-meter accuracy despite the sparsity of its map. 
Deploying it on mobile robots, we completed $12$ autonomous image-goal navigation trials across simulation and the real world, demonstrating that a sparse, landmark-free map can reliably support downstream visual navigation while matching the visual localization accuracy of globally consistent 3D models and substantially reducing scene construction complexity. Source code and datasets will be released publicly.}
\bibliographystyle{IEEEtran}
\bibliography{bib_abbr,bible}
\endgroup

\clearpage
\onecolumn

\section*{Supplementary Material}
\addcontentsline{toc}{section}{Supplementary Material}
\renewcommand{\thesection}{S\arabic{section}}
\setcounter{section}{0}

This supplementary material provides extended experimental details and dataset specifications that complement the main manuscript above:
\begin{itemize}
  \item Sec.~\ref{sec:s1_dataset_self} describes the construction, ground truth generation, and evaluation protocols for the self-collected dataset (R$0$--R$2$), including topological localization configurations, metric localization sampling, and extended map merging visualizations.
  \item Sec.~\ref{sec:s2_dataset_360loc} details the cross-device adaptation of the $360$Loc benchmark~\cite{huang2024360loc} used to evaluate heterogeneous-device map merging.
  \item Sec.~\ref{sec:s3_nav} provides the full benchmark setup and extended per-session results for the multi-session mapping benefit study (Exp~5 in the main paper), quantifying gains in reachability and path optimality as successive sessions accumulate.
\end{itemize}

\section{Self-Collected Dataset: Construction and Evaluation Protocols}\label{sec:s1_dataset_self}

This section details the construction of datasets used in our experiments, with emphasis on the self-collected dataset.
Throughout, visual place recognition (VPR) denotes the task of identifying revisited locations from image appearance alone; visual-inertial odometry (VIO) denotes pose estimation by fusing camera and inertial measurements; and geometric verification (GV) refers to RANSAC-based filtering that retains only geometrically consistent correspondences.

\subsection{Data Acquisition and Ground Truth Generation}
\begin{figure}
    \centering
    \subfigure[R$0$]{
        \includegraphics[width=0.25\linewidth]{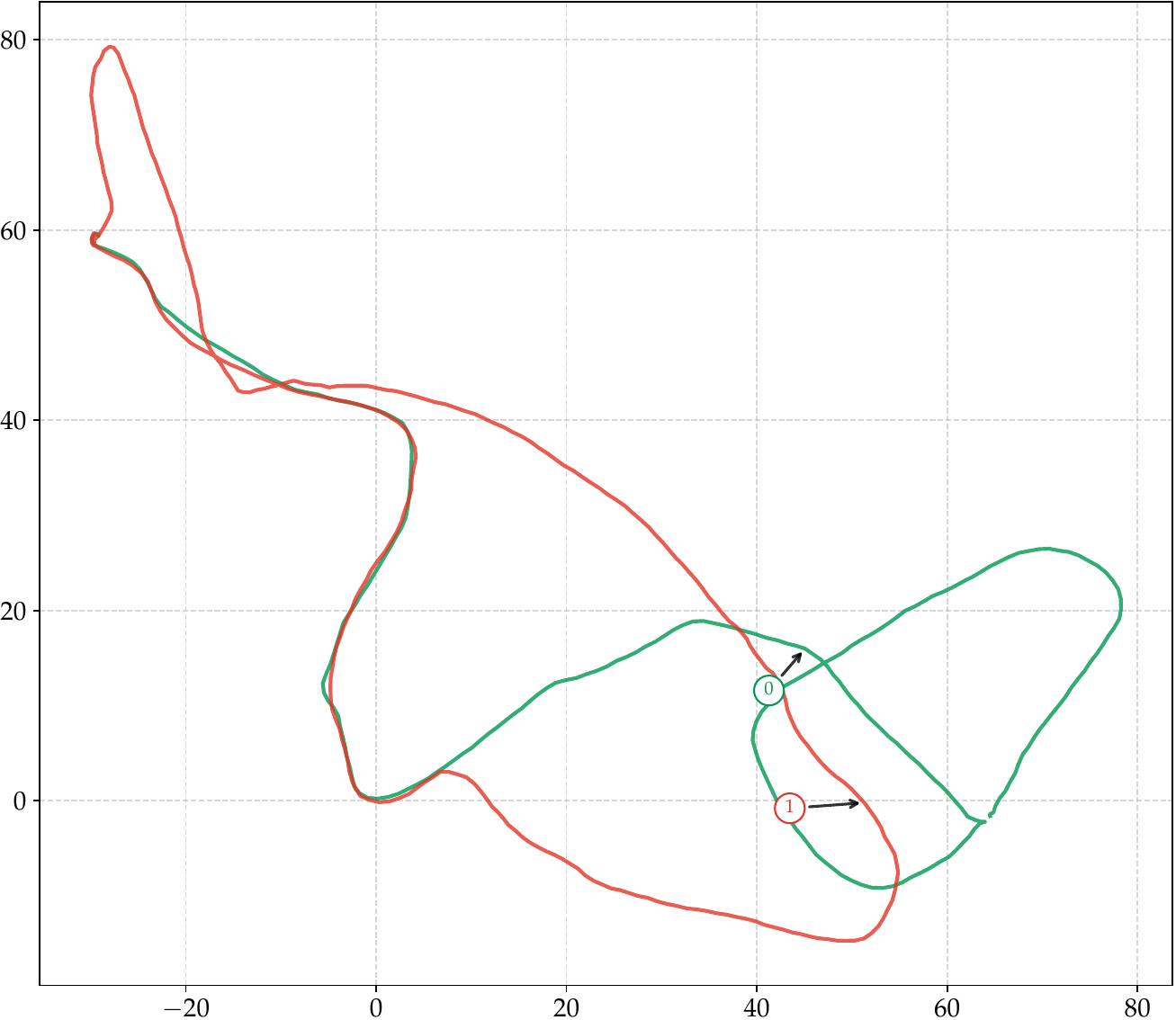}
        \label{fig:s1_aria_traj_r0}
    }
    \hspace{-0.4cm}
    \subfigure[R$1$]{
        \includegraphics[width=0.25\linewidth]{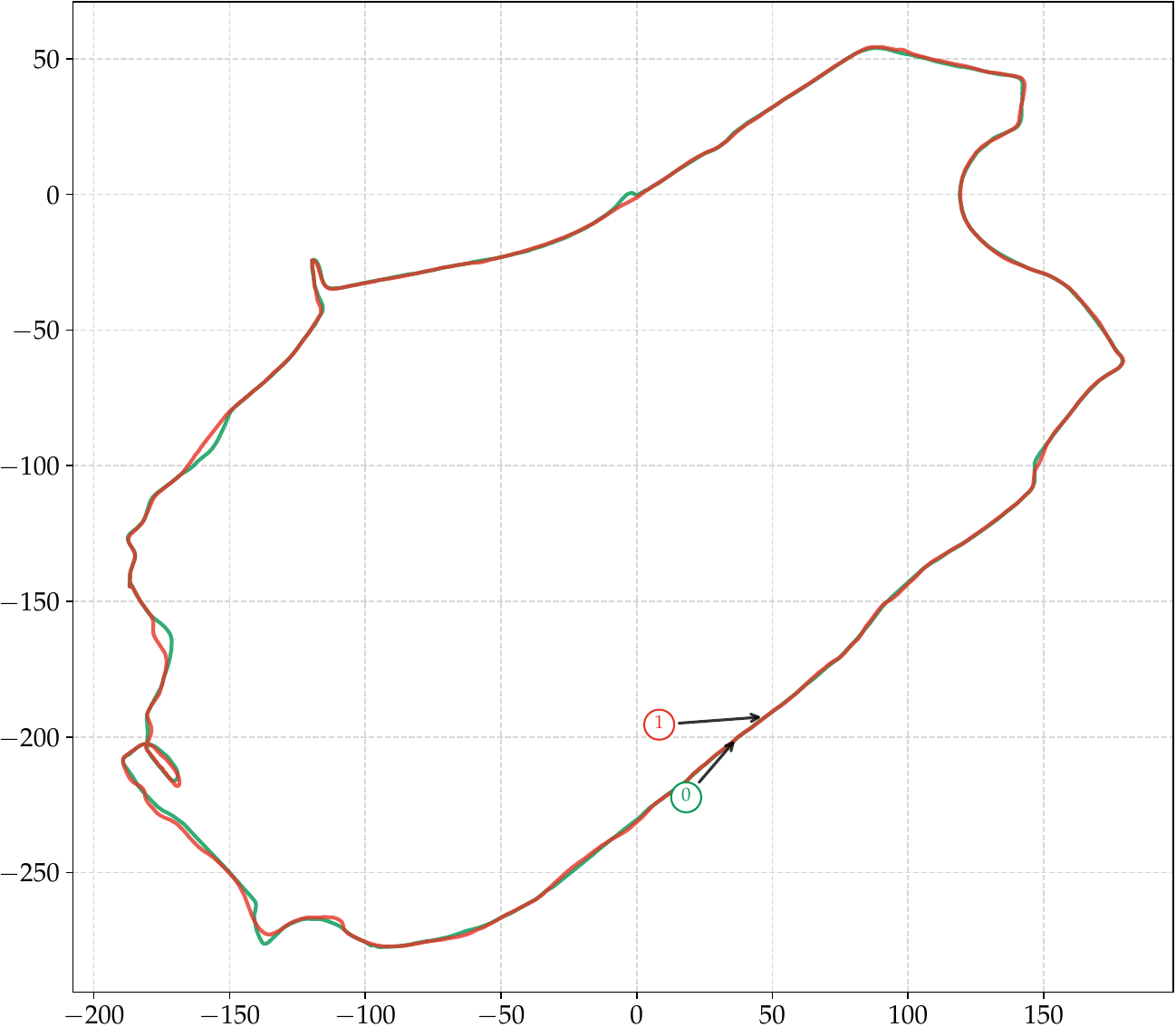}
        \label{fig:s1_aria_traj_r1}
    }
    \hspace{-0.4cm}
    \subfigure[R$2$]{
        \includegraphics[width=0.45\linewidth]{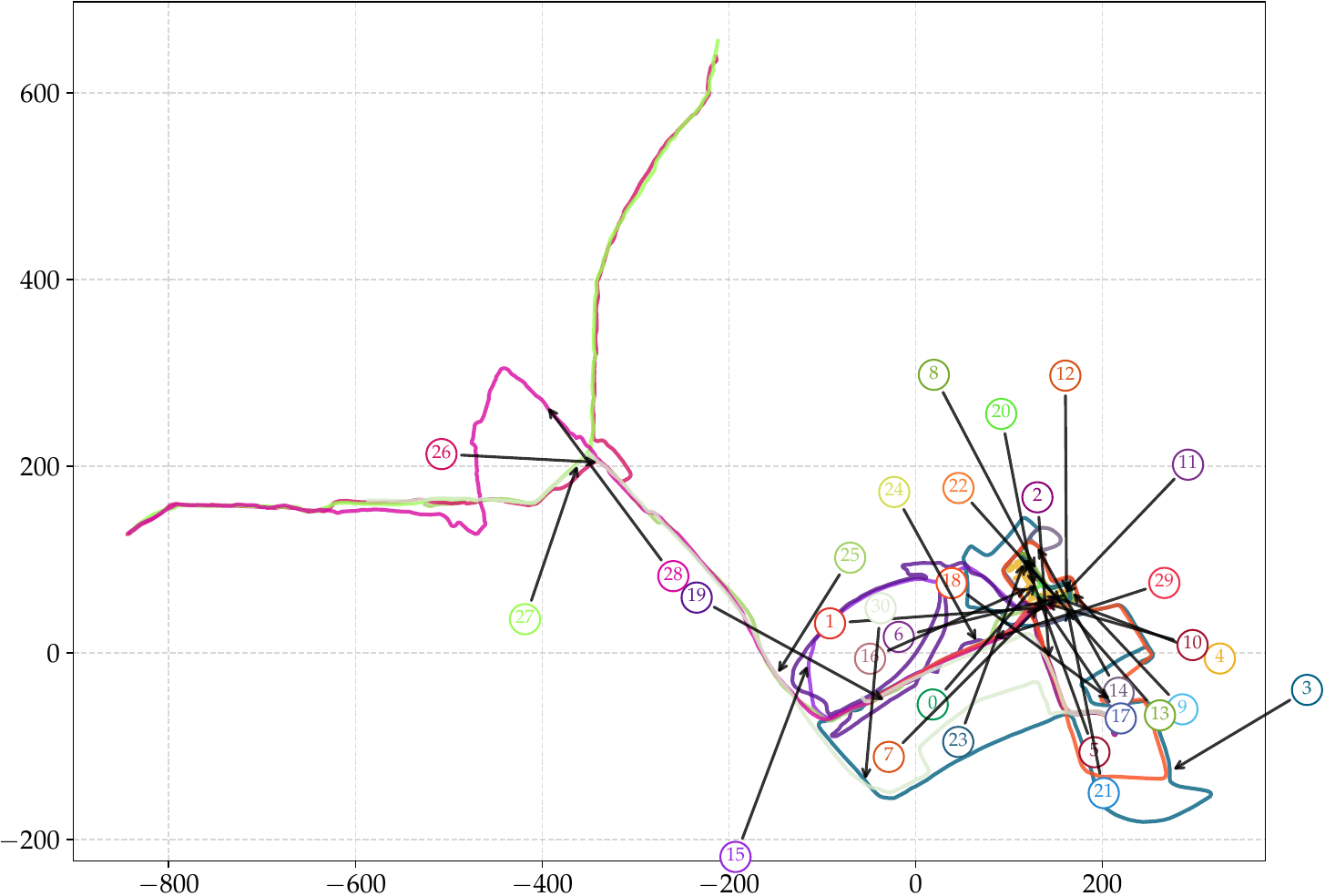}
        \label{fig:s1_aria_traj_r2}
    }
    \caption{Ground truth positions of all sequences from R$0$-R$2$ after multi-session SLAM and global alignment, with sequence IDs corresponding to Tab.~\ref{tab:s1_sequences}. Colored trajectories indicate different sessions; the spatial extents of the three regions span vineyard (R$0$), campus road (R$1$), and mixed building/park/shopping-center (R$2$) environments. Best viewed in color.}
    \label{fig:s1_aria_traj}
    \vspace{-0.3cm}
\end{figure}

As outlined in Sec.~\ref{sec:exp_datasets_self_collected} of the main paper, we collected $37$ sequences across various scenarios using Meta Project Aria glasses~\cite{engel2023project}.
Tab.~\ref{tab:s1_sequences} summarizes the attributes of these sequences, which constitute the R$0$-R$2$ datasets.
Ground truth (GT) trajectories for each region were derived from multi-session SLAM, as illustrated in Fig.~\ref{fig:s1_aria_traj}.

\begin{table}[t]
  \centering
  \caption{Summary of the self-collected dataset (R$0$-R$2$). Each row lists a sequence's device type, collection date, duration, path length, and environment.
  All data were collected with Meta Project Aria glasses.}
  \renewcommand\arraystretch{1.1}
  \renewcommand\tabcolsep{9.0pt}
  % \scriptsize
  \begin{tabular}{c|c|c|c|c|c|c}
    \toprule[0.03cm]
    \textbf{Region} & \textbf{ID} & \textbf{Device}  & \textbf{Date}    & \textbf{Time Spans} $[$min$]$ & \textbf{Length} $[$m$]$ & \textbf{Scenario} \\
    \bottomrule[0.03cm]
    \multirow{2}{*}{R$0$}
                    & 0           & \text{Aria}    & 20241008 - 16:41 & $3.9$                           & $280$  & Vineyard       \\
    \cline{2-7}     & 1           & \text{Aria}    & 20241008 - 16:41 & $3.1$                           & $311$  & Vineyard       \\
    \toprule[0.03cm]
    \bottomrule[0.03cm]
    \multirow{6}{*}{R$1$}
                    & 0           & \text{Aria}    & 20241226 - 19:47 & $14.0$                          & $1222$ & Campus Road    \\
    \cline{2-7}     & 1           & \text{Aria}    & 20241227 - 13:43 & $12.4$                          & $1223$ & Campus Road    \\
    \cline{2-7}     & 2           & \text{Phone}   & 20250117 - 12:00 & $3.3$                           & $169$  & Building       \\
    \cline{2-7}     & 3           & \text{Phone}   & 20250117 - 12:00 & $3.1$                           & $124$  & Building       \\
    \cline{2-7}     & 4           & \text{Phone}   & 20250117 - 12:10 & $3.1$                           & $174$  & Building       \\
    \cline{2-7}     & 5           & \text{Vehicle} & 20230620 - 09:12 & $10.2$                          & $2708$ & Campus Road    \\
    \toprule[0.03cm]
    \bottomrule[0.03cm]
    \multirow{31}{*}{R$2$}
                    & 0           & \text{Aria}    & 20240904 - 08:35 & $2.2$                           & $162$  & Building       \\
    \cline{2-7}
                    & 1           & \text{Aria}    & 20241127 - 17:16 & $10.1$                          & $671$  & Building, Park \\
    \cline{2-7}
                    & 2           & \text{Aria}    & 20241128 - 11:05 & $7.8$                           & $709$  & Building, Park \\
    \cline{2-7}
                    & 3           & \text{Aria}    & 20241129 - 11:45 & $19.5$                          & $1787$ & Park           \\
    \cline{2-7}
                    & 4           & \text{Aria}    & 20241129 - 12:06 & $8.3$                           & $643$  & Building       \\
    \cline{2-7}
                    & 5           & \text{Aria}    & 20241202 - 14:15 & $3.0$                           & $219$  & Building       \\
    \cline{2-7}
                    & 6           & \text{Aria}    & 20241202 - 14:21 & $1.4$                           & $66$   & Building       \\
    \cline{2-7}
                    & 7           & \text{Aria}    & 20241202 - 14:25 & $1.0$                           & $58$   & Building       \\
    \cline{2-7}
                    & 8           & \text{Aria}    & 20241202 - 17:41 & $3.4$                           & $275$  & Building       \\
    \cline{2-7}
                    & 9           & \text{Aria}    & 20241202 - 17:46 & $2.5$                           & $202$  & Building       \\
    \cline{2-7}
                    & 10          & \text{Aria}    & 20241204 - 13:56 & $1.5$                           & $62$   & Building       \\
    \cline{2-7}
                    & 11          & \text{Aria}    & 20241204 - 14:00 & $1.4$                           & $65$   & Building       \\
    \cline{2-7}
                    & 12          & \text{Aria}    & 20241204 - 14:11 & $1.8$                           & $78$   & Building       \\
    \cline{2-7}
                    & 13          & \text{Aria}    & 20241204 - 14:20 & $3.6$                           & $230$  & Building       \\
    \cline{2-7}
                    & 14          & \text{Aria}    & 20241204 - 14:32 & $6.0$                           & $438$  & Building       \\
    \cline{2-7}
                    & 15          & \text{Aria}    & 20241204 - 14:39 & $9.1$                           & $707$  & Park           \\
    \cline{2-7}
                    & 16          & \text{Aria}    & 20241204 - 16:48 & $1.7$                           & $90$   & Building       \\
    \cline{2-7}
                    & 17          & \text{Aria}    & 20241204 - 16:51 & $2.1$                           & $111$  & Building       \\
    \cline{2-7}
                    & 18          & \text{Aria}    & 20241204 - 16:59 & $9.5$                           & $857$  & Park           \\
    \cline{2-7}
                    & 19          & \text{Aria}    & 20241204 - 17:00 & $13.9$                          & $1300$ & Park           \\
    \cline{2-7}
                    & 20          & \text{Aria}    & 20241204 - 17:18 & $3.2$                           & $183$  & Building       \\
    \cline{2-7}
                    & 21          & \text{Aria}    & 20241204 - 18:04 & $1.0$                           & $53$   & Building       \\
    \cline{2-7}
                    & 22          & \text{Aria}    & 20241205 - 10:09 & $1.3$                           & $79$   & Building       \\
    \cline{2-7}
                    & 23          & \text{Aria}    & 20241205 - 10:18 & $3.6$                           & $218$  & Building       \\
    \cline{2-7}
                    & 24          & \text{Aria}    & 20241221 - 17:29 & $3.0$                           & $250$  & Park           \\
    \cline{2-7}
                    & 25          & \text{Aria}    & 20241222 - 12:43 & $10.5$                          & $921$  & Park           \\
    \cline{2-7}
                    & 26          & \text{Aria}    & 20241222 - 15:45 & $12.3$                          & $1125$ & SC, Road       \\
    \cline{2-7}
                    & 27          & \text{Aria}    & 20241222 - 16:37 & $11.7$                          & $1040$ & SC, Road       \\
    \cline{2-7}
                    & 28          & \text{Aria}    & 20241223 - 17:14 & $18.6$                          & $1450$ & Park, SC       \\
    \cline{2-7}
                    & 29          & \text{Aria}    & 20241223 - 17:33 & $3.2$                           & $248$  & Park           \\
    \cline{2-7}
                    & 30          & \text{Aria}    & 20241223 - 18:47 & $17.0$                          & $1459$ & Park, SC       \\
    \toprule[0.03cm]
    \multicolumn{7}{l}{\#SC: Shopping Center.}                                   
  \end{tabular}
  \label{tab:s1_sequences}
  \vspace{-0.4cm}
\end{table}

\subsection{Topological Localization Configuration}
\begin{figure}[H]
  \centering
  \subfigure[]{
    \includegraphics[width=0.192\linewidth]{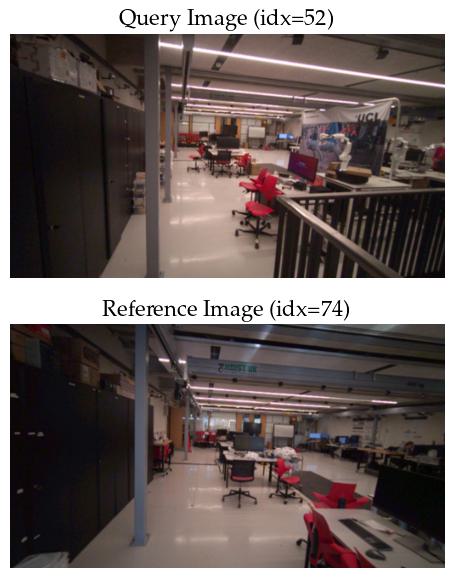}
  }
  \hspace{-0.4cm}
  \subfigure[]{
    \includegraphics[width=0.192\linewidth]{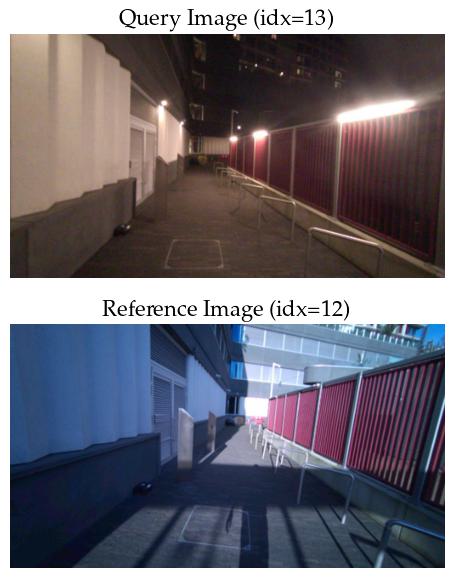}
  }
  \hspace{-0.4cm}
  \subfigure[]{
    \includegraphics[width=0.192\linewidth]{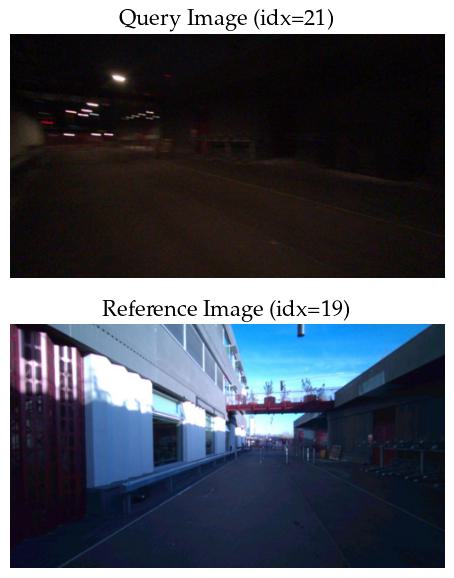}
  }
  \hspace{-0.4cm}
  \subfigure[]{
    \includegraphics[width=0.192\linewidth]{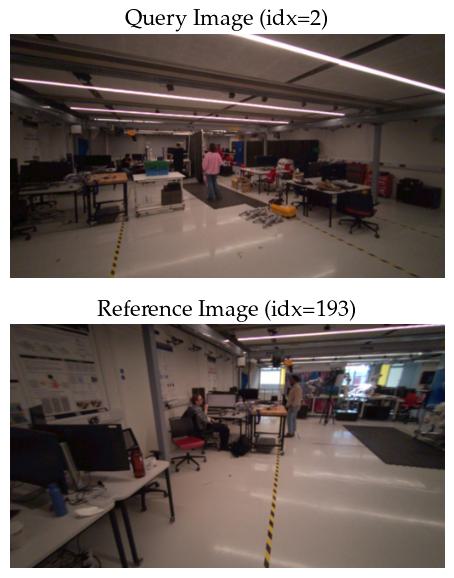}
  }
  \hspace{-0.4cm}
  \subfigure[]{
    \includegraphics[width=0.192\linewidth]{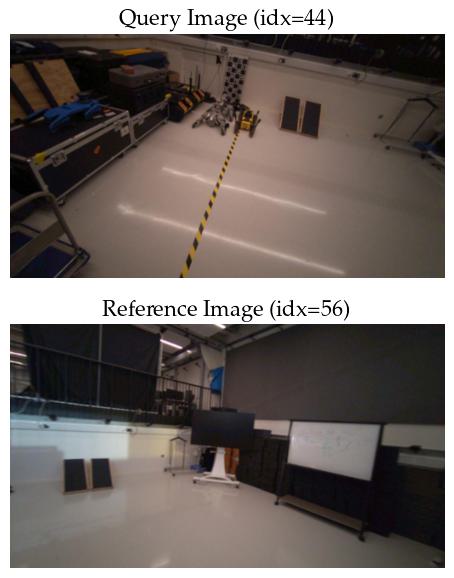}
  }
  \caption{Matched query (top) and reference (bottom) pairs for topological localization (Reference~$0$; see Fig.~\ref{fig:s1_topo_traj_ref0}). Pairs exhibit pronounced appearance changes and viewpoint variations, confirming the challenge posed by multi-session crowdsourced data.}
  \label{fig:s1_topo_matched_ref0}
\end{figure}

\begin{figure}[H]
  \centering
  \subfigure[]{
    \includegraphics[width=0.192\linewidth]{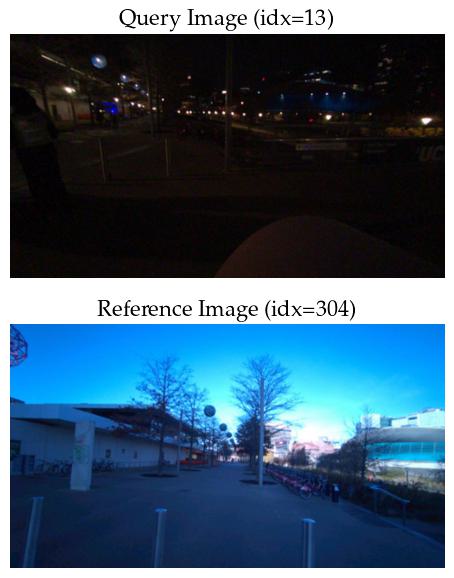}
  }
  \hspace{-0.4cm}
  \subfigure[]{
    \includegraphics[width=0.192\linewidth]{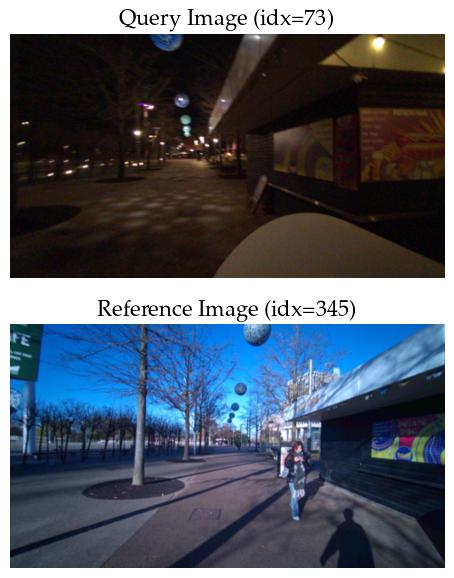}
  }
  \hspace{-0.4cm}
  \subfigure[]{
    \includegraphics[width=0.192\linewidth]{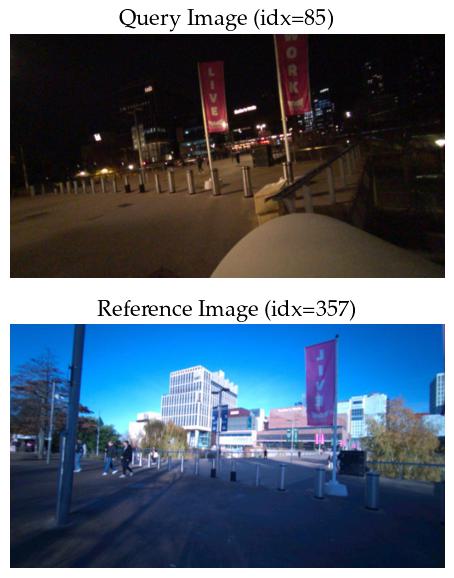}
  }
  \hspace{-0.4cm}
  \subfigure[]{
    \includegraphics[width=0.192\linewidth]{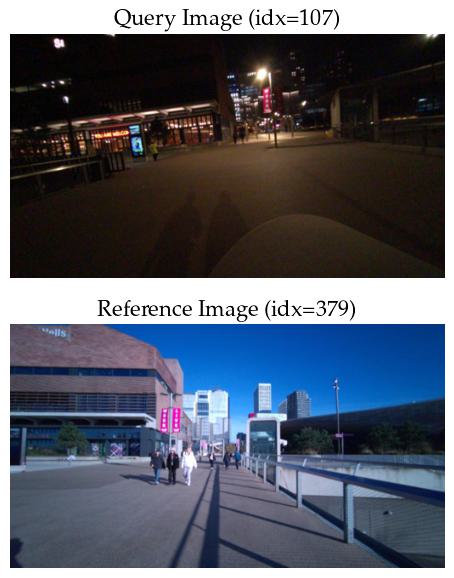}
  }
  \hspace{-0.4cm}
  \subfigure[]{
    \includegraphics[width=0.192\linewidth]{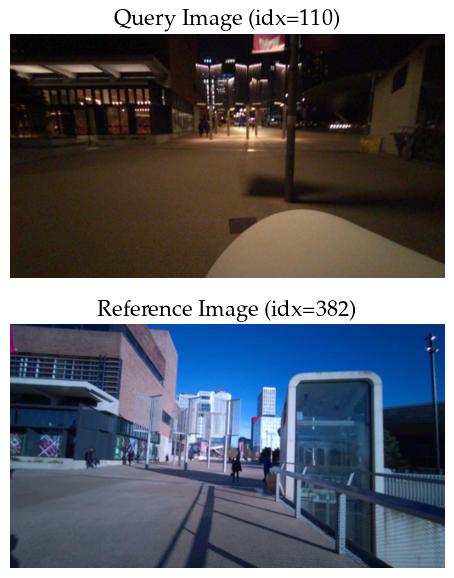}
  }
  \caption{Matched query (top) and reference (bottom) pairs for topological localization (Reference~$1$; see Fig.~\ref{fig:s1_topo_traj_ref1}). Pairs exhibit pronounced appearance changes and viewpoint variations across $4$ query trajectories collected on multiple dates.}
  \label{fig:s1_topo_matched_ref1}
\end{figure}

\begin{figure}[H]
  \centering
  \subfigure[]{
    \includegraphics[width=0.192\linewidth]{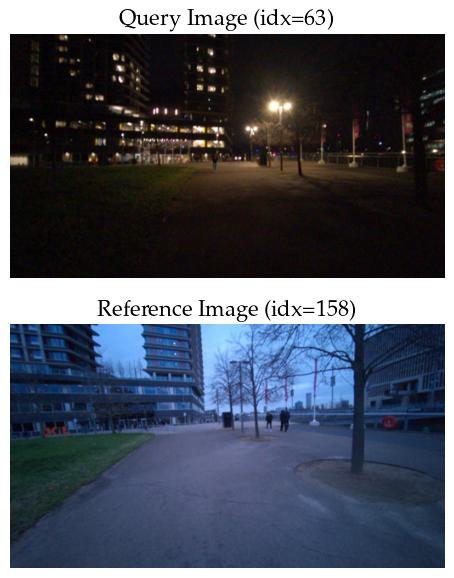}
  }
  \hspace{-0.4cm}
  \subfigure[]{
    \includegraphics[width=0.192\linewidth]{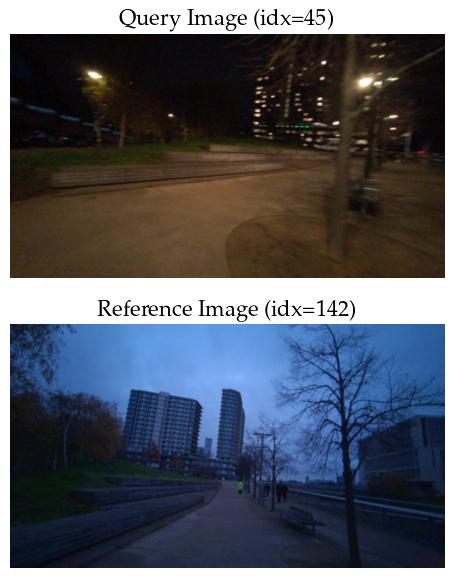}
  }
  \hspace{-0.4cm}
  \subfigure[]{
    \includegraphics[width=0.192\linewidth]{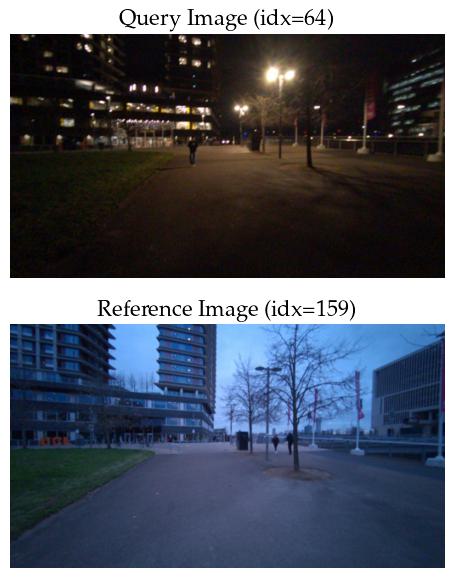}
  }
  \hspace{-0.4cm}
  \subfigure[]{
    \includegraphics[width=0.192\linewidth]{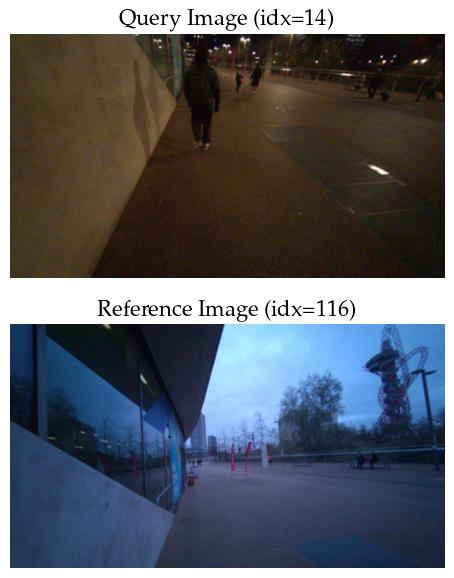}
  }
  \hspace{-0.4cm}
  \subfigure[]{
    \includegraphics[width=0.192\linewidth]{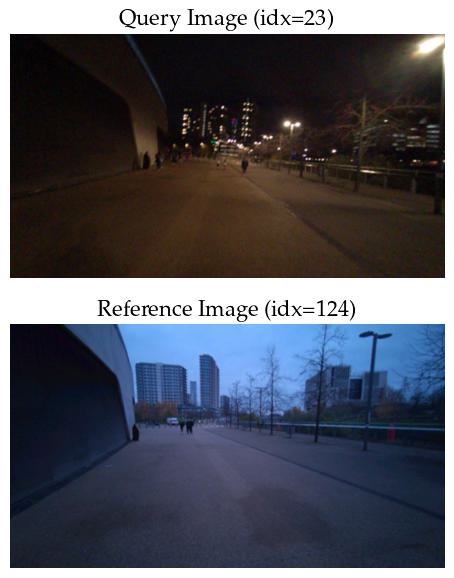}
  }
  \caption{Matched query (top) and reference (bottom) pairs for topological localization (Reference~$2$; see Fig.~\ref{fig:s1_topo_traj_ref2}). Pairs exhibit pronounced appearance changes and viewpoint variations across $5$ query trajectories collected on multiple dates.}
  \label{fig:s1_topo_matched_ref2}
\end{figure}

\begin{figure}[H]
  \centering
  \subfigure[]{
    \includegraphics[width=0.192\linewidth]{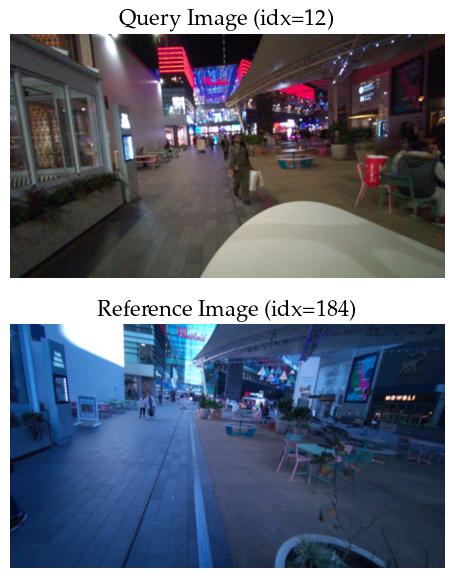}
  }
  \hspace{-0.4cm}
  \subfigure[]{
    \includegraphics[width=0.192\linewidth]{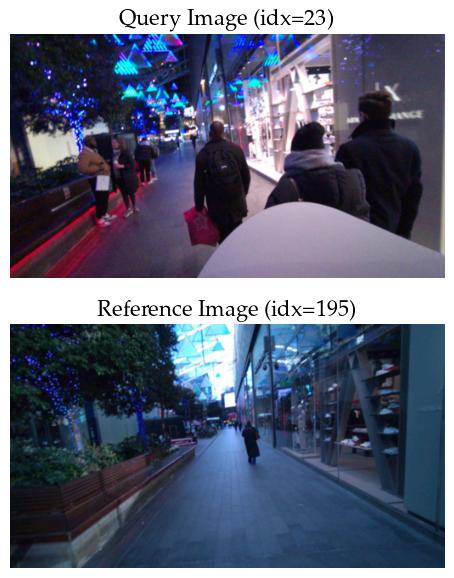}
  }
  \hspace{-0.4cm}
  \subfigure[]{
    \includegraphics[width=0.192\linewidth]{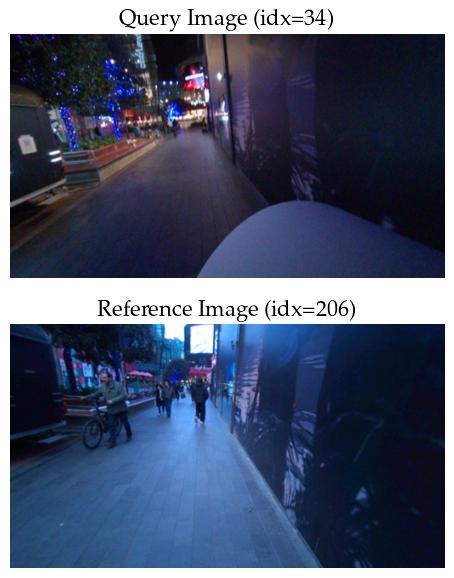}
  }
  \hspace{-0.4cm}
  \subfigure[]{
    \includegraphics[width=0.192\linewidth]{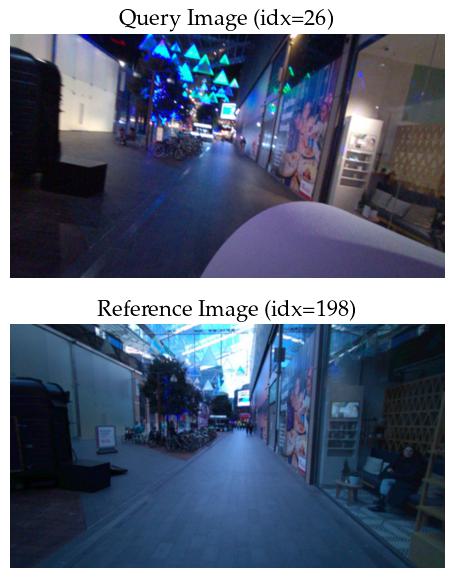}
  }
  \hspace{-0.4cm}
  \subfigure[]{
    \includegraphics[width=0.192\linewidth]{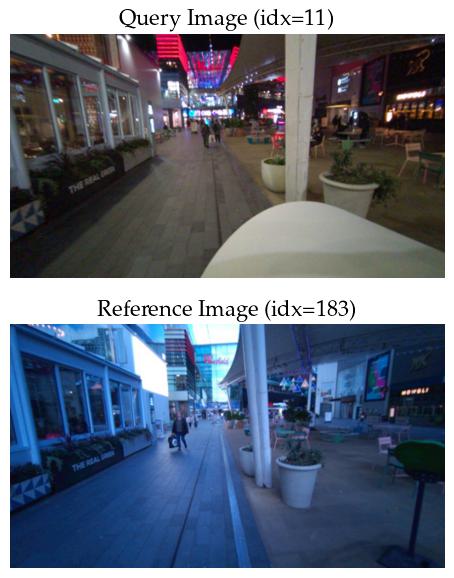}
  }
  \caption{Matched query (top) and reference (bottom) pairs for topological localization (Reference~$3$; see Fig.~\ref{fig:s1_topo_traj_ref3}). Pairs exhibit pronounced appearance changes and viewpoint variations across $3$ query trajectories collected on multiple dates.}
  \label{fig:s1_topo_matched_ref3}
\end{figure}

\begin{figure}[H]
  \centering
  \subfigure[2024/09/04]{
    \includegraphics[width=0.196\linewidth]{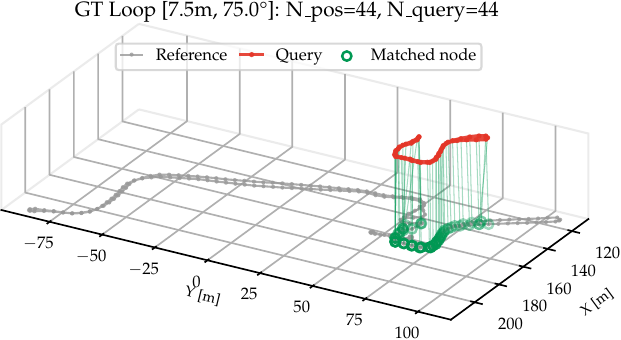}
  }
  \hspace{-0.45cm}
  \subfigure[2024/11/27]{
    \includegraphics[width=0.196\linewidth]{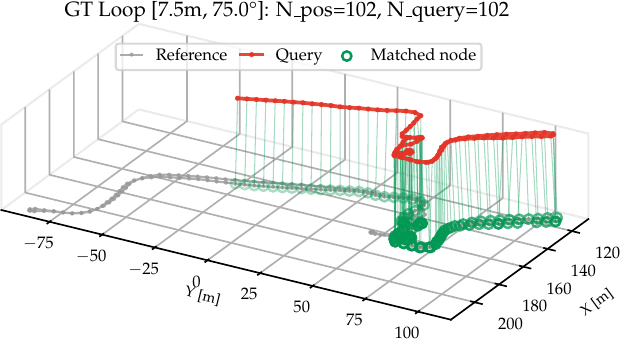}
  }
  \hspace{-0.45cm}
  \subfigure[2024/11/27]{
    \includegraphics[width=0.196\linewidth]{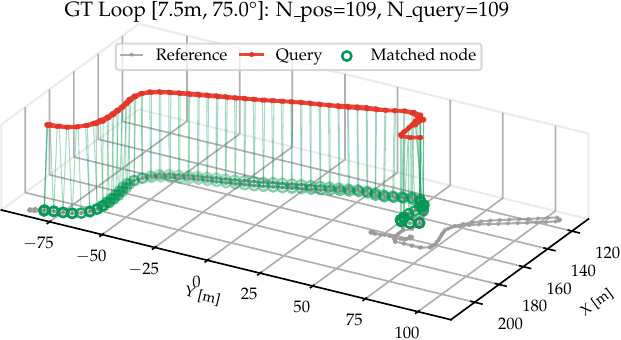}
  }
  \hspace{-0.45cm}
  \subfigure[2024/12/02]{
    \includegraphics[width=0.196\linewidth]{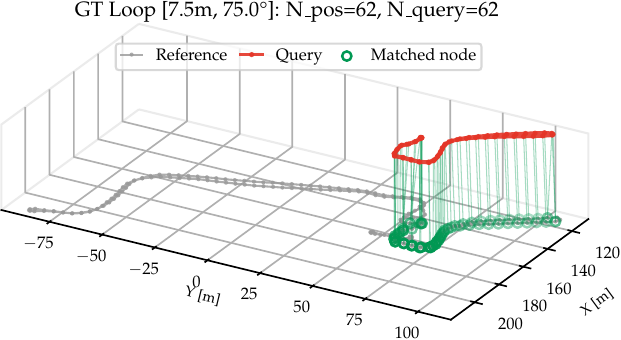}
  }
  \hspace{-0.45cm}
  \subfigure[2024/12/02]{
    \includegraphics[width=0.196\linewidth]{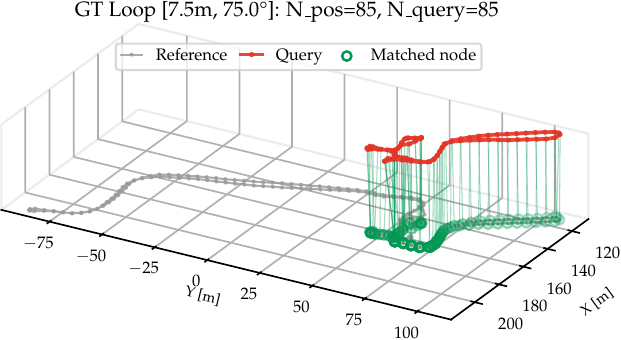}
  }
  \hspace{-0.45cm}
  \subfigure[2024/12/02]{
    \includegraphics[width=0.196\linewidth]{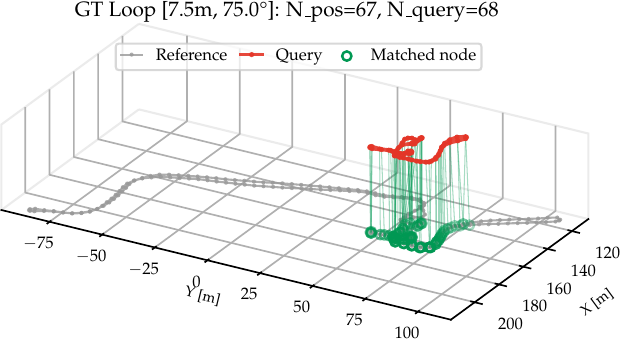}
  }
  \hspace{-0.45cm}
  \subfigure[2024/12/04]{
    \includegraphics[width=0.196\linewidth]{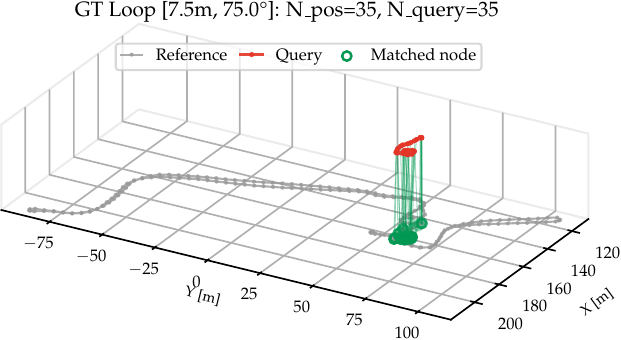}
  }
  \hspace{-0.45cm}
  \subfigure[2024/12/04]{
    \includegraphics[width=0.196\linewidth]{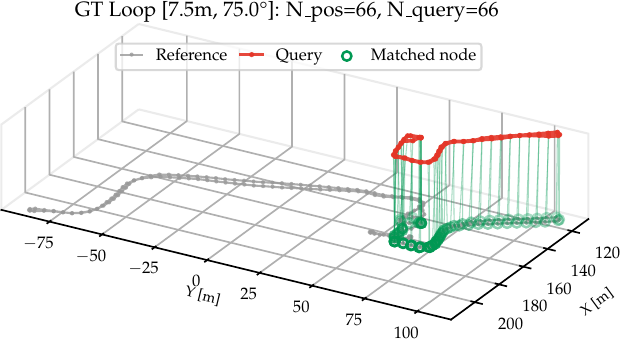}
  }
  \hspace{-0.45cm}
  \subfigure[2024/12/04]{
    \includegraphics[width=0.196\linewidth]{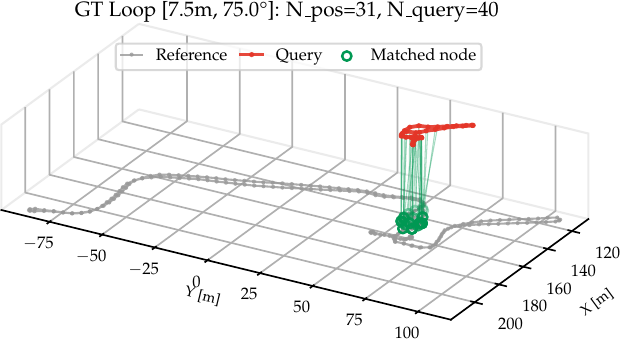}
  }
  \hspace{-0.45cm}
  \subfigure[2024/12/04]{
    \includegraphics[width=0.196\linewidth]{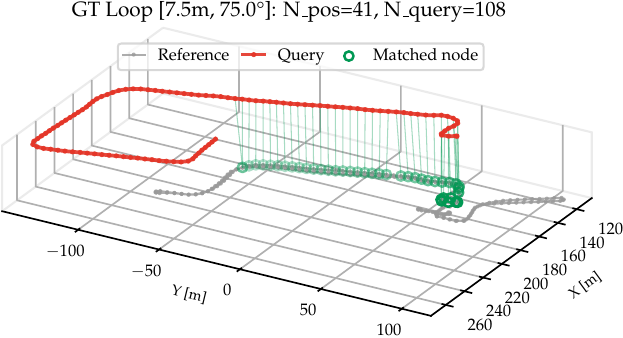}
  }
  \hspace{-0.45cm}
  \subfigure[2024/12/04]{
    \includegraphics[width=0.196\linewidth]{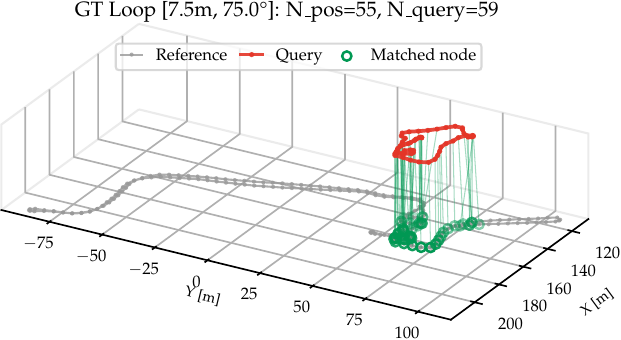}
  }
  \hspace{-0.45cm}
  \subfigure[2024/12/05]{
    \includegraphics[width=0.196\linewidth]{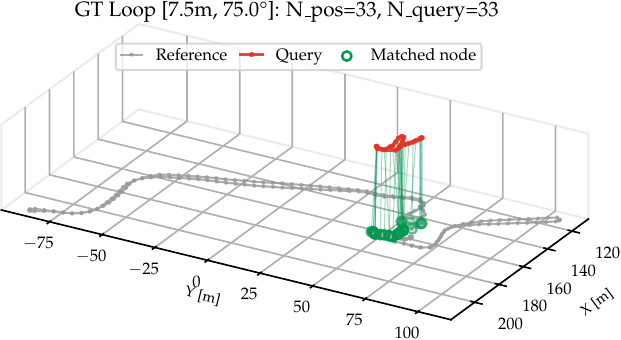}
  }
  \hspace{-0.45cm}
  \subfigure[2024/12/05]{
    \includegraphics[width=0.196\linewidth]{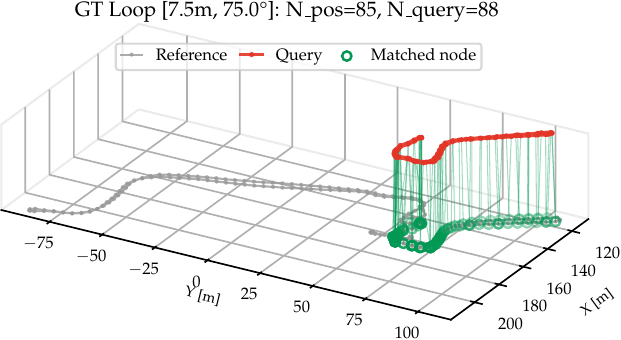}
  }
  \hspace{-0.45cm}
  \subfigure[2024/12/23]{
    \includegraphics[width=0.196\linewidth]{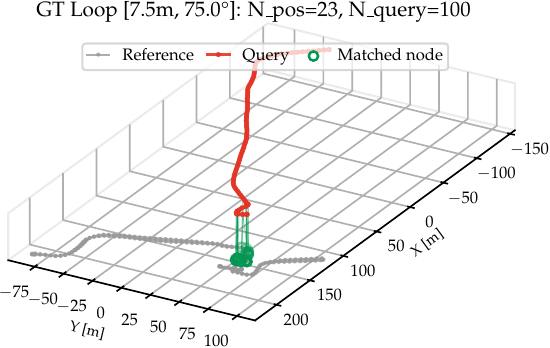}
  }
  \caption{Reference-query trajectory pairs for topological localization (Reference~$0$, collected 2024/11/28). Valid pairs lie within $[7.5\text{m},\,75^{\circ}]$; co-located node pairs are shown as edge connections. The $14$ query trajectories span four months and exhibit partial, reversed, and interleaved overlap patterns, illustrating the diversity of crowdsourced traversals over the same region.}
  \label{fig:s1_topo_traj_ref0}
\end{figure}

\begin{figure}[H]
  \centering
  \subfigure[2024/12/04]{
    \includegraphics[width=0.196\linewidth]{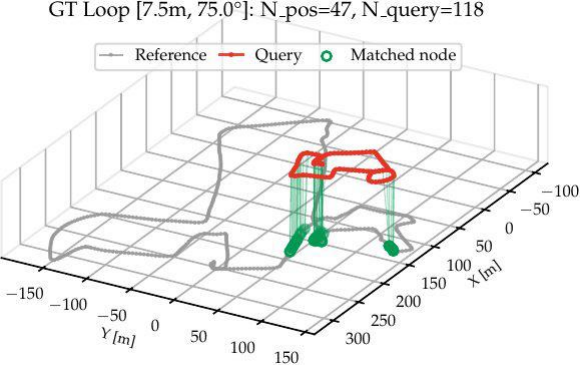}
  }
  \hspace{-0.45cm}
  \subfigure[2024/12/04]{
    \includegraphics[width=0.196\linewidth]{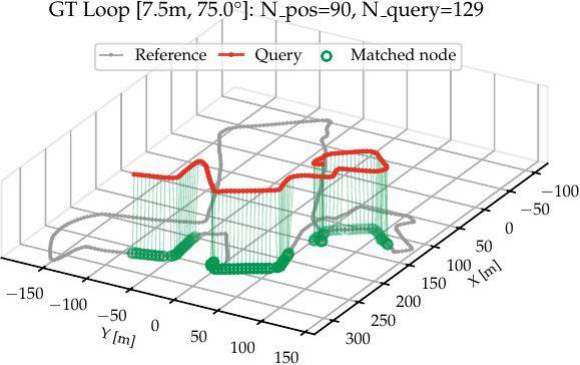}
  }
  \hspace{-0.45cm}
  \subfigure[2024/12/23]{
    \includegraphics[width=0.196\linewidth]{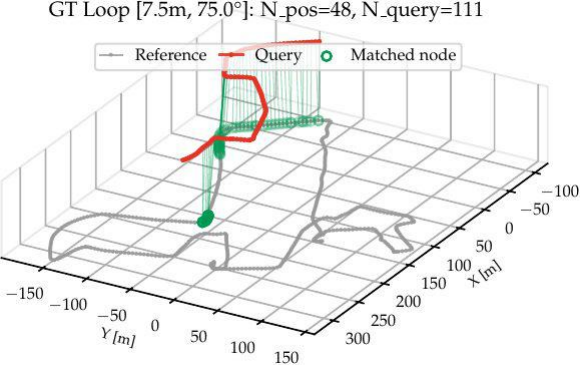}
  }
  \hspace{-0.45cm}
  \subfigure[2024/12/23]{
    \includegraphics[width=0.196\linewidth]{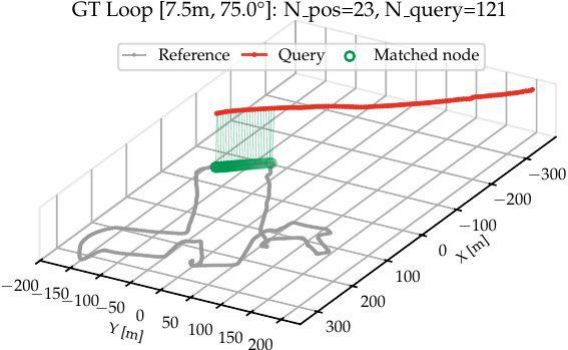}
  }
  \caption{Reference-query trajectory pairs for topological localization (Reference~$1$, collected 2024/11/29). Valid pairs lie within $[7.5\text{m},\,75^{\circ}]$; co-located node pairs are shown as edge connections. The $4$ query trajectories illustrate diverse overlap patterns and appearance changes across dates.}
  \label{fig:s1_topo_traj_ref1}
\end{figure}

\begin{figure}[H]
  \centering
  \subfigure[2024/12/04]{
    \includegraphics[width=0.196\linewidth]{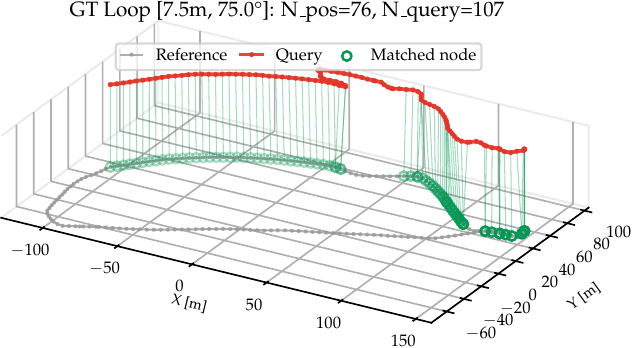}
  }
  \hspace{-0.45cm}
  \subfigure[2024/12/04]{
    \includegraphics[width=0.196\linewidth]{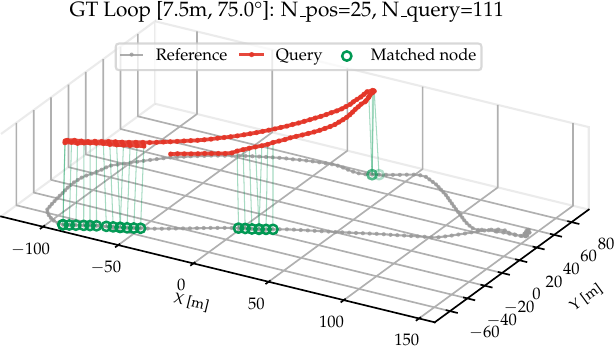}
  }
  \hspace{-0.45cm}
  \subfigure[2024/12/04]{
    \includegraphics[width=0.196\linewidth]{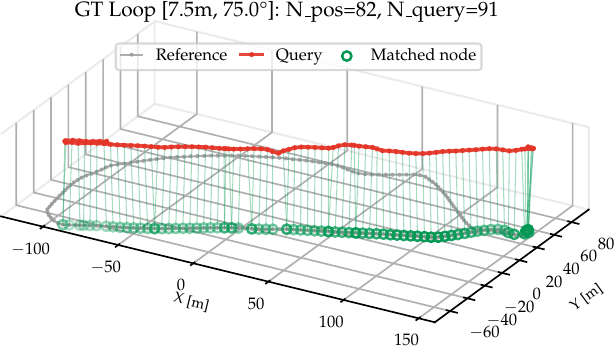}
  }
  \hspace{-0.45cm}
  \subfigure[2024/12/23]{
    \includegraphics[width=0.196\linewidth]{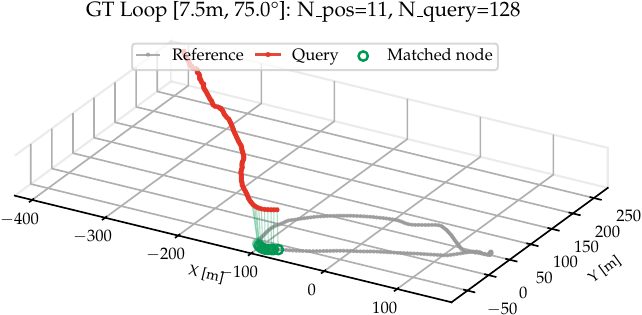}
  }
  \hspace{-0.45cm}
  \subfigure[2024/12/23]{
    \includegraphics[width=0.196\linewidth]{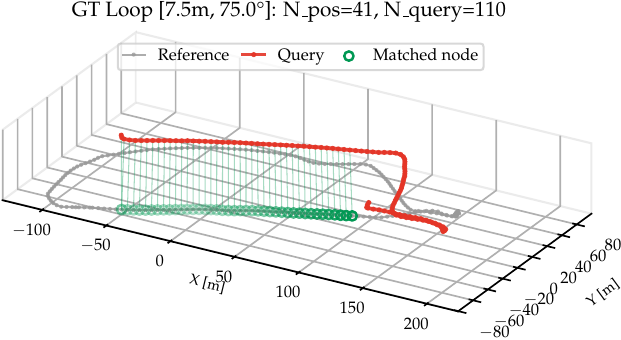}
  }
  \caption{Reference-query trajectory pairs for topological localization (Reference~$2$, collected 2024/12/04). Valid pairs lie within $[7.5\text{m},\,75^{\circ}]$; co-located node pairs are shown as edge connections. The $5$ query trajectories demonstrate branching and partial-overlap patterns characteristic of crowdsourced multi-session data.}
  \label{fig:s1_topo_traj_ref2}
\end{figure}

\begin{figure}[H]
  \centering
  \subfigure[2024/12/21]{
    \includegraphics[width=0.196\linewidth]{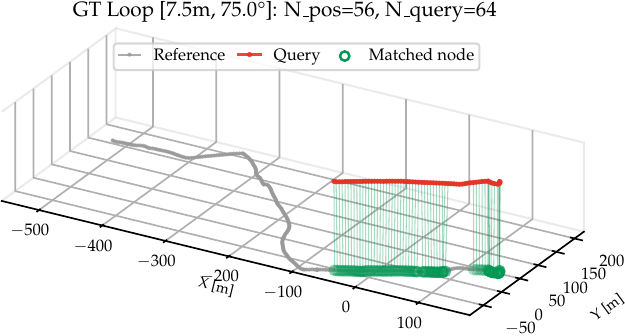}
  }
  \hspace{-0.45cm}
  \subfigure[2024/12/23]{
    \includegraphics[width=0.196\linewidth]{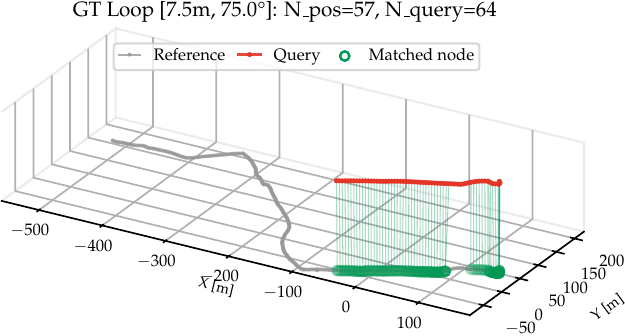}
  }
  \hspace{-0.45cm}
  \subfigure[2024/12/23]{
    \includegraphics[width=0.196\linewidth]{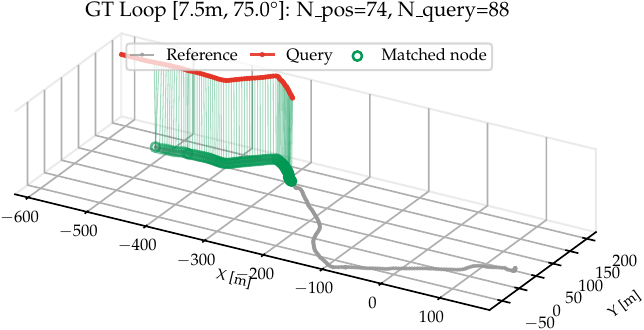}
  }
  \caption{Reference-query trajectory pairs for topological localization (Reference~$3$, collected 2024/12/22). Valid pairs lie within $[7.5\text{m},\,75^{\circ}]$; co-located node pairs are shown as edge connections. The $3$ query trajectories exhibit interleaved traversal patterns over two days.}
  \label{fig:s1_topo_traj_ref3}
\end{figure}

From the R$2$ dataset, we manually curated $4$ reference sequences.
For each reference, corresponding query sequences with significant spatial and visual overlap were identified, yielding $26$ query sequences in total.
Fig.~\ref{fig:s1_topo_matched_ref0}-\ref{fig:s1_topo_matched_ref3} present examples of matched reference-query image pairs (References~$0$-$3$), which exhibit pronounced appearance changes and viewpoint variations; co-located pairs satisfy a proximity threshold of $[7.5\text{m},\,75^{\circ}]$.
Fig.~\ref{fig:s1_topo_traj_ref0}-\ref{fig:s1_topo_traj_ref3} display the GT trajectories of these reference-query pairs, with co-located node pairs connected by edges.
A defining challenge of crowdsourced mapping is that query sequences overlap the reference along highly irregular patterns: partial coverage, reversed traversals, branching detours, or interleaved paths, rather than a single contiguous overlap.
This structural diversity violates the core assumption of SeqSLAM, which requires a monotonically aligned subsequence between query and reference.
Our dynamic-programming (DP)-based matching, by contrast, accommodates arbitrary overlap patterns through a flexible path search on the appearance-distance matrix.

% Matched images for topological localization - reference ID 0

% Matched images for topological localization - reference ID 1

% Matched images for topological localization - reference ID 2

% Matched images for topological localization - reference ID 3

% Reference ID 0 - 14 query trajectories

% Reference ID 1 - 4 query trajectories

% Reference ID 2 - 5 query trajectories

% Reference ID 3 - 3 query trajectories

\begin{figure}[H]
  \centering
  \subfigure[Reference~$0$ (2024/11/28) vs. query (2024/09/04)]{
    \label{fig:s1_topo_dmatrix_ref0}
    \includegraphics[width=0.99\linewidth]{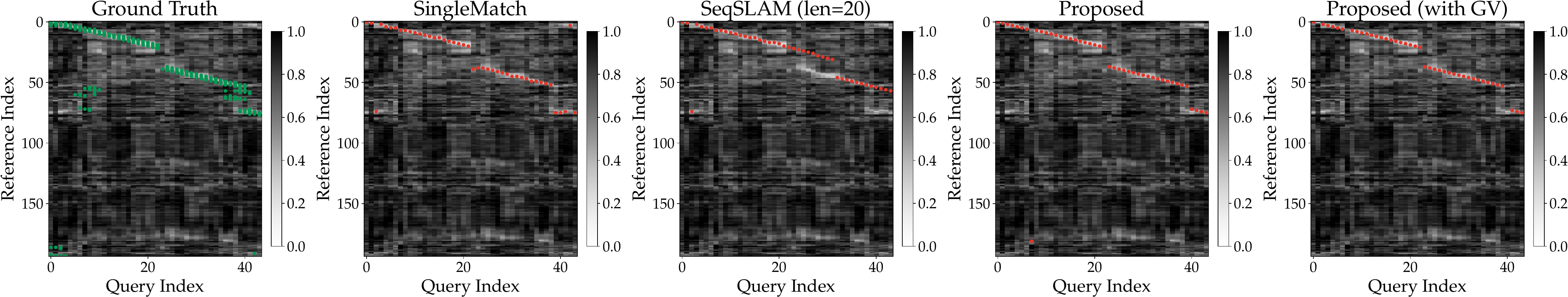}
  }
  \\
  \subfigure[Reference~$1$ (2024/11/29) vs. query (2024/12/04)]{
    \label{fig:s1_topo_dmatrix_ref1}
    \includegraphics[width=0.99\linewidth]{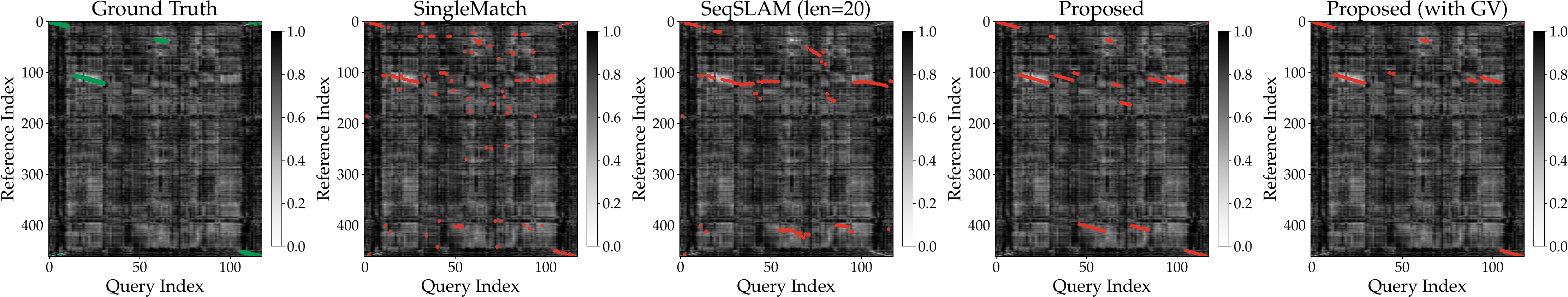}
  }
  \\
  \subfigure[Reference~$2$ (2024/12/04) vs. query (2024/12/04)]{
    \label{fig:s1_topo_dmatrix_ref2}
    \includegraphics[width=0.99\linewidth]{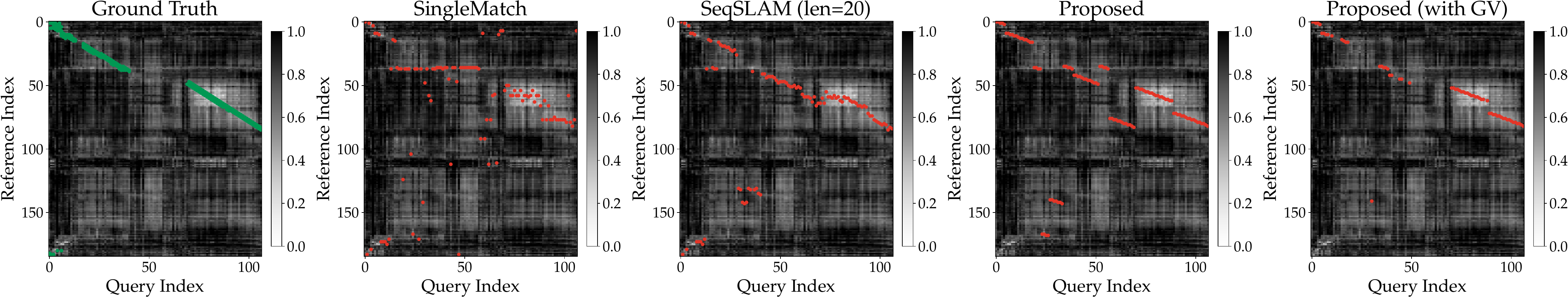}
  }
  \\
  \subfigure[Reference~$3$ (2024/12/22) vs. query (2024/12/23)]{
    \label{fig:s1_topo_dmatrix_ref3}
    \includegraphics[width=0.99\linewidth]{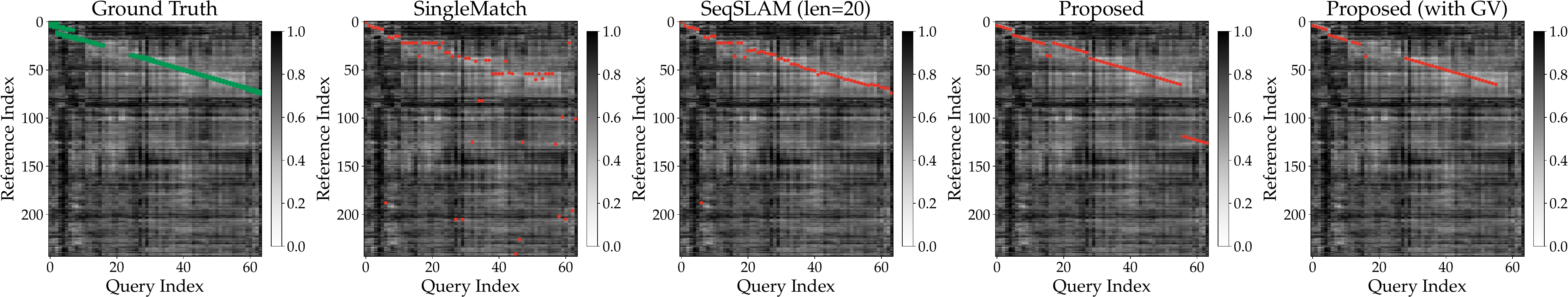}
  }
  \caption{Sequence-matching difference matrices for topological localization, one representative query per reference sequence (collection dates in subcaptions). Each row of panels shows the appearance-distance matrix between reference (rows) and query (columns) frames, overlaid with matched paths from: ground truth, single-frame matching, SeqSLAM, the proposed dynamic-programming (DP)-based matching, and DP after geometric verification (GV) (rightmost column); darker cells indicate larger distances. The DP-based matching recovers valid pairs under irregular overlap patterns, while GV prunes the residual false-positive correspondences retained by the other methods.}
  \label{fig:s1_topo_dmatrix}
\end{figure}

Fig.~\ref{fig:s1_topo_dmatrix} visualizes the appearance-distance matrices for a representative query of each reference sequence, contrasting matched paths recovered by the GT, single-frame matching, SeqSLAM with sequence length $20$, DP-based matching, and DP after GV (rightmost column).
Matrices are computed from CosPlace descriptors (ResNet-$18$ backbone, $256$-D).
DP-based matching yields more consistent detections along the sequence and remains flexible under irregular overlap, as seen for Reference~$0$ in Fig.~\ref{fig:s1_topo_dmatrix_ref0}.
All sequence-matching methods tend to retain an excess of candidate pairs, accumulating false positives that require post-hoc filtering; GV effectively prunes these spurious correspondences, retaining only geometrically consistent pairs.

% Sequence-matching difference matrices for topological localization

\subsection{Metric Localization Protocol}
\begin{figure}[H]
    \centering
    \subfigure[s00002]{
        \includegraphics[height=0.385\linewidth]{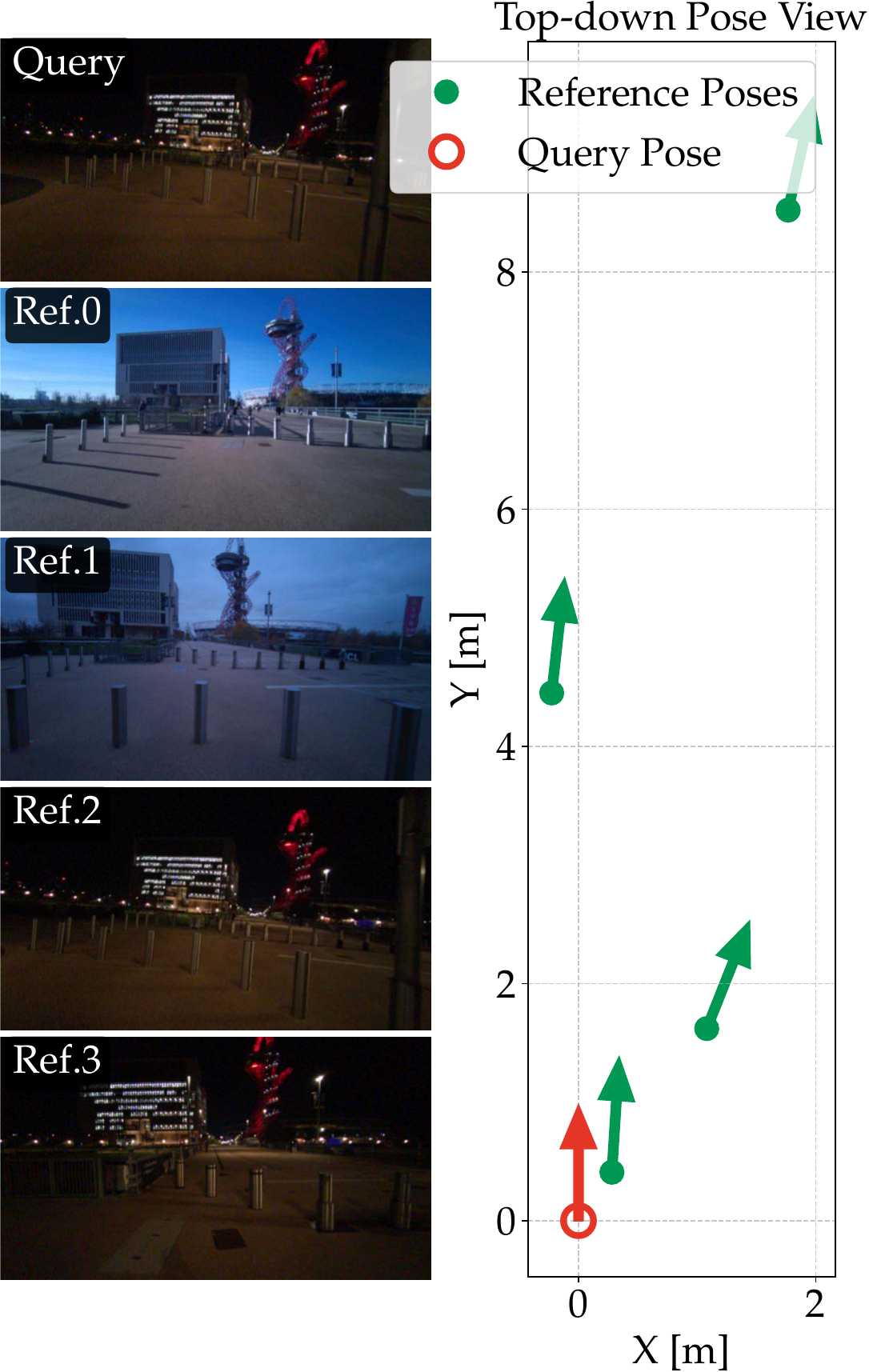}
        \label{fig:s1_metric_ex0}
    }
    \hspace{0.2cm}
    \subfigure[s00038]{
        \includegraphics[height=0.385\linewidth]{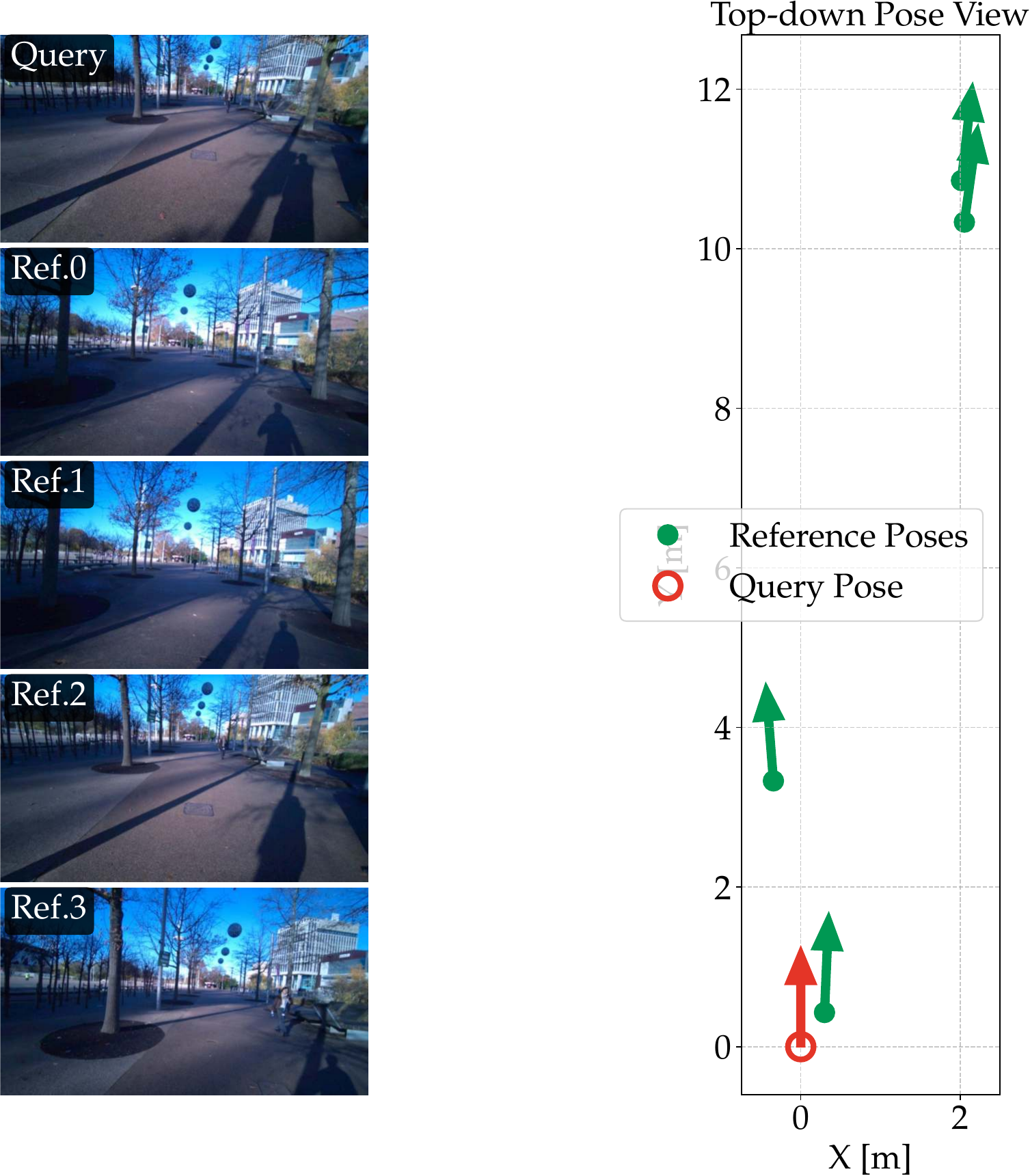}
        \label{fig:s1_metric_ex1}
    }
    \hspace{0.2cm}
    \subfigure[s00053]{
        \includegraphics[height=0.385\linewidth]{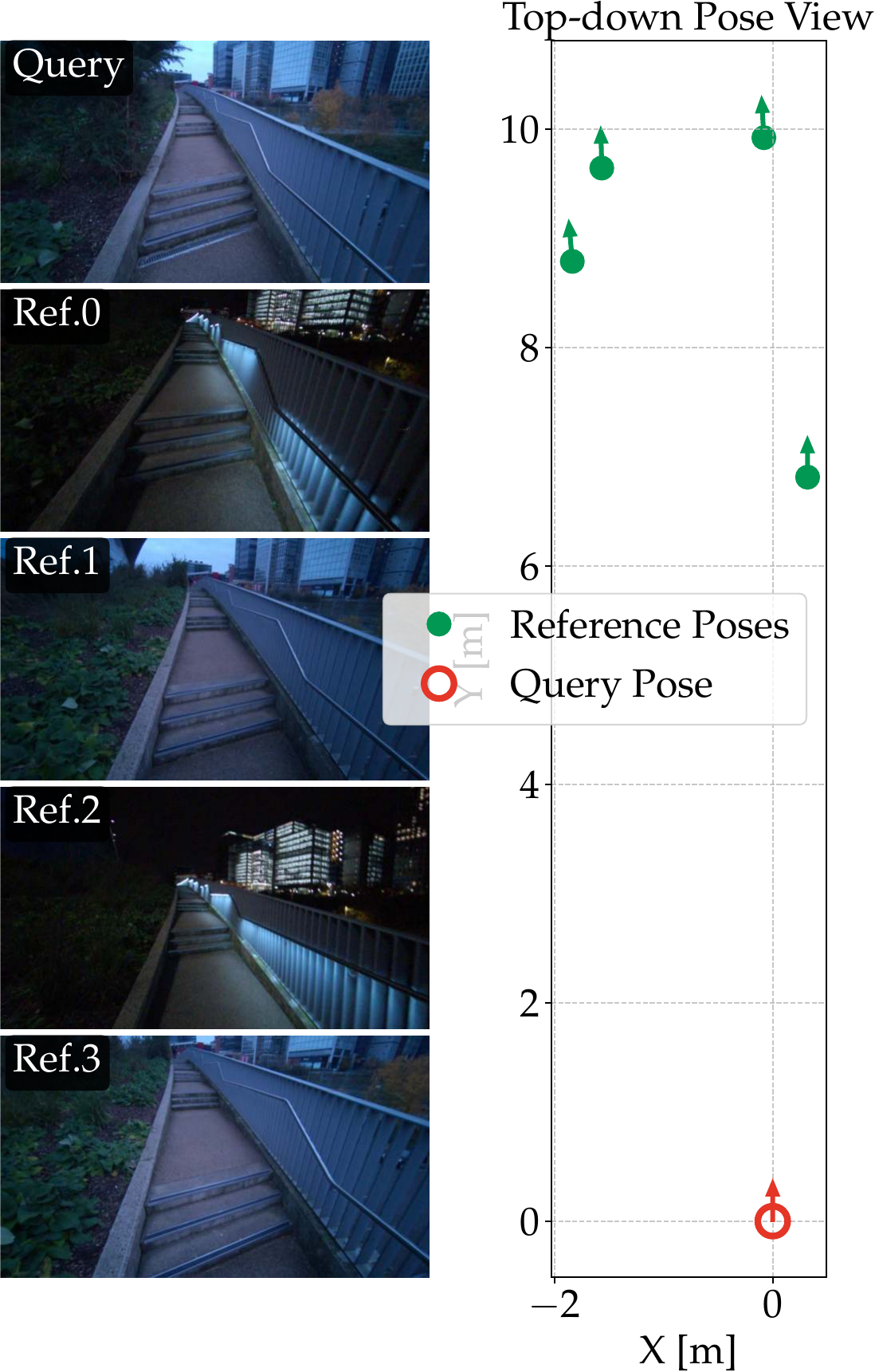}
        \label{fig:s1_metric_ex2}
    }
    \\[4pt]
    \subfigure[s00055]{
        \includegraphics[height=0.41\linewidth]{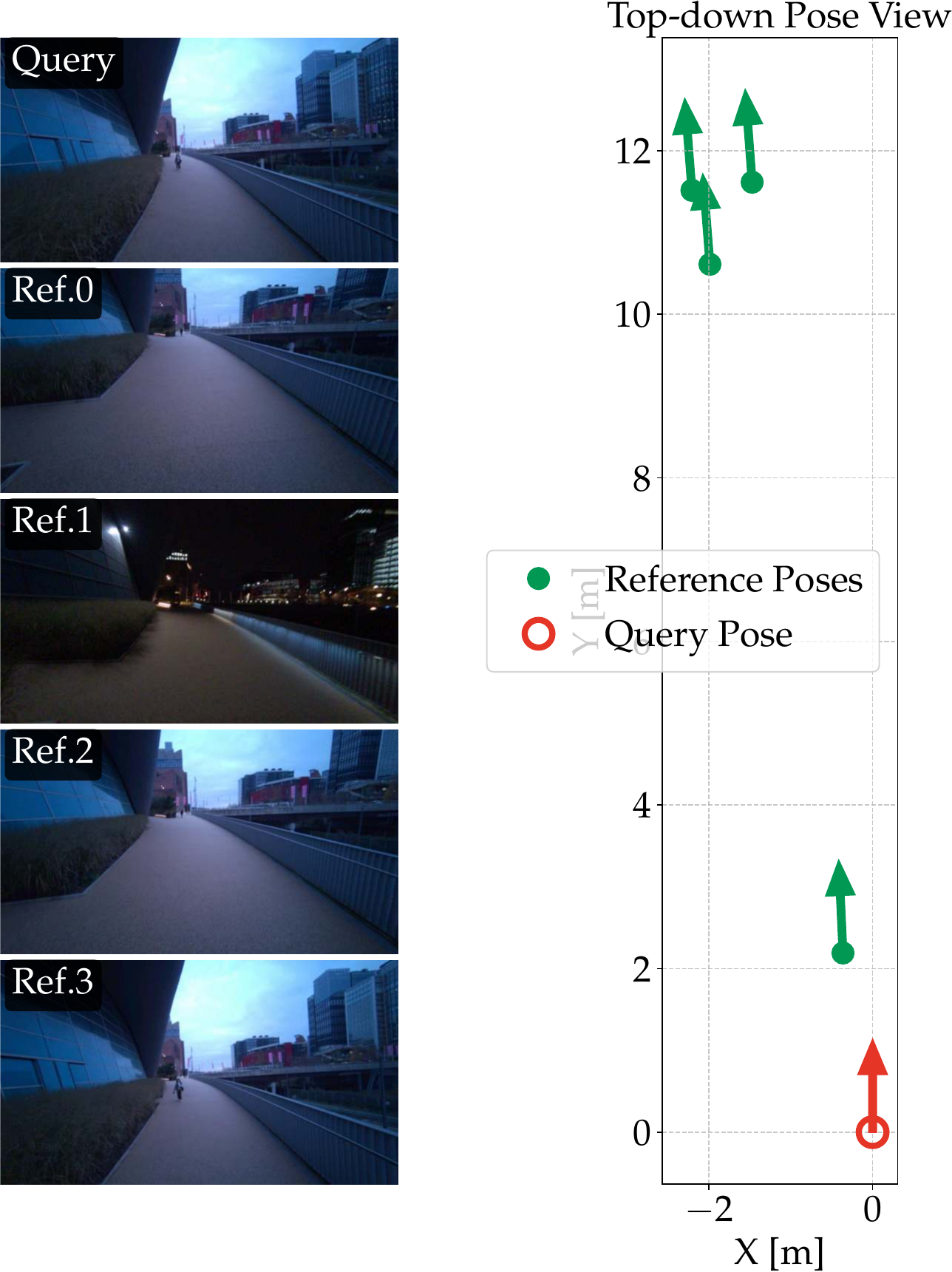}
        \label{fig:s1_metric_ex3}
    }
    \hspace{0.2cm}
    \subfigure[s00074]{
        \includegraphics[height=0.41\linewidth]{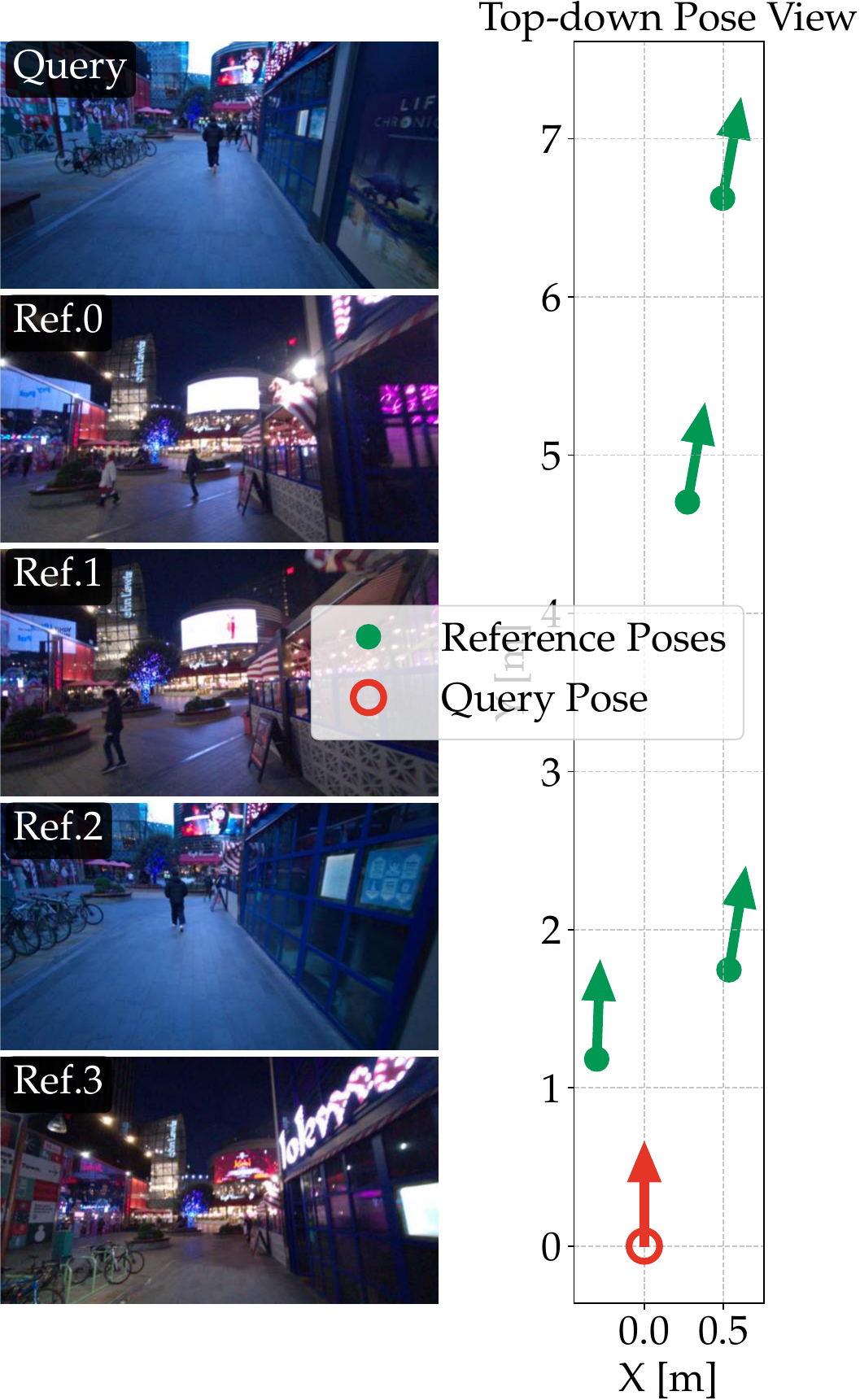}
        \label{fig:s1_metric_ex4}
    }
    \hspace{0.2cm}
    \subfigure[s00075]{
        \includegraphics[height=0.41\linewidth]{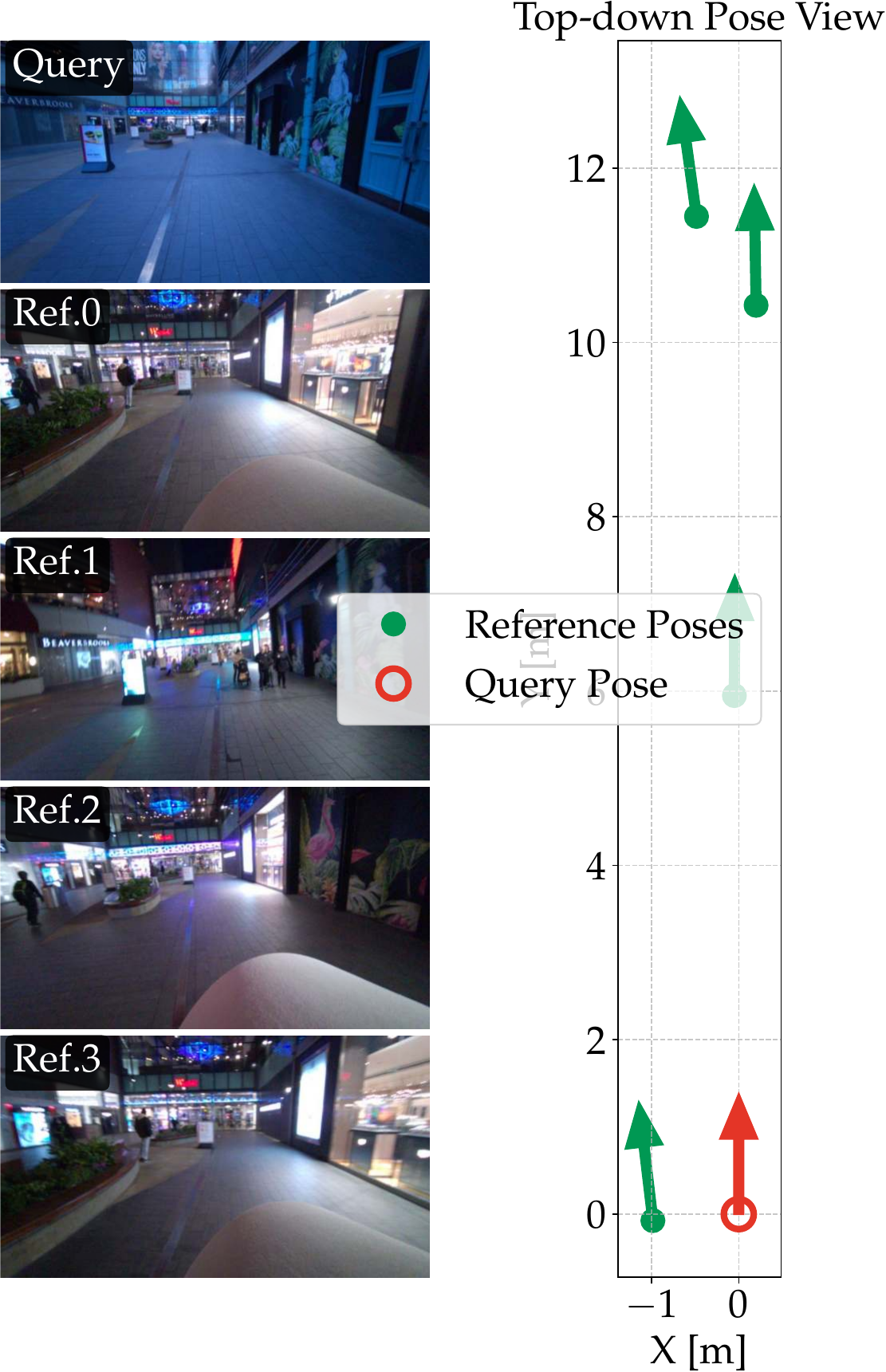}
        \label{fig:s1_metric_ex5}
    }
    \caption{Metric localization examples randomly sampled from the self-collected dataset (six scenes, query IDs in subcaptions). For each example, the left column shows the query image alongside its retrieved reference images (Ref.~$0$-$3$); the right panel gives a top-down pose view comparing the ground truth (GT) reference poses (green) with the estimated query pose (red). The examples highlight three key challenges: large spatial baselines, significant viewpoint variations, and cross-session appearance changes.}
    \label{fig:s1_metric_examples}
\end{figure}

For metric localization evaluation, we randomly sampled $57$ query images per scene from R$2$.
Reference images were retrieved by spatial proximity (within $5$m) and verified visual overlap, quantified by the number of matched keypoints.
Query images exhibit diverse viewpoints and illumination conditions.
Fig.~\ref{fig:s1_metric_examples} visualizes $5$ randomly sampled examples: each panel shows the query image alongside its retrieved reference images and a top-down pose view comparing the GT reference poses against the estimated query pose, illustrating key challenges: large spatial baselines, viewpoint variations, and appearance changes.

\subsection{Extended Map Merging Visualization}
\begin{figure}[H]
    \centering
    \subfigure[R$0$-InOrder]{\includegraphics[width=0.325\linewidth]{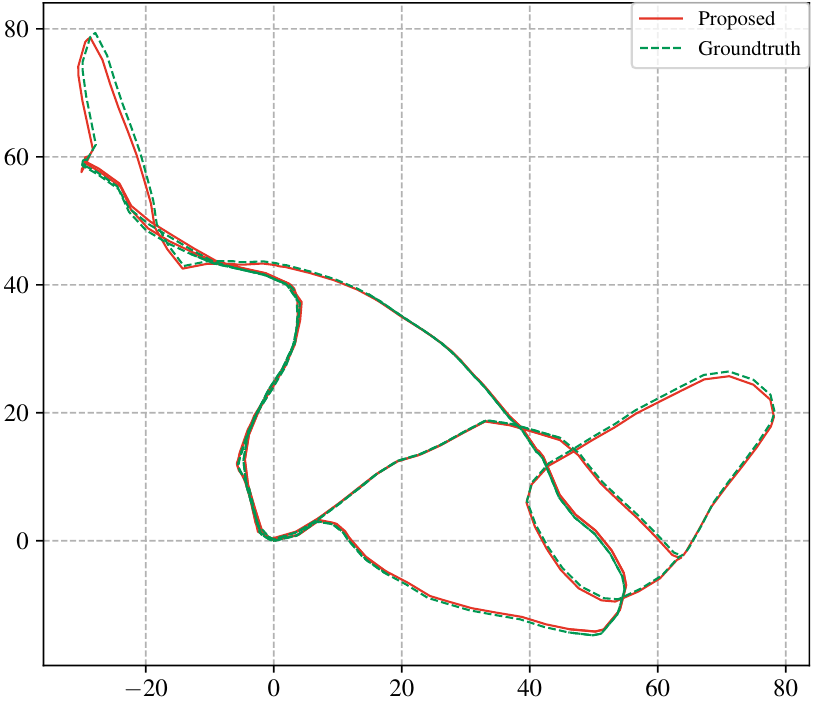}}
    \hspace{-0.2cm}
    \subfigure[R$0$-$0$]{\includegraphics[width=0.325\linewidth]{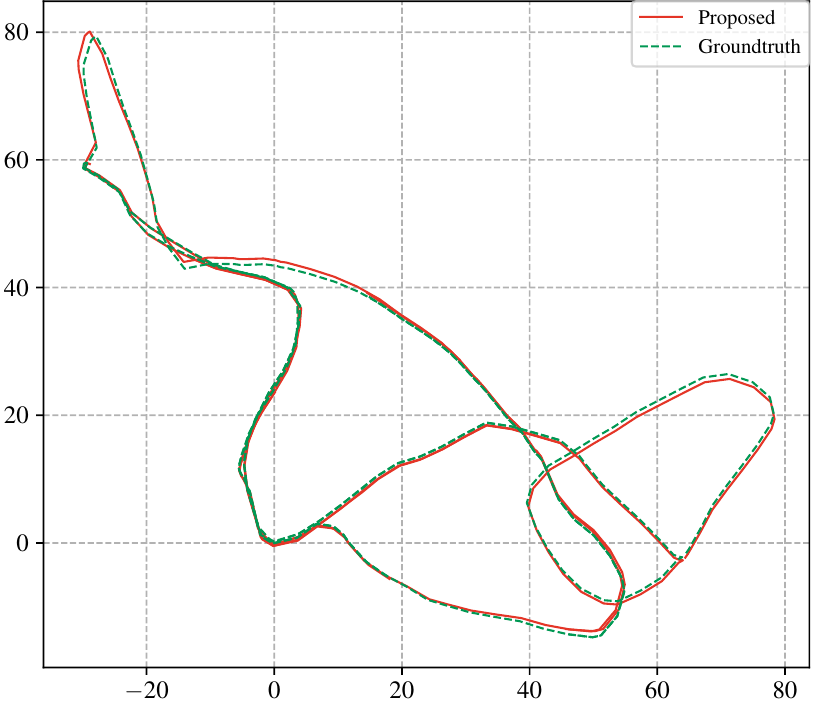}}
    \hspace{-0.2cm}
    \subfigure[R$0$-$1$]{\includegraphics[width=0.325\linewidth]{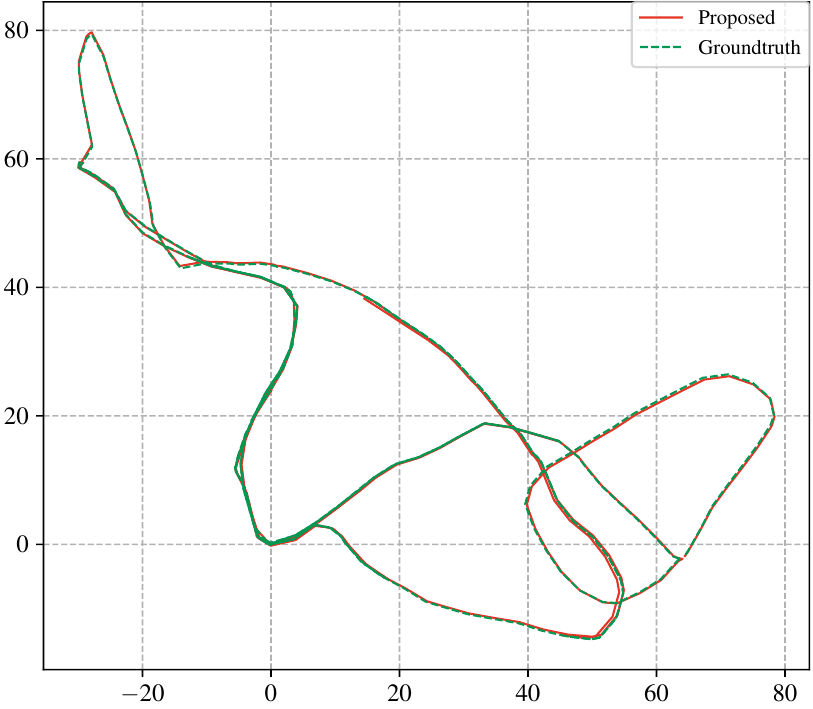}}
    \caption{Trajectory alignment results for the map merging pipeline on R$0$ under shuffled segment ordering. Estimated trajectories (colored) are compared against ground truth (gray). InOrder shows the in-order baseline; R$0$-$k$ denotes the $k$-th shuffled permutation.}
    \label{fig:s1_mapmerge_r0}
\end{figure}

\begin{figure}[H]
    \centering
    \subfigure[R$1$-InOrder]{\includegraphics[width=0.33\linewidth]{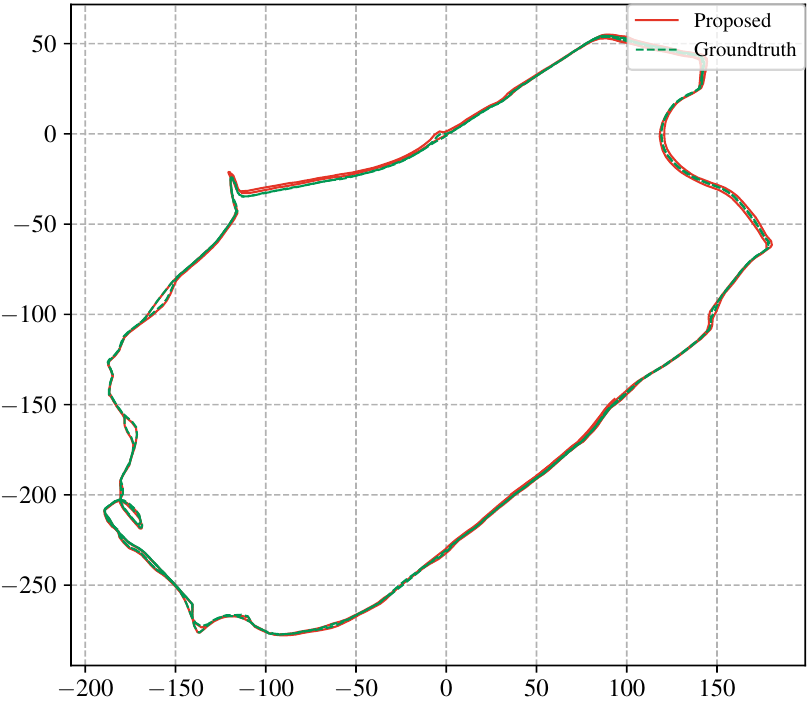}}
    \hspace{-0.2cm}
    \subfigure[R$1$-$0$]{\includegraphics[width=0.33\linewidth]{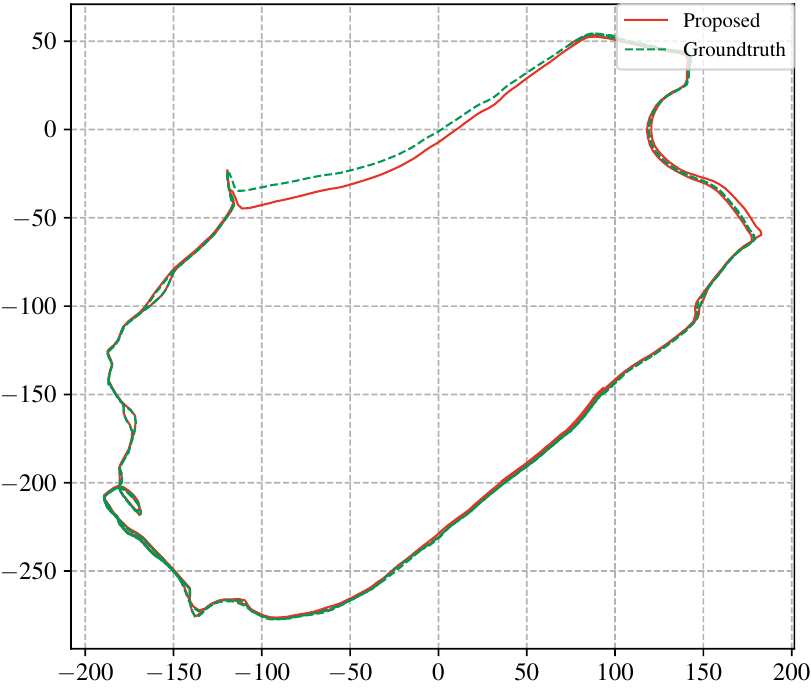}}
    \hspace{-0.2cm}
    \subfigure[R$1$-$1$]{\includegraphics[width=0.33\linewidth]{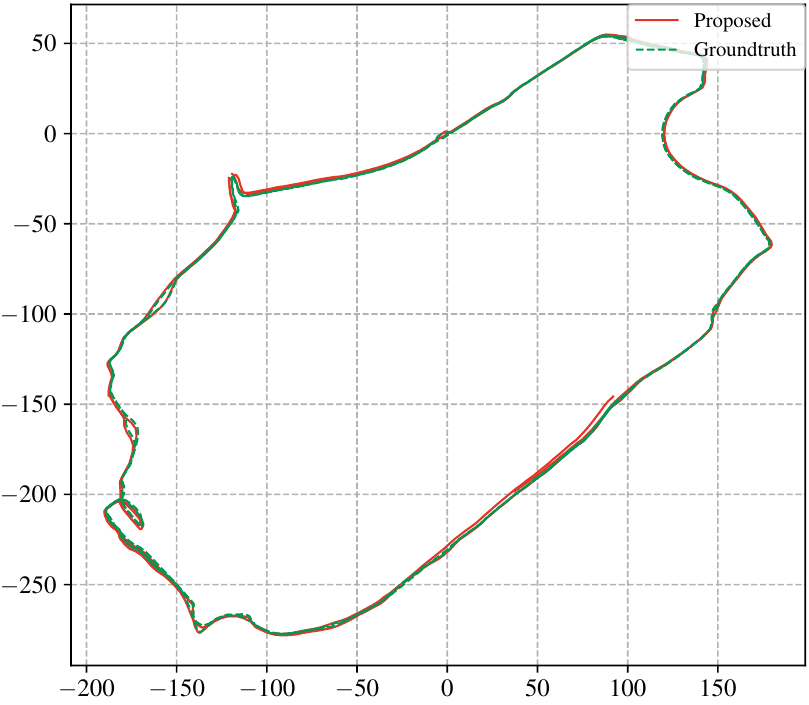}}
    \caption{Trajectory alignment results for the map merging pipeline on R$1$ under shuffled segment ordering. Estimated trajectories (colored) are compared against ground truth (gray).}
    \label{fig:s1_mapmerge_r1}
\end{figure}

\begin{figure}[H]
    \centering
    \subfigure[R$2$-InOrder]{\includegraphics[width=0.19\linewidth]{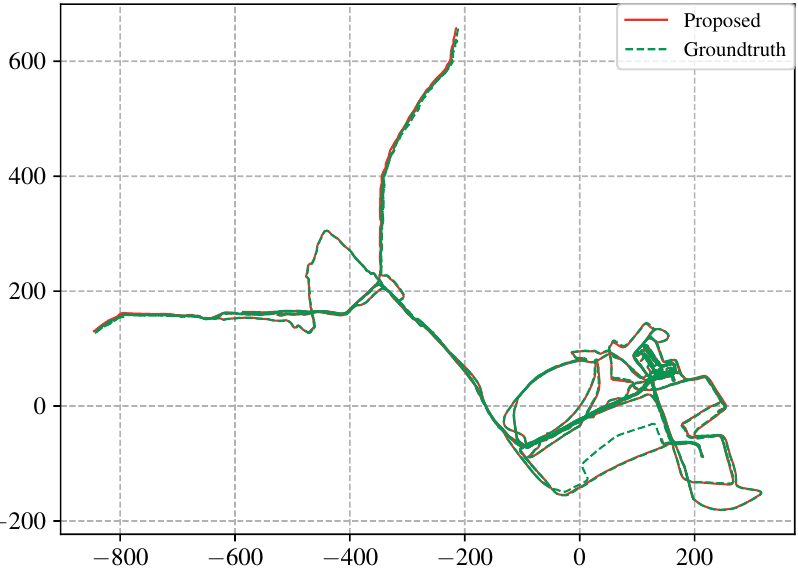}}
    \subfigure[R$2$-$0$]{\includegraphics[width=0.19\linewidth]{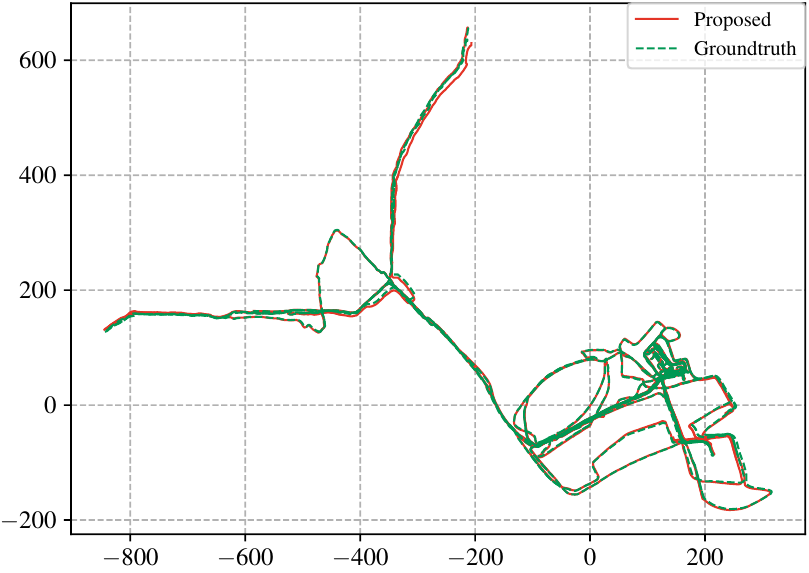}}
    \subfigure[R$2$-$1$]{\includegraphics[width=0.19\linewidth]{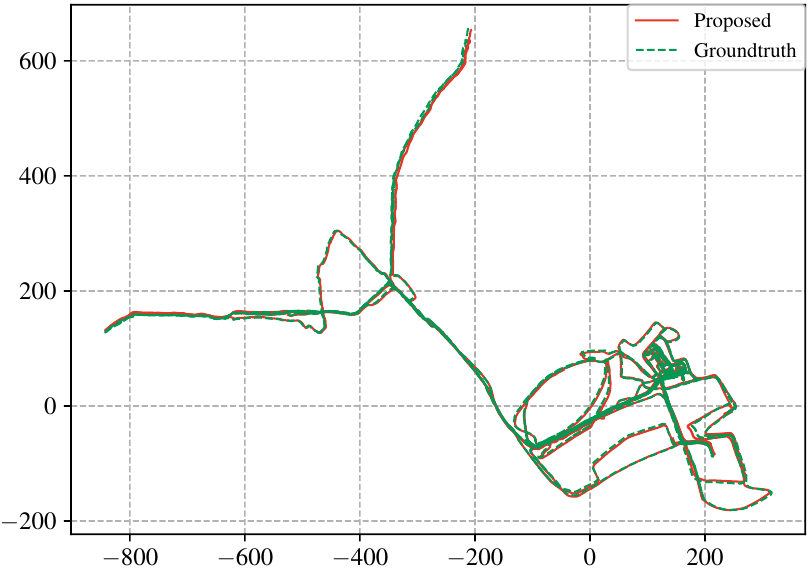}}
    \subfigure[R$2$-$2$]{\includegraphics[width=0.19\linewidth]{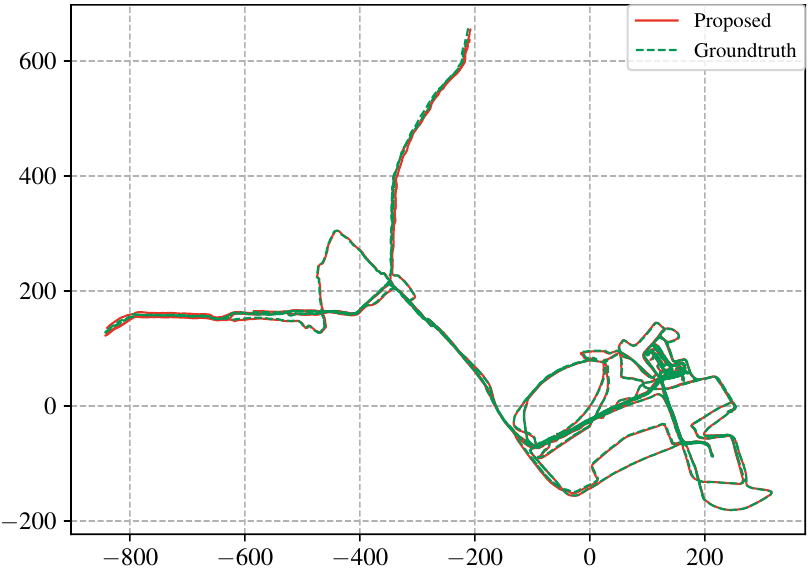}}
    \subfigure[R$2$-$3$]{\includegraphics[width=0.19\linewidth]{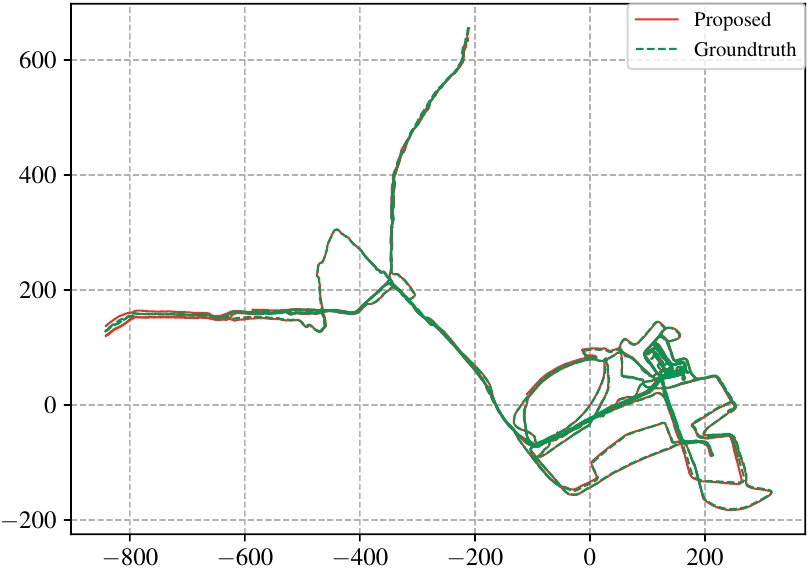}}\\
    \subfigure[R$2$-$4$]{\includegraphics[width=0.19\linewidth]{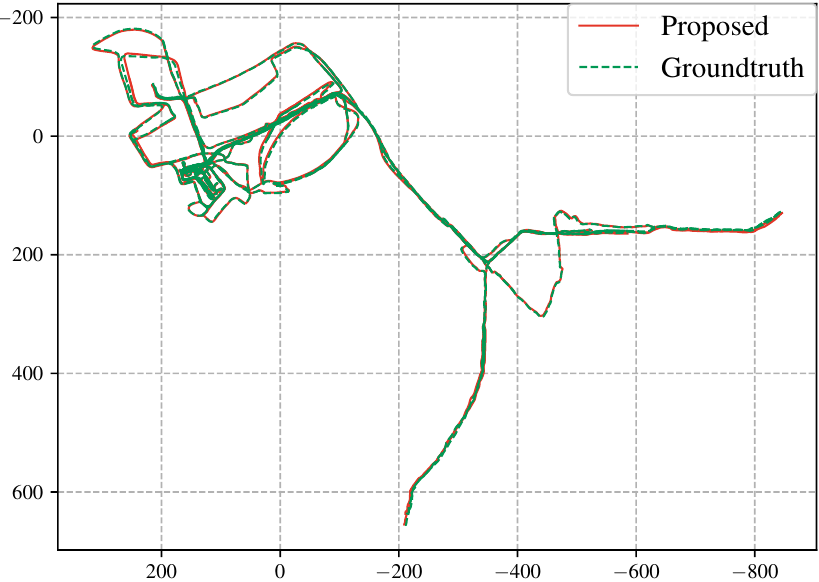}}
    \subfigure[R$2$-$5$]{\includegraphics[width=0.19\linewidth]{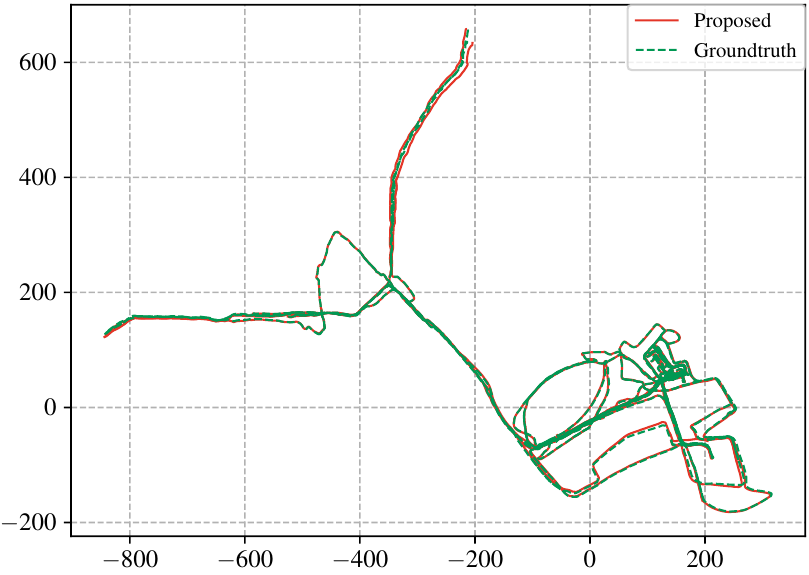}}
    \subfigure[R$2$-$6$]{\includegraphics[width=0.19\linewidth]{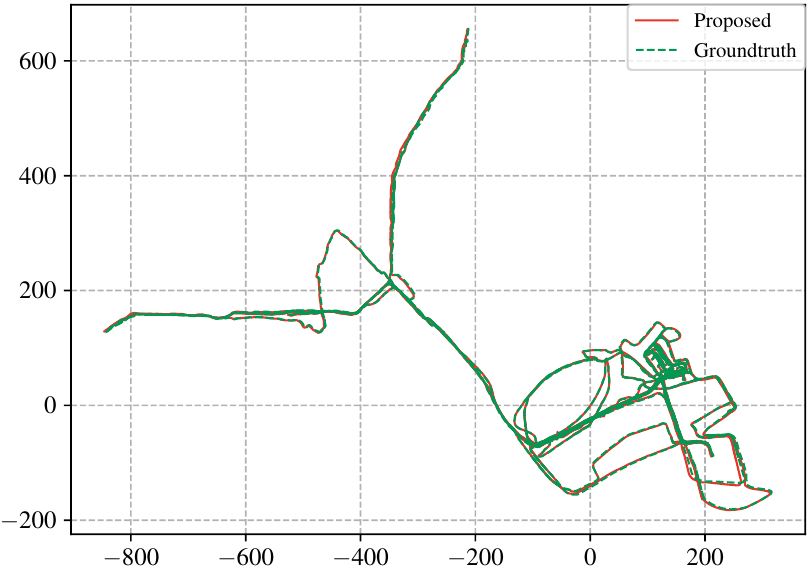}}
    \subfigure[R$2$-$7$]{\includegraphics[width=0.19\linewidth]{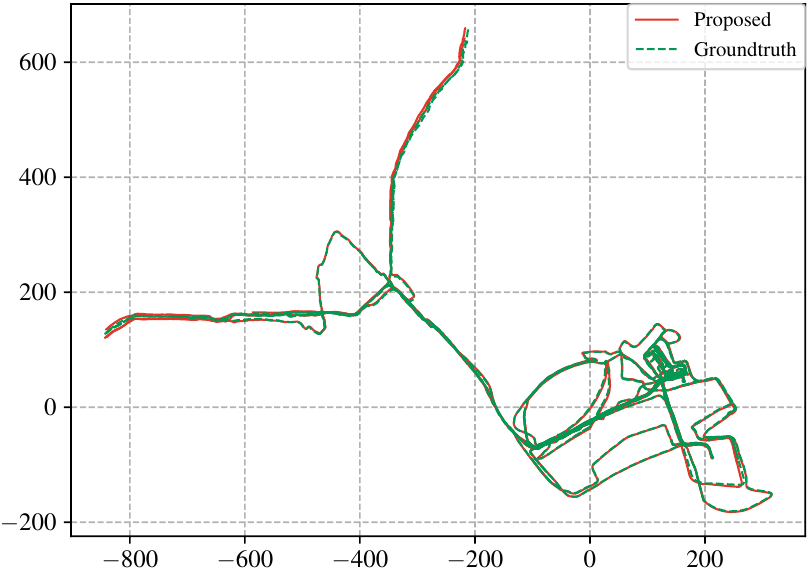}}
    \subfigure[R$2$-$8$]{\includegraphics[width=0.19\linewidth]{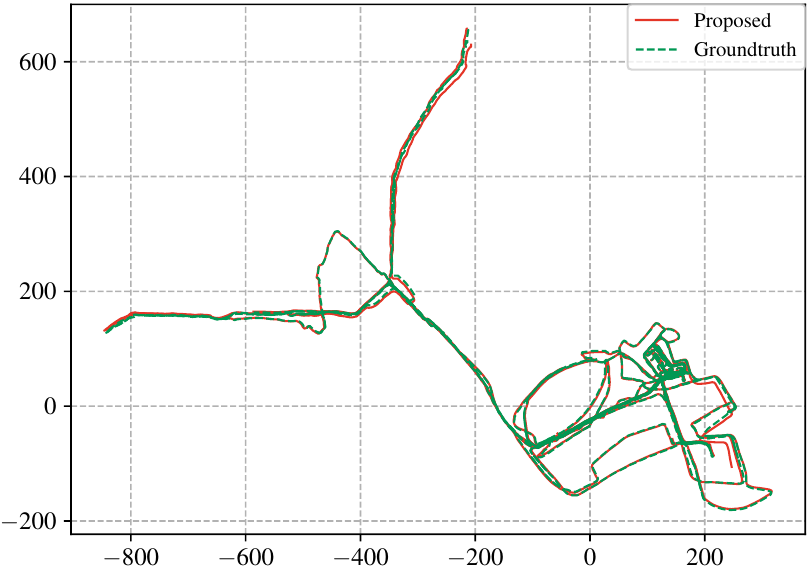}}
    \caption{Trajectory alignment results for the map merging pipeline on R$2$ under shuffled segment ordering ($9$ permutations plus in-order baseline). Estimated trajectories (colored) are compared against ground truth (gray), demonstrating globally consistent alignment across the large mixed building/park/shopping-center environment.}
    \label{fig:s1_mapmerge_r2}
\end{figure}

For the map merging experiments, we used all sequences from R$0$-R$2$ (summarized in Tab.~\ref{tab:s1_sequences}).
Each sequence was partitioned into segments of at most $300$m, with keyframes selected via a distance-based threshold\footnote{Translation and rotation thresholds: $3.9$m and $60^{\circ}$, respectively.}, yielding $68$ distinct segments that simulate short-term data capture from multiple distributed users.
Camera intrinsics, extrinsics, timestamps, local VIO poses, and GT poses are released alongside the dataset.
Fig.~\ref{fig:s1_mapmerge_r0}-\ref{fig:s1_mapmerge_r2} present the alignment results of the map merging pipeline, comparing estimated against GT trajectories for R$0$-R$2$ under shuffled execution orders.

\newpage
\section{$360$Loc Dataset: Cross-Device Adaptation}\label{sec:s2_dataset_360loc}

The $360$Loc dataset~\cite{huang2024360loc} provides raw equirectangular imagery from pedestrian-mounted omnidirectional cameras, as described in the experimental setup of the main paper (Sec.~\ref{sec:exp_datasets}).
This section details the cross-device adaptation procedure used for our map merging evaluation.

\subsection{Cross-Device Map Merging Configuration}
\begin{figure}[H]
    \centering
    % Atrium scene
    \subfigure[Atrium - Raw]{
        \includegraphics[width=0.176\linewidth]{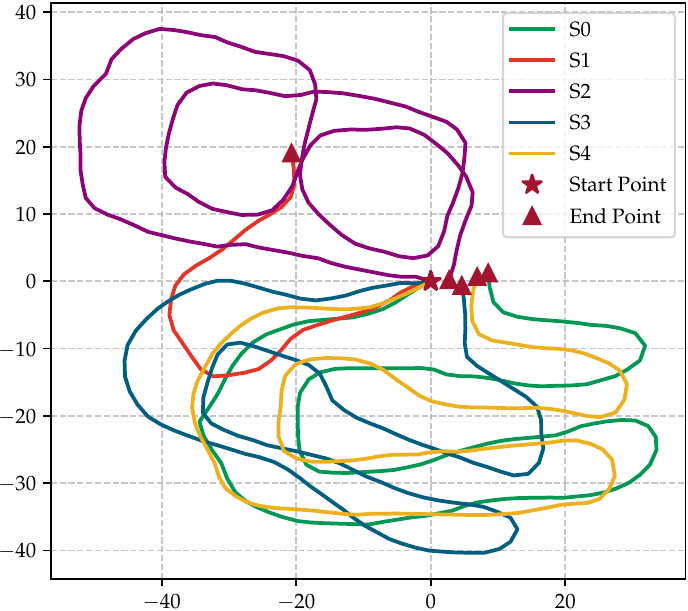}
        \label{fig:s2_mapmerge_atrium_raw}
    }
    \hspace{-0.15cm}
    \subfigure[Atrium - Merged]{
        \includegraphics[width=0.250\linewidth]{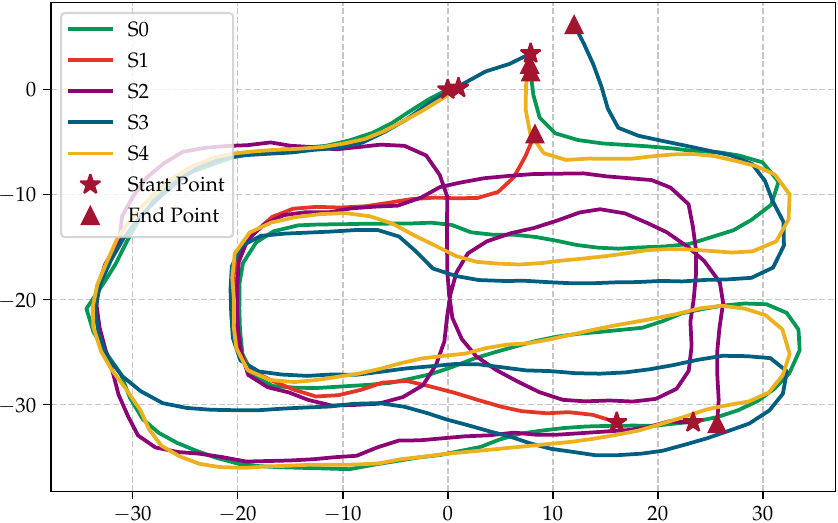}
        \label{fig:s2_mapmerge_atrium_merged}
    }
    % Concourse scene
    \subfigure[Concourse - Raw]{
        \includegraphics[width=0.175\linewidth]{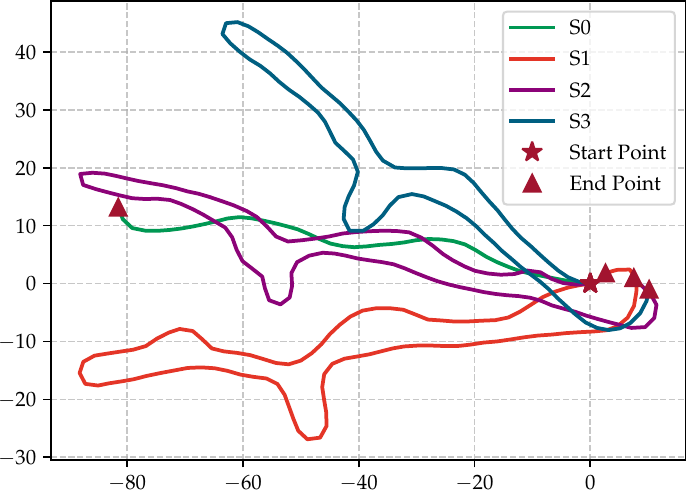}
        \label{fig:s2_mapmerge_concourse_raw}
    }
    \hspace{-0.15cm}
    \subfigure[Concourse - Merged]{
        \includegraphics[width=0.300\linewidth]{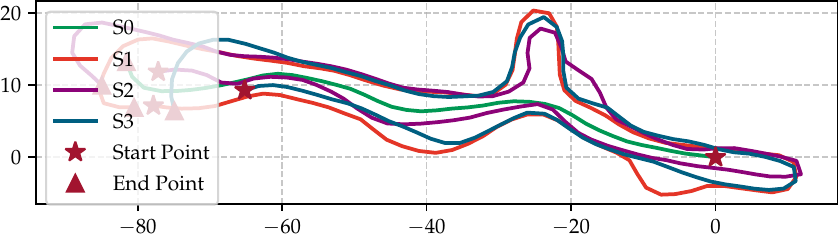}
        \label{fig:s2_mapmerge_concourse_merged}
    }
    % Hall scene
    \subfigure[Hall - Raw]{
        \includegraphics[width=0.153\linewidth]{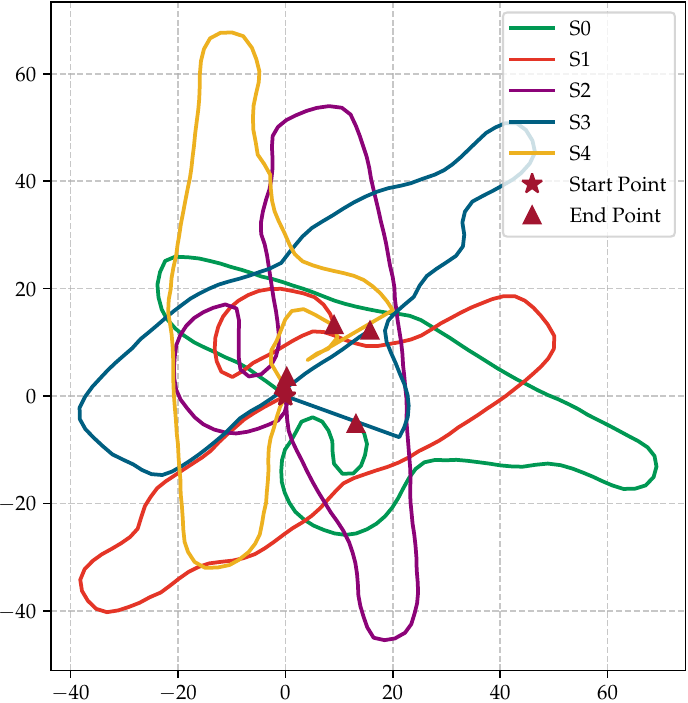}
        \label{fig:s2_mapmerge_hall_raw}
    }
    \hspace{-0.15cm}
    \subfigure[Hall - Merged]{
        \includegraphics[width=0.273\linewidth]{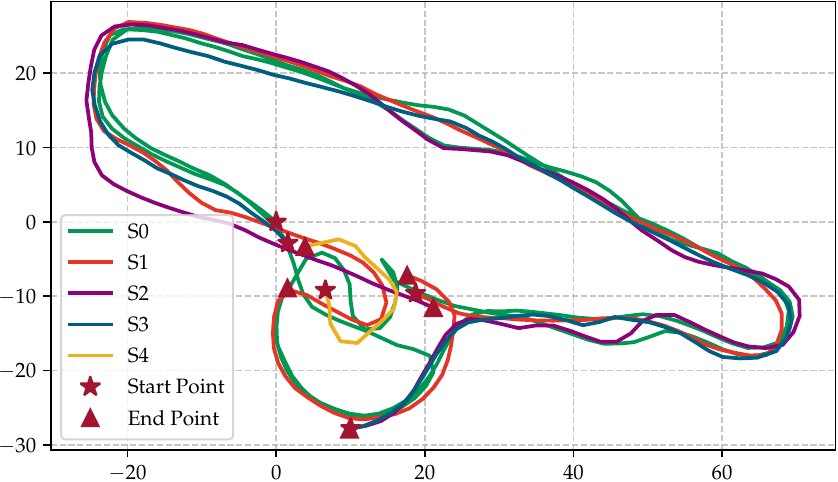}
        \label{fig:s2_mapmerge_hall_merged}
    }
    % Piatrium scene
    \subfigure[Piatrium - Raw]{
        \includegraphics[width=0.225\linewidth]{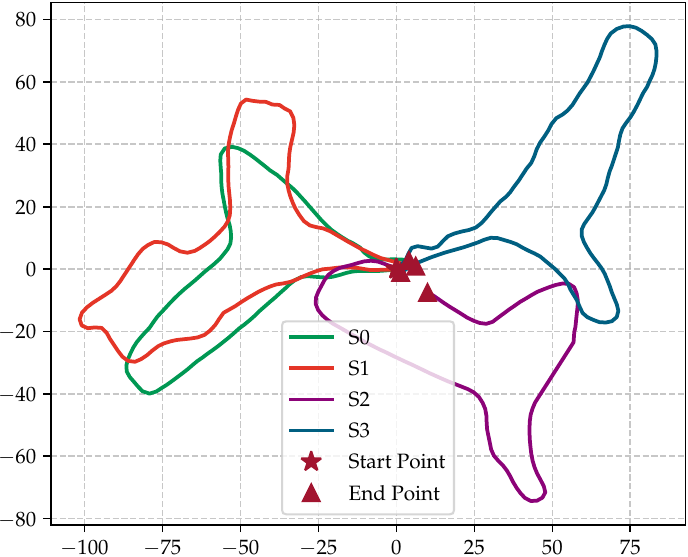}
        \label{fig:s2_mapmerge_piatrium_raw}
    }
    \hspace{-0.15cm}
    \subfigure[Piatrium - Merged]{
        \includegraphics[width=0.200\linewidth]{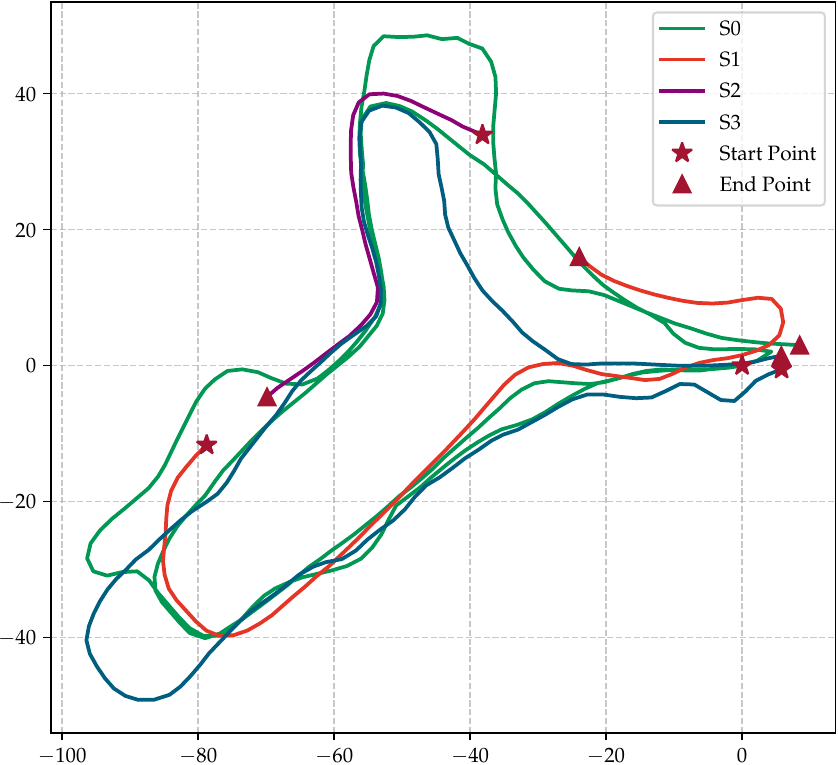}
        \label{fig:s2_mapmerge_piatrium_merged}
    }
    \caption{Map merging results for all four $360$Loc environments across daytime and nighttime sessions. Raw: independent disconnected submaps before merging. Merged: unified topometric map after cross-device session alignment. The pipeline successfully consolidates heterogeneous perspective projections of the same space into a single globally consistent map.}
    \label{fig:s2_mapmerge_overview}
\end{figure}

\begin{figure}[H]
    \centering
    \subfigure[Atrium]{
        \includegraphics[width=0.226\linewidth]{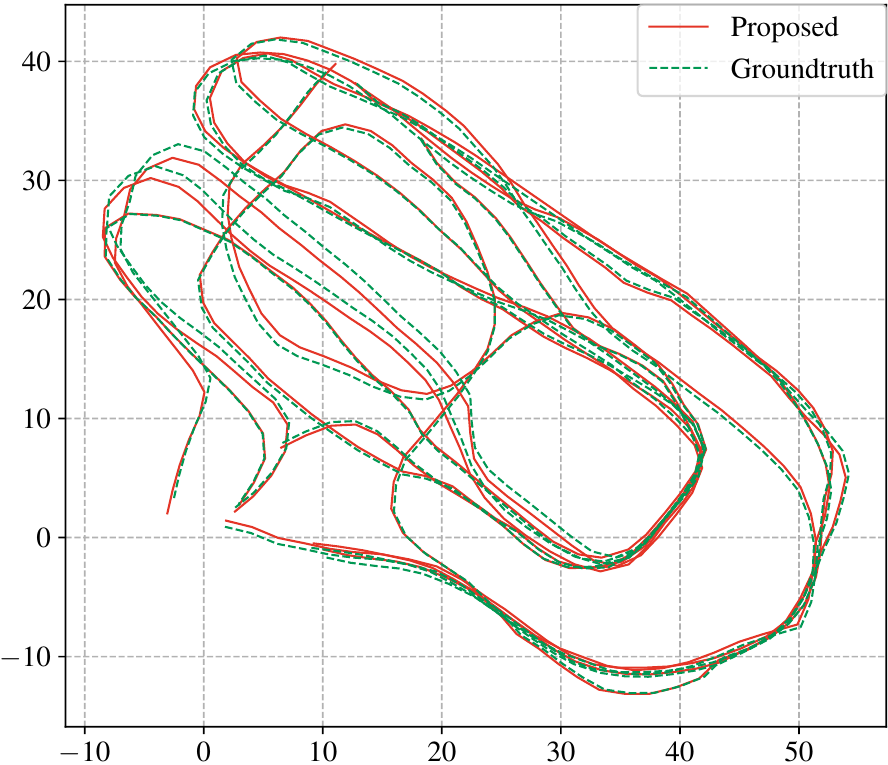}
    }
    \hspace{-0.2cm}
    \subfigure[Concourse]{
        \includegraphics[width=0.231\linewidth]{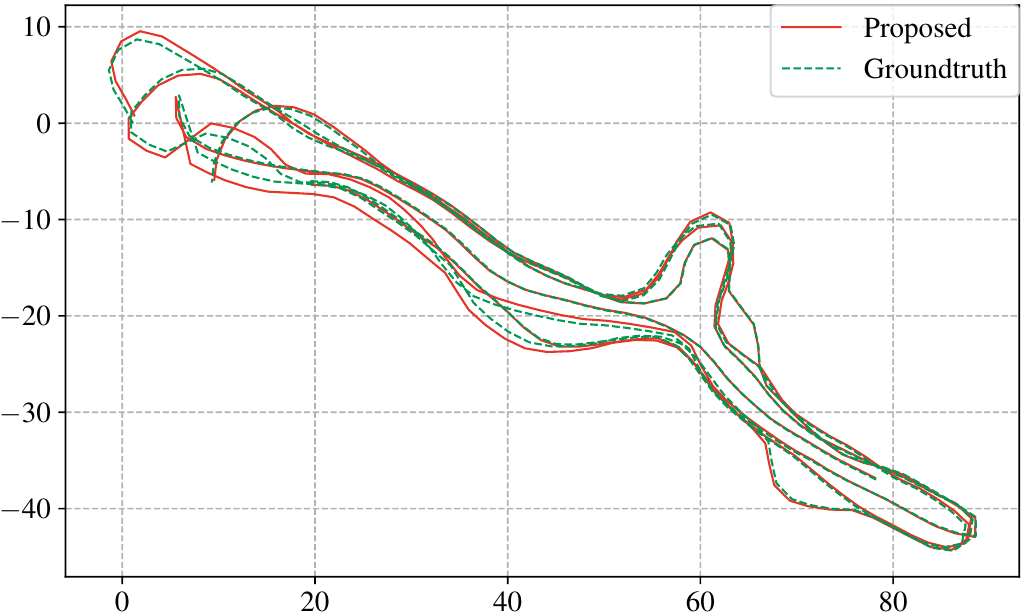}
    }
    \hspace{-0.2cm}
    \subfigure[Hall]{
        \includegraphics[width=0.259\linewidth]{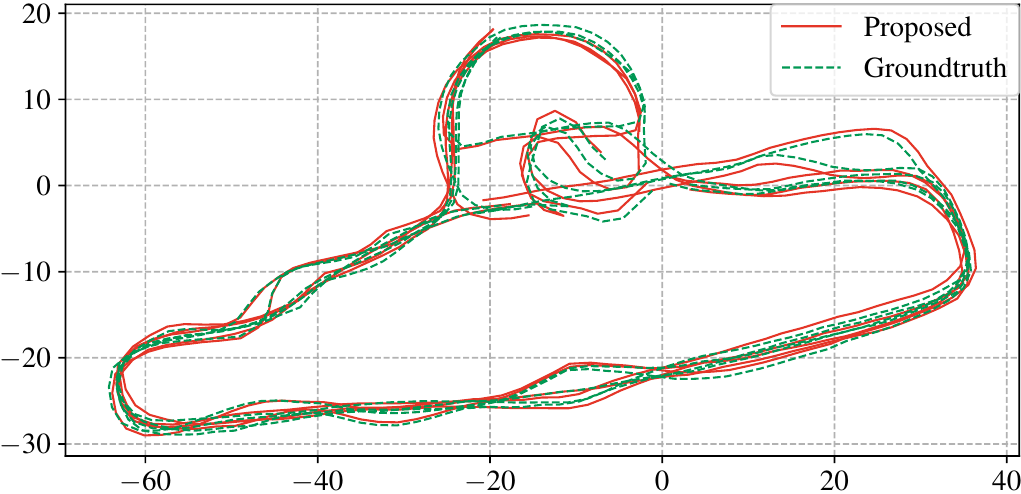}
    }
    \hspace{-0.2cm}
    \subfigure[Piatrium]{
        \includegraphics[width=0.200\linewidth]{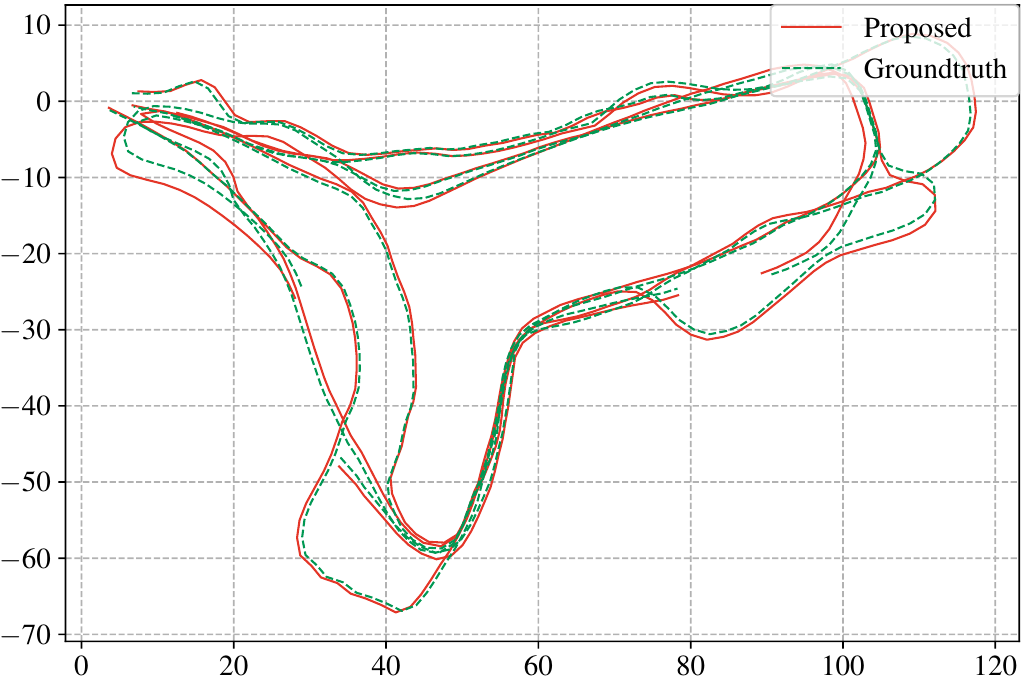}
    }
    \caption{Qualitative trajectory alignment on the $360$Loc dataset after cross-device map merging. Estimated trajectories (colored) are overlaid against ground truth (gray) for all four environments (Atrium, Concourse, Hall, Piatrium). The alignment quality demonstrates that the pipeline successfully handles heterogeneous perspective projections from omnidirectional sources captured under both daytime and nighttime conditions.}
    \label{fig:s2_mapmerge_results}
\end{figure}

The dataset spans four distinct architectural environments: Atrium, Concourse, Hall, and Piatrium.
Each environment contains $4$ to $5$ temporally distinct sequences (\eg \texttt{daytime\_360\_0}, \texttt{nighttime\_360\_1}).
We projected each panoramic image into a perspective frame using the predefined camera intrinsics and extrinsics listed in Tab.~\ref{tab:s2_intrinsics_mapmerge}.
Session identifiers $0$-$4$ correspond to the sequence subsets \texttt{daytime\_360\_0-2} and \texttt{nighttime\_360\_0-1}, respectively.
Fig.~\ref{fig:s2_mapmerge_overview} illustrates the trajectory evolution from raw independent segments to the unified merged map for each environment.
Fig.~\ref{fig:s2_mapmerge_results} presents the qualitative alignment of estimated trajectories against the GT.
Together, these visualizations confirm that the map merging pipeline generalizes across heterogeneous camera models and challenging lighting conditions (daytime vs.\ nighttime).

\begin{table}[t]
  \centering
  \caption{Predefined camera intrinsics and extrinsic rotation (ZYX Euler angles) for each simulated device (session) in the $360$Loc cross-device map merging evaluation. Sessions $0$-$4$ correspond to \texttt{daytime\_360\_0-2} and \texttt{nighttime\_360\_0-1}.}
  \renewcommand\arraystretch{1.00}
  \renewcommand\tabcolsep{8pt}
  \begin{tabular}{c|cccccc|ccc}
    \toprule[0.03cm]
    \multirow{2}{*}{\textbf{Session}} & \multicolumn{6}{c|}{\textbf{Intrinsics}} & \multicolumn{3}{c}{\textbf{Extrinsic Rotation [deg]}}                                                                                                \\
                                      & \textbf{W}                               & \textbf{H}                                       & \textbf{Fx} & \textbf{Fy} & \textbf{Cx} & \textbf{Cy} & \textbf{Z} & \textbf{Y} & \textbf{X} \\
    \midrule[0.03cm]
    \textbf{0}
                                      & $1024$                                   & $576$                                            & $444.5$     & $444.5$     & $512.0$     & $288.0$
                                      & $0.0$                                    & $0.0$                                            & $0.0$                                                                                        \\
    \midrule[0.01cm]
    \textbf{1}
                                      & $1024$                                   & $576$                                            & $593.6$     & $593.6$     & $512.0$     & $288.0$
                                      & $0.0$                                    & $0.0$                                            & $0.0$                                                                                        \\
    \midrule[0.01cm]
    \textbf{2}
                                      & $1024$                                   & $576$                                            & $400.0$     & $400.0$     & $512.0$     & $288.0$
                                      & $0.0$                                    & $0.0$                                            & $0.0$                                                                                        \\
    \midrule[0.01cm]
    \textbf{3}
                                      & $1024$                                   & $576$                                            & $593.6$     & $593.6$     & $512.0$     & $288.0$
                                      & $0.0$                                    & $0.0$                                            & $15.0$                                                                                       \\
    \midrule[0.01cm]
    \textbf{4}
                                      & $1024$                                   & $576$                                            & $444.5$     & $444.5$     & $512.0$     & $288.0$
                                      & $0.0$                                    & $0.0$                                            & $15.0$                                                                                       \\
    \bottomrule[0.03cm]
    \multicolumn{10}{l}{*daytime\_360\_0-2 and nighttime\_360\_0-1 correspond to sessions $0$-$4$.}
  \end{tabular}
  \label{tab:s2_intrinsics_mapmerge}
  \vspace{-0.3cm}
\end{table}

\newpage
\section{Multi-Session Mapping Benefit Study: Extended Setup and Results}\label{sec:s3_nav}

\begin{figure}[H]
  \centering
  \subfigure[Normal setting: per-environment merged-map visualizations and session-by-session exploration trajectories without environmental change. Each column shows one environment (Office, Maze, Tunnel); rows correspond to successive sessions. The traversability graph grows as sessions accumulate, eventually connecting start and goal with a near-optimal path.]{
    \includegraphics[width=\linewidth]{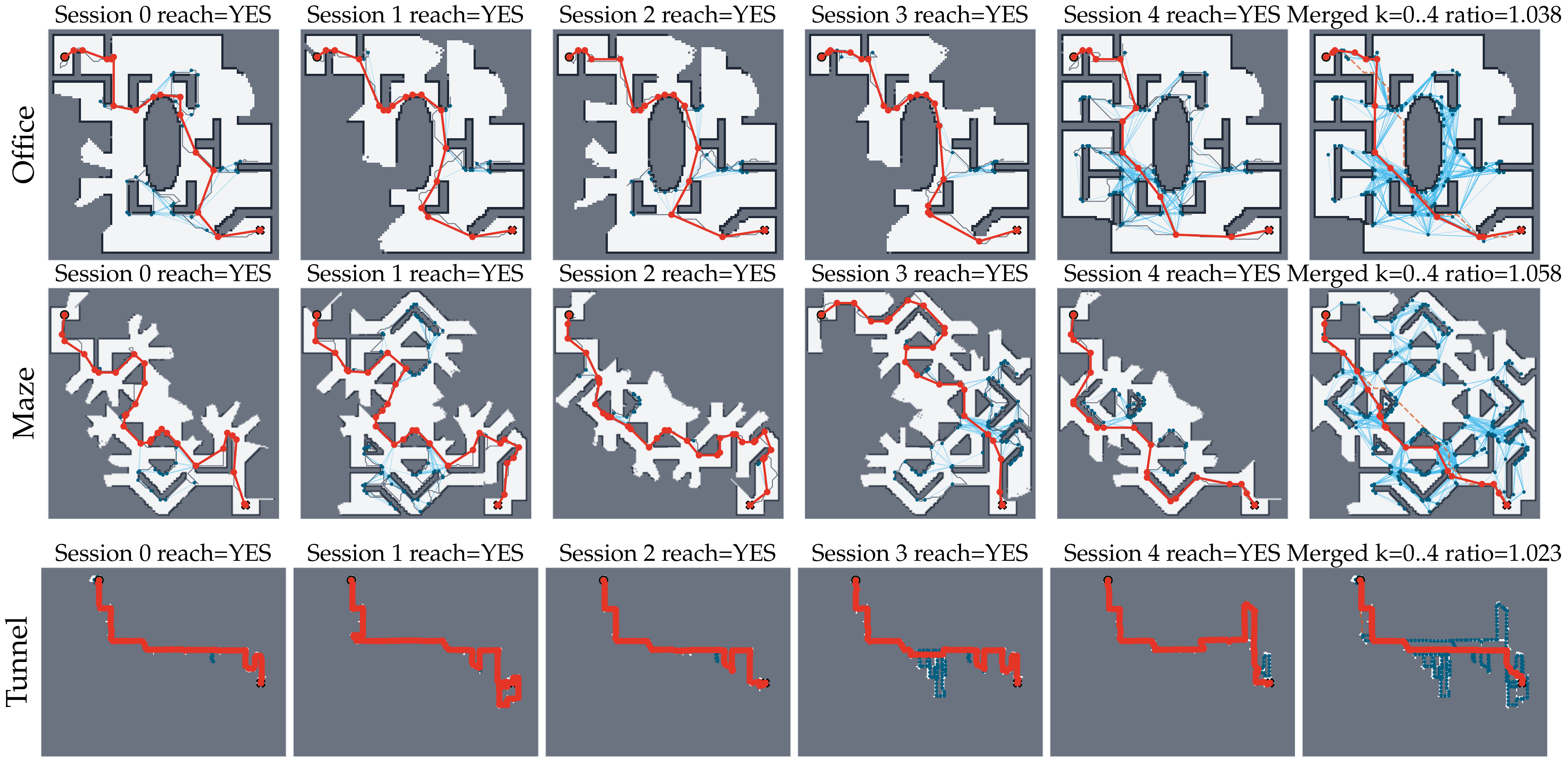}
    \label{fig:s3_nav_normal}
  }
  \subfigure[Dynamic setting: same environments with a temporary rectangular obstacle introduced before session~$0$ and removed from session~$2$ onward. Edges blocked by the obstacle are restored once all traversability criteria are re-satisfied, demonstrating successful re-planning through the reopened corridor.]{
    \includegraphics[width=\linewidth]{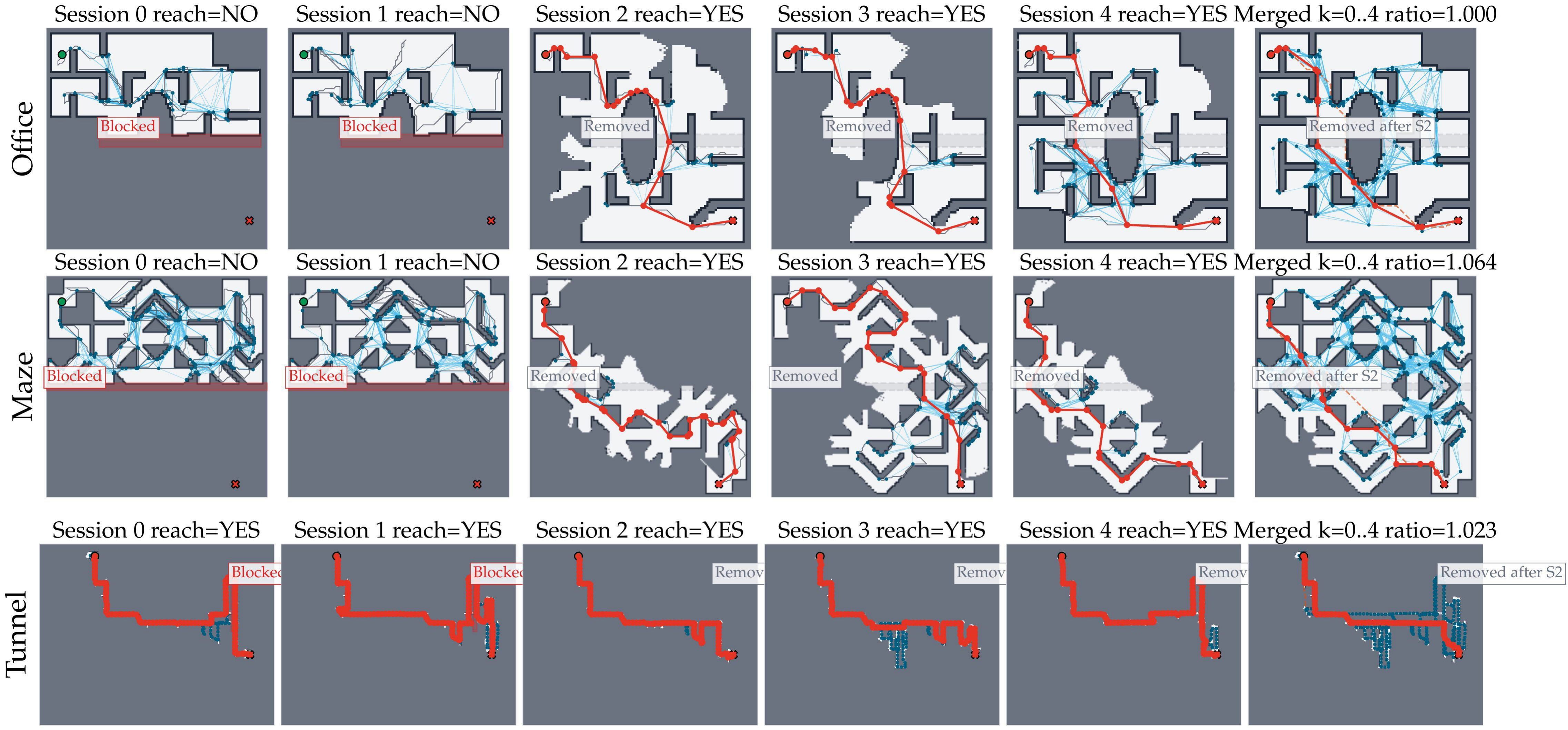}
    \label{fig:s3_nav_dynamic}
  }
  \caption{Session-by-session merged-map and exploration trajectory visualizations for the multi-session navigation benchmark. (a) Normal setting: the traversability graph expands monotonically as sessions merge, achieving near-optimal path ratios. (b) Dynamic setting: a temporary corridor blockage reduces reachability during sessions $0$-$1$; traversability is restored once the obstacle is removed from session~$2$ onward.}
  \label{fig:s3_nav_sessions}
\end{figure}

This section provides full details and extended per-session results for Exp~5 (Sec.~\ref{sec:exp_single_vs_multi} of the main paper).
Whereas the main paper reports final aggregate statistics, here we describe the simulation environment configurations, session collection protocol, traversability edge criteria, and the dynamic obstacle variant in detail, and supplement the summary table with session-by-session trajectory visualizations in Fig.~\ref{fig:s3_nav_sessions}.

\subsection{Benchmark Setup}

\subsubsection{Environments}
We use three simulated environments: Office, Maze, and Tunnel~\cite{zhang2024falcon,cao2022autonomous}, spanning a range of layout types:
Office (${\approx}298$\,m$^2$, compact multi-room), Maze (${\approx}846$\,m$^2$, structured narrow-corridor), and Tunnel (${\approx}75{,}000$\,m$^2$, large open-topology space).
Each environment is provided as a point-cloud map and rasterized into a $2$D occupancy grid at $0.2$m/cell resolution, with the $XY$-plane as the ground plane.
A single start-goal pair, fixed in world coordinates, is shared across all sessions within each environment.
This design holds the navigation objective constant, isolating the effect of cumulative map coverage on reachability and path optimality.

\subsubsection{Session Collection Protocol}
Within each environment, a frontier-based exploration agent autonomously collects mapping sessions from the fixed start location.
(\ie A frontier is any navigable cell adjacent to unvisited space; the agent selects among frontiers to maximize coverage.)
A randomly sampled initial yaw angle (\ie shared random seed $=42$ for reproducibility) and a softmax temperature of $2.5$ introduce trajectory diversity across sessions
(\ie the softmax temperature controls the frontier selection distribution: higher temperature broadens the sampling toward distant frontiers, producing more varied traversal paths across sessions).
Each session terminates when the agent reaches the goal or exhausts its session budget.
We collect $5$ sessions for Office and Maze, and $10$ sessions for Tunnel, whose larger extent requires more traversals to achieve adequate coverage.

\subsubsection{Cumulative Map Merging}
After each session, a topometric map is constructed from the exploration trajectory and merged into the accumulated map.
A traversability edge is added between two nodes only when all three criteria are simultaneously satisfied:
\begin{enumerate}[label=\arabic*), noitemsep, topsep=2pt]
  \item their Euclidean distance is below a threshold,
  \item the line-of-sight between them is unobstructed in the occupancy grid, and
  \item $A^*$ path planning confirms reachability.
\end{enumerate}
This conservative policy ensures that every edge in the traversability graph corresponds to a physically navigable transition.

\subsubsection{Dynamic Variant}
To evaluate adaptation to environmental change, we introduce a rectangular obstacle blocking a key corridor in each environment's occupancy grid before session~$0$ and remove it from session~$2$ onward, emulating a temporarily blocked corridor.
The merged traversability graph is re-evaluated after each new session: edges previously blocked by the obstacle are restored once all three traversability criteria are again satisfied, enabling re-planning through the newly reopened path.

\begin{table}
  \centering
  \caption{Final-session results of the multi-session navigation benchmark. Reachability reports the number of sessions in which Dijkstra's algorithm finds a connected path from start to goal out of the total. Ratio $r = \ell_\text{topo}/\ell_\text{GT}$ measures path optimality ($r = 1.0$ is optimal). All normal-setting environments achieve full reachability with near-optimal ratios ($r \leq 1.06$). In the dynamic setting, reachability drops to $3/5$ for Office and Maze while the corridor obstacle is present, and recovers once it is removed.}
  \renewcommand\arraystretch{1.05}
  \renewcommand\tabcolsep{10.6pt}
  \label{tab:s3_nav_results}
  \begin{tabular}{ccccc}
    \toprule
    Environment & Variant & Reachability & Ratio $r$ & Coverage \\
    \midrule
    \multirow{2}{*}{Office} & Normal  & $5/5$   & $1.038$ & $259.0$\,m$^2$ ($87.0\%$) \\
                             & Dynamic & $3/5$   & $1.000$ & $265.5$\,m$^2$ ($89.2\%$) \\
    \midrule
    \multirow{2}{*}{Maze}   & Normal  & $5/5$   & $1.058$ & $663.7$\,m$^2$ ($78.5\%$) \\
                             & Dynamic & $3/5$   & $1.064$ & $681.5$\,m$^2$ ($80.6\%$) \\
    \midrule
    \multirow{2}{*}{Tunnel} & Normal  & $10/10$ & $1.010$ & $5111.2$\,m$^2$ ($6.8\%$) \\
                             & Dynamic & $10/10$ & $1.010$ & $5270.7$\,m$^2$ ($7.1\%$) \\
    \bottomrule
  \end{tabular}
\end{table}

\subsubsection{Evaluation Metrics}
We report three metrics:
\begin{itemize}
  \item \textbf{Reachability}: the number of accumulated sessions in which Dijkstra's algorithm finds a connected path from start to goal on the merged traversability graph, out of the total number of sessions.
  \item \textbf{Topometric shortest-path ratio} $r = \ell_{\text{topo}} / \ell_{\text{GT}}$: the planned path length on the topometric graph normalized by the ground-truth $A^*$ shortest-path length on the occupancy grid; values near $1.0$ indicate near-optimal planning.
  \item \textbf{Explored free-space coverage} (m$^2$): the cumulative free area visited across all sessions, quantifying spatial coverage growth.
\end{itemize}

\subsection{Results}

Tab.~\ref{tab:s3_nav_results} reports final-session results across all six configurations.
In the normal setting, every environment achieves full reachability with a shortest-path ratio in the range $1.01$-$1.06$, confirming that the merged traversability graph supports near-optimal path planning.

In the dynamic setting, reachability drops to $3/5$ for Office and Maze during the obstacle-present phase (sessions $0$-$1$), because the single blocked corridor prevents path completion.
Reachability is restored from session~$2$ onward once the obstacle is removed; the recovered path ratios ($r=1.000$ for Office, $r=1.064$ for Maze) confirm near-optimal re-planning through the reopened corridor.
Tunnel remains fully reachable ($10/10$) in both settings because its open topology provides alternative routes around the blocked region.

Tunnel achieves only ${\approx}7\%$ coverage despite accumulating ${\approx}5{,}000$\,m$^2$ of explored area, because the total navigable space exceeds $70{,}000$\,m$^2$; its large, open-topology layout means that even extensive multi-session traversal covers a small fraction of the environment.
Session-by-session trajectory plots and per-environment merged-map visualizations are provided in Fig.~\ref{fig:s3_nav_sessions}.

\end{document}